\documentclass[letterpaper,twocolumn,10pt]{article}
\usepackage{usenix}

\usepackage{tikz}
\usepackage{amsmath}

\usepackage{multirow}
\usepackage{pifont}
\usepackage{caption}
\usepackage{enumitem}
\usepackage{amssymb}
\usepackage{subcaption}
\usepackage{graphicx}
\usepackage{booktabs}
\usepackage[ruled,linesnumbered]{algorithm2e}
\usepackage{etoolbox}
\usepackage{mdframed}

\newmdenv[
    linewidth=2pt,
    leftline=true,
    rightline=false,
    topline=false,
    bottomline=false,
    linecolor={rgb:red,0.180;green,0.459;blue,0.714},    
    backgroundcolor=gray!10,
    innerleftmargin=5pt,
    innerrightmargin=5pt,
    innertopmargin=5pt,
    innerbottommargin=5pt
]{observationbox}
\usepackage{xspace}
\newcommand{\sys}{{\sc Void}\xspace}
\usepackage[table]{xcolor}
\definecolor{Gray}{gray}{0.9}

\SetCommentSty{BlueCommentSty}

\usepackage[capitalise]{cleveref}

\usepackage[available]{usenixbadges}

\begin{document}
\pagestyle{empty}

\title{\sys: Defeating Unauthorized Mimicry in Latent Diffusion Models}

\author{
{\rm Chunlin Qiu$^{1}$, \quad Ang Li$^{1}$, \quad Tianxiao Huang$^{1}$, \quad Ruilin Gan$^{2}$, \quad Yunjie Ge$^{3}$,}\\
{\rm Shenyi Zhang$^{3}$, \quad Huayi Duan$^{4}$, \quad Lingchen Zhao$^{1}$, \quad Chao Shen$^{5}$, \quad Qian Wang$^{1}$\thanks{Corresponding author.}}\\
{$^1$} School of Cyber Science and Engineering, Wuhan University,\\
$^2$School of Computer Science, Wuhan University,\\
$^3$Institute for Math\&AI, Wuhan University,\\
$^4$The Hong Kong University of Science and Technology (Guangzhou),\\
$^5$School of Cyber Science and Engineering, Xi'an Jiaotong University}

\maketitle

\begin{abstract}

While Latent Diffusion Models (LDMs) have revolutionized visual synthesis, they are increasingly exploited for unauthorized mimicry of individuals. Existing defenses inject deceptive perturbations to steer the generated images toward irrelevant targets. However, this approach hinges on an ungrounded assumption: subtle perturbations can maintain their deceptive efficacy throughout an LDM's extensive generation process. In reality, the model's innate restoration mechanism will remove such perturbations and cause individual identities to re-emerge in the images generated.

We propose \sys, a defense framework that overcomes this conundrum by manipulating an LDM's intrinsic stochasticity. \sys perturbs the diffusion pipeline in two novel ways: 1) amplifying the latent encoding errors to shatter an image's semantic structure, and 2) counteracting the target guidance signals to suppress the model's restoration capabilities. This results in a semantic corruption that thwarts any unauthorized mimicry. Notably, the security gain does not come at the price of visual utility, as \sys simultaneously manages to confine perturbations to human-imperceptible regions of protected images. Our comprehensive evaluation of 24 state-of-the-art defenses against 10 mimicry attacks on 5 datasets demonstrates \sys's unprecedented protection power: it increases the average Fréchet Inception Distance (FID) from 113 to 365, a 223\% improvement over the strongest defense to date.

\end{abstract}
\section{Introduction}
\label{sec:intro}

Latent Diffusion Models (LDMs)~\cite{ldm,sd,lcm} have emerged as the dominant framework for high-fidelity visual synthesis. Compared to traditional generative models, LDMs demonstrate superior capabilities in zero-shot image editing~\cite{sdedit,controlnet} and few-shot model personalization~\cite{dreambooth, lora}. These large-scale pre-trained models function as highly personalized creation tools for generating specific contents or artistic styles across diverse scenarios. Furthermore, their computational efficiency significantly reduces hardware requirements, enabling individual users to leverage sophisticated synthesis capabilities on consumer-grade devices. However, the barriers are equally low for malicious actors to perform unauthorized mimicry, e.g., cloning sensitive facial identities or distinctive artistic styles with only a handful of reference images. This poses severe threats to personal privacy~\cite{antidb,metacloak} and intellectual property rights~\cite{advdm, glaze}, raising critical concerns about the abuse of generative AI.

\begin{figure}[t]
\centering
\includegraphics[width=\columnwidth]{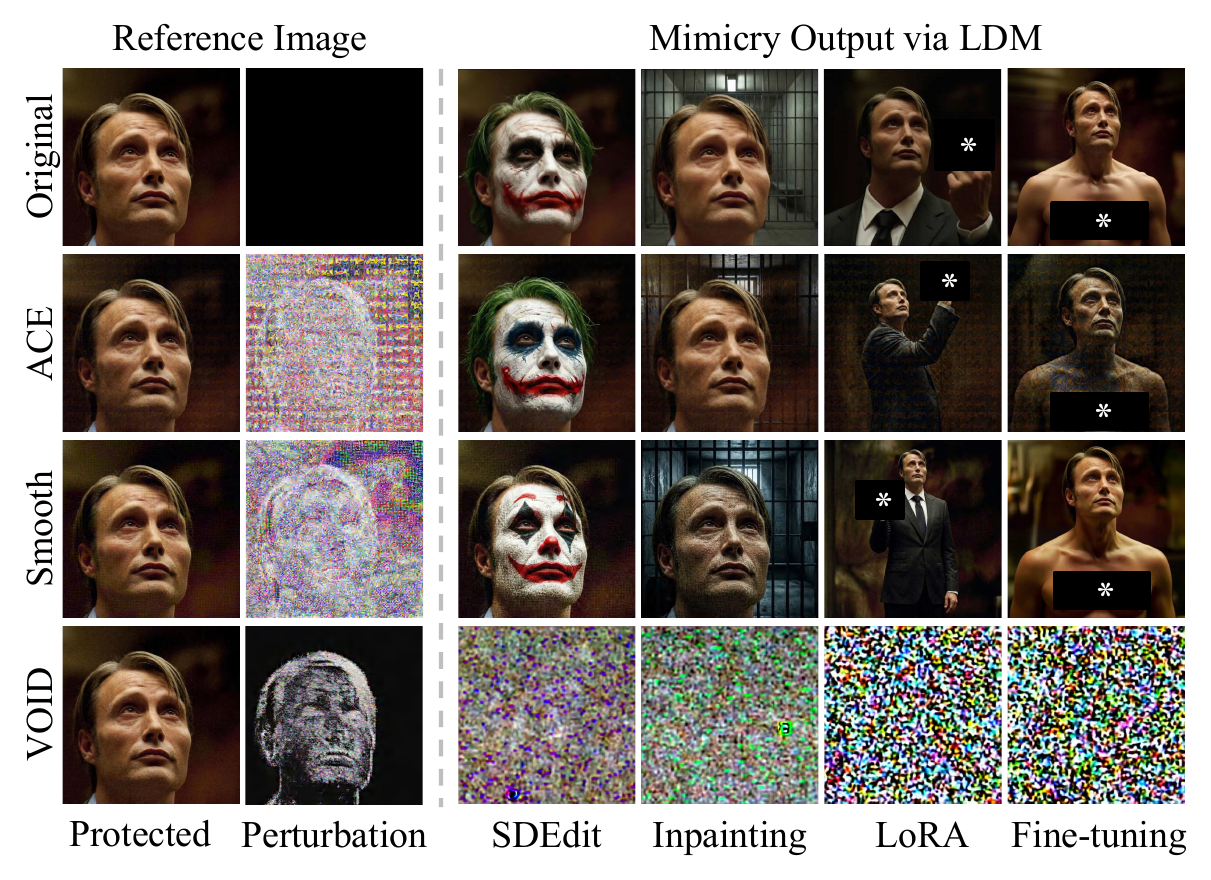}
\caption{Visual comparison of defenses against unauthorized mimicry. While SOTA methods fail to prevent high-fidelity identity reconstruction, \sys fundamentally collapses the generation trajectory, rendering synthesized outputs semantically void. \emph{Sensitive content is redacted for presentation.}}
\label{fig:mimicry_comparison}
\end{figure}

To counter such threats, a list of defense mechanisms~\cite{glaze, mist, advdm, antidb} have been proposed. All of them follow a \emph{semantic-steering} paradigm:  
injecting subtle perturbations into an image that needs protection prior to its release, in the hope of deceiving LDMs and steer their denoising trajectory away from the image's original semantics. The steering process can apply to either the initial latent encoding~\cite{glaze, mist} or the subsequent denoising process~\cite{advdm, antidb}.

However, our extensive empirical analysis reveals a critical vulnerability in this approach, where defensive perturbations fail to sustain their deceptive efficacy along an LDM's generation process. As illustrated by the visual comparisons in~\cref{fig:mimicry_comparison}, even the most advanced defense mechanisms, ACE~\cite{ace} and Smooth~\cite{smooth}, fail to prevent the reconstruction of identities they intend to protect. 
Evidently, superficial visual perturbations are insufficient to prevent the synthesis of semantic content.

We further reveal that the failure of semantic-steering defenses is inevitable due to the fundamental design of LDMs.
Specifically, a protected identity effectively acts as a robust semantic anchor that allows LDMs to discard protective perturbations as additive noise. Moreover, defensive signals will undergo structural suppression through the VAE~\cite{vae} component's encoding and the subsequent diffusion noise injection process. Consequently, the weakened perturbations fail to override the model's strong generative restoration priors.

\begin{figure}[t]
\centering
\includegraphics[width=\columnwidth]{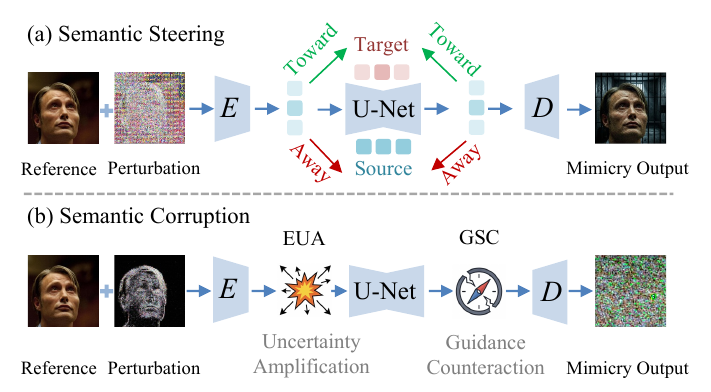}
\caption{Comparison between the existing semantic steering paradigm and our proposed semantic corruption paradigm.}
\label{fig:paradigm_comparison}
\end{figure}

To overcome these systemic limitations, we propose \sys, a defense framework that realizes an alternative \emph{semantic-corruption} paradigm. As shown in \cref{fig:paradigm_comparison}, rather than contending with an LDM's denoising instincts, \sys embraces and weaponizes the model's intrinsic stochasticity. Our core insight is that, 
while the model always purifies external protective perturbations, it cannot distinguish or suppress its own \emph{internal probabilistic errors}.
By amplifying these errors, we can trigger a cascading trajectory divergence that shatters semantic structure and ensure that mimicry of protected images degenerates into meaningless noise.

Specifically, \sys perturbs the LDM pipeline in two novel ways. First, the variational autoencoder (VAE)~\cite{vae} always introduces probabilistic noise when encoding an image into a latent space, and we amplify such encoding uncertainty to an extent that it becomes irreducible. Second, after the U-Net diffusion process the latent feature vector will be decoded back to an image steered by the Classifier-Free Guidance (CFG) mechanism~\cite{cfg}, yet we manage to neutralize its conditional steering effect so that the restoration trajectory collapses. We realize both techniques as perturbations on the image under protection, without modifying any part of the LDM itself.

A distorted image will be of little use. We thereby incorporate a Just Noticeable Difference (JND)~\cite{tdjnd} model to estimate perceptual sensitivity, effectively confining perturbations to imperceptible regions. Since this restricts our security budget, we further develop a customized perturbation selection strategy to ensure robust defense. Orchestrating all these techniques into a joint optimization process, \sys offers strong resilience to mimicry while maintaining the utility of images.

We validate \sys through an extensive evaluation involving 10 mimicry attacks and 24 state-of-the-art defenses across 5 diverse datasets. Quantitative results show that \sys raises the average FID (Fréchet Inception Distance, the standard quality metric for generative models) from 113 to 365, a 223\% improvement over the second best defense, while reducing latent identity correlation to near-zero. Meanwhile, \sys maintains superior visual utility with a PSNR~\cite{psnr} of 34.41 dB and an SSIM~\cite{ssim} of 0.89; it stands as the only defense in our benchmarks to achieve a perceptual LPIPS~\cite{lpips} below 0.1. Moreover, our stress tests demonstrate its transferability across diverse LDM variants like SD v2.1~\cite{sdv2.1} and LCM~\cite{lcm} as well as its resilience against advanced adaptive countermeasures including prior-guided purification and mechanism-aware adaptation. These comprehensive results establish \sys as a powerful and highly effective tool for proactive protection against generative mimicry.

In summary, we make the following contributions:
\begin{itemize}
\item We provide the first systematic study to dissect the semantic-steering paradigm for countering LDM-based mimicry threats and reveal its inherent limitations.
\item We propose \sys, a novel framework that shifts the defensive paradigm to semantic corruption. It leverages the inherent stochasticity of LDMs to defeat mimicry attacks without degrading the utility of protected images.
\item 
We establish comprehensive benchmarks with state-of-the-art defenses against attacks over diverse datasets, demonstrating the superior performance of \sys.
\end{itemize}
\section{Background and Related Work}

\vspace{0.5em}
\subsection{Unauthorized Mimicry in LDMs}
Latent Diffusion Models (LDMs)~\cite{ldm} enable efficient image editing and personalization, but the same capabilities also lower the cost of mimicking identities, objects, and artistic styles from reference images. We defer technical details of LDMs to \cref{sec:appendix_ldm}. Unauthorized mimicry generally follows two routes: training-based model personalization and inference-based image editing.

\noindent\textbf{Training-based model personalization:}
This attack family uses the victim's images as unauthorized training data to bind an identity, object, or style to a model or a learnable token~\cite{ldm,dreambooth,ti}. The adversary first uses the VAE~\cite{vae} encoder~$\mathcal{E}$ to map each image~$x$ into the latent space, obtaining $z_0 = \mathcal{E}(x)$. Personalization is then performed by minimizing the diffusion training loss on these unauthorized latents:
\begin{equation}
\label{eq:training_objective}
\min_{\theta'} \mathbb{E}_{t, z_0, \epsilon, c} [\|\epsilon - \epsilon_{\theta'}(z_t, t, c)\|_{2}^2],
\end{equation}
where $z_t = \sqrt{\bar{\alpha}_t} z_0 + \sqrt{1 - \bar{\alpha}_t} \epsilon$ is the noisy latent at timestep~$t$.

Here, $\epsilon_{\theta'}$ denotes the noise predictor after updating the trainable variables~$\theta'$, and $c$ denotes the conditioning signal, such as a text prompt. These attacks mainly differ in what variables are updated. \emph{Full Fine-tuning (FT)} updates all model parameters, with \emph{DreamBooth (DB)}~\cite{dreambooth} further using prior preservation to improve fidelity. \emph{Parameter-Efficient Fine-Tuning (PEFT)}~\cite{peft}, such as \emph{LoRA}~\cite{lora} and \emph{SVDiff}~\cite{svd}, updates only a small set of low-rank or singular-value parameters. \emph{Embedding Optimization}, such as \emph{Textual Inversion}~\cite{ti}, freezes the model and optimizes a textual embedding that serves as a trigger for the unauthorized concept.

\noindent\textbf{Inference-based image editing:}
This attack family uses off-the-shelf LDMs to mimic a protected concept without updating model parameters. It typically consists of inversion and guided generation. First, inversion maps a reference image into a latent state $z_t$, either by adding noise to the image, as in \emph{SDEdit}~\cite{sdedit} and \emph{Inpainting}~\cite{ldm}, or by reversing the diffusion trajectory, as in \emph{DDIM Inversion}~\cite{ddim} and \emph{DiffEdit}~\cite{diffedit}. This step preserves source information that can later be reused during generation. The subsequent generation is then steered toward a target concept using conditioning signals. Classifier-Free Guidance (CFG)~\cite{cfg} provides semantic control, while \emph{ControlNet}~\cite{controlnet} adds structural constraints such as edges or poses to preserve spatial alignment with the source image. Compared with training-based personalization, these attacks avoid fine-tuning and therefore reduce the cost of unauthorized mimicry, but they still rely on the reference image to provide semantic or structural guidance.

Together, training-based personalization and inference-based editing cover two major unauthorized mimicry surfaces. The former learns the protected concept through parameter or embedding updates, while the latter reuses a protected image through inversion and guidance. These attacks differ in cost, access assumptions, and affected pipeline stages, but both can reproduce identities, objects, or styles from reference images. Therefore, a practical defense should be evaluated against both training-time and inference-time mimicry.

\subsection{Defenses against LDM-based Mimicry}

To safeguard visual privacy against unauthorized mimicry, existing defenses protect images by adding subtle perturbations~$\delta$ ($\|\delta\|_p \leq \alpha$) before release. Most methods follow semantic steering, where the perturbation redirects the LDM~\cite{ldm} away from the original semantics or toward an irrelevant target. These defenses can be described by their adversarial setting, such as evasion against frozen models~\cite{advdm,photoguard} or poisoning against future personalization~\cite{antidb,simac}, and by their main target in the LDM~\cite{ldm} pipeline, such as VAE~\cite{vae} encoding, U-Net~\cite{unet} denoising, or full-pipeline optimization.

Defenses that target encoding mainly perturb the representation before denoising. Methods such as \emph{Glaze}~\cite{glaze} and \emph{NightShade}~\cite{nightshade} add pixel-space perturbations so that the VAE~\cite{vae} encoder~$\mathcal{E}$ maps the protected image to a misleading latent representation. Methods that operate in the latent space, such as LDS~\cite{lds}, further optimize the steering objective in the latent domain, but they still rely on shifting the protected representation toward an external adversarial direction. The objective can be written as
\begin{equation}
\label{eq:encoding_defense}
    \min_{\delta} L(\mathcal{E}(x+\delta), z_{ref}) \quad \text{s.t.} \quad \|\delta\|_p \leq \alpha,
\end{equation}
where $L$ is a distance metric and $z_{ref}$ specifies either a decoy latent target or a direction away from the original latent.

Defenses that target the denoising trajectory optimize perturbations against the U-Net~\cite{unet} noise predictor~$\epsilon_\theta$ at selected diffusion timesteps under the conditioning signal. In evasion settings, methods such as \emph{AdvDM}~\cite{advdm} and \emph{Smooth}~\cite{smooth} optimize perturbations against frozen LDMs by changing their noise estimates across timesteps, with Smooth~\cite{smooth} emphasizing later timesteps where editing noise is stronger to improve robustness for inference-based editing. Other methods use different timestep ranges. For example, \emph{SimAC}~\cite{simac} and \emph{GAPDiff}~\cite{gapdiff} focus on earlier timesteps during diffusion training and report stronger protection against personalization-based mimicry, while \emph{AntiDB}~\cite{antidb} incorporates denoising objectives into a training-aware optimization for personalization. The objective can be written as
\begin{equation}
\label{eq:denoising_defense}
    \min_{\delta} \mathbb{E}_{t} \left[ L(\epsilon_\theta(z_t(x+\delta), t, c), y_{ref}) \right] \quad \text{s.t.} \quad \|\delta\|_p \leq \alpha,
\end{equation}
where $y_{ref}$ specifies the steering target, either a decoy noise pattern or a signal that increases prediction error.

Beyond the steering target, defenses also differ in how the perturbation is optimized. Evasion-based methods keep the LDM parameters fixed and update only the perturbation~$\delta$, often with PGD~\cite{pgd} optimization, to construct protected samples. This setting may use VAE losses, U-Net losses, or their combination, as in \emph{Mist}~\cite{mist} and \emph{PhotoGuard}~\cite{photoguard}. Poisoning-based methods further account for model updates during personalization, using training-aware or bi-level objectives to optimize protected samples, as in \emph{ACE}~\cite{ace}, \emph{MetaCloak}~\cite{metacloak} and \emph{Pretender}~\cite{pretender}. Despite these different optimization regimes, they still rely on steering losses that shift latent representations or denoising trajectories away from protected semantics.

Overall, existing defenses differ in where and how they optimize perturbations, but share the same semantic-steering assumption. A small external signal is expected to persist through the LDM pipeline and redirect generation away from protected semantics. However, LDMs restore coherent semantics from noisy latents, suppressing such external signals while retaining the original semantic anchor. As a result, these defenses often provide limited and attack-specific protection.
\section{Threat Model}
The threat model involves two parties: a \textit{Data Owner} (defender) and an \textit{Adversary} (attacker).

\vspace{.25em}

\noindent\underline{\textit{{Adversary.}}}
The adversary aims to perform successful mimicry using the data owner's images without authorization.
\begin{itemize}[leftmargin=*]
\item \textbf{Capabilities:}
The adversary’s capabilities are defined by their knowledge of the defense. Black-box attackers possess no specific knowledge of the protective perturbations but are capable of employing general pre-processing schemes~\cite{jpeg}, adversarial purification~\cite{gridpure}, or extensive hyper-parameter tuning~\cite{impress++} to evade potential protection. White-box attackers are assumed to have full access to a defense's internals, including the specific LDM used to craft the defense and the underlying algorithm. This comprehensive knowledge allows them to devise specialized adaptive attacks~\cite{impress,impress++} or modify the mimicry pipeline~\cite{diffshortcut} to directly circumvent the protection.
\item \textbf{Objectives:} The primary goal of an adversary is to produce high-fidelity unauthorized mimicry that aligns with adversarial instructions. This involves two competing requirements, which are the successful replication of identity or style features from the source images and the accurate execution of desired modifications, such as changes in background or context.
\end{itemize}

\noindent\underline{\textit{{Data Owner.}}}
The data owner seeks to protect personal images by applying protective perturbations to render the aforementioned mimicry attempts ineffective.
\begin{itemize}[leftmargin=*]
\item \textbf{Capabilities:} We assume the data owner relies solely on a locally available LDM to construct defensive perturbations, with no prior knowledge of the adversary's specific LDM variants, mimicry regimes, or configurations across adversarial settings.
\item \textbf{Objectives:} The data owner adheres to three primary goals for the protection: (1) Effectiveness: The defense should fundamentally disrupt the adversary's generation process ensuring that any mimicry attempt fails both identity preservation and intent alignment; (2) Utility: The defense should preserve the high visual quality of the protected images, ensuring they retain their utility for legitimate public display with satisfactory perceptual quality; (3) Robustness: The defense should demonstrate transferability across heterogeneous mimicry pipelines and maintain strong resilience against diverse adversarial countermeasures, ranging from generic countermeasures to white-box adaptive attacks.
\end{itemize}
\section{Analysis of Defense Vulnerability}
\label{sec:fragility_analysis}

Existing semantic-steering defenses hinge on the assumption that subtle perturbations can maintain their deceptive efficacy throughout the extensive denoising process of LDMs. In reality, models effectively purify these signals to recover the protected identity. We trace this vulnerability to two systemic root causes: intrinsic feature persistence (\cref{sec:limitations_feature}) and structural signal degradation (\cref{sec:limitations_signal}).

\begin{figure}[tbp!]
    \captionsetup[subfigure]{labelformat=empty}
    \centering
    \begin{subfigure}[b]{0.3\linewidth}
        \centering
        \includegraphics[width=\linewidth, keepaspectratio]{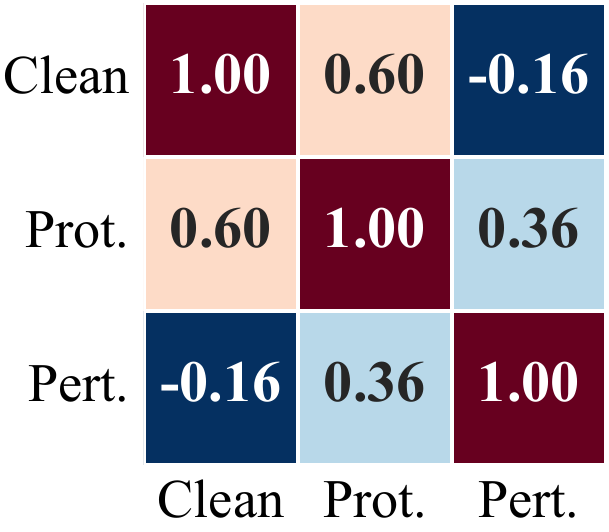}
        \label{fig:adversarial0}
    \end{subfigure}
    \hspace{10pt}
    \begin{subfigure}[b]{0.3\linewidth}
        \centering
        \includegraphics[width=\linewidth, keepaspectratio]{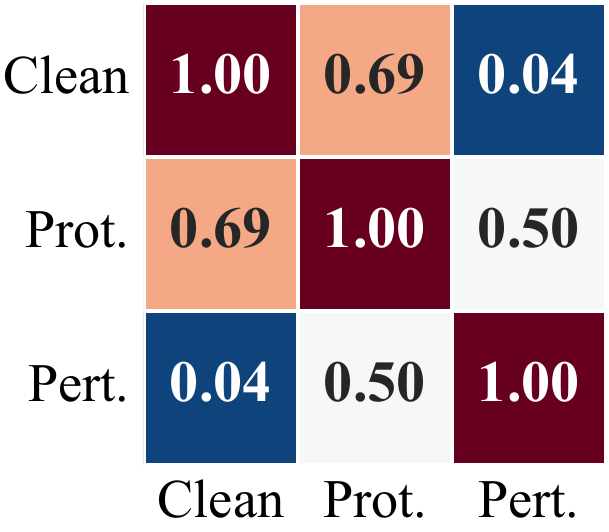}
        \label{fig:poisoning0}
    \end{subfigure}
    \hspace{10pt}
    \raisebox{1.5em}{\includegraphics[height=2.15cm]{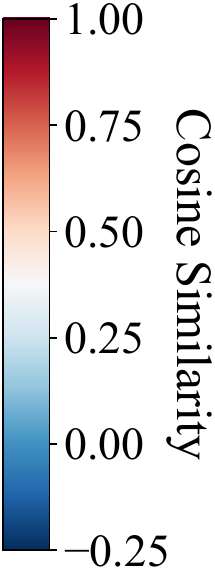}}
    \caption{Pairwise latent cosine similarity between clean, protected, and perturbation samples, comparing evasion-based~\cite{advdm} (left) and poisoning-based~\cite{antidb} (right) defenses.}
    \label{fig:latent_similarity}
\end{figure}
\begin{figure}[tbp!]
    \centering
    \begin{subfigure}[b]{0.4\linewidth}
        \centering
        \includegraphics[width=\linewidth]{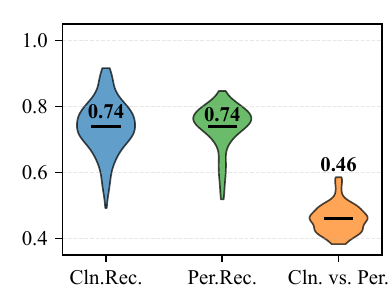}
        \caption{CLIP similarity}
        \label{fig:clip_violin}
    \end{subfigure}
    \begin{subfigure}[b]{0.4\linewidth}
        \centering
        \includegraphics[width=\linewidth]{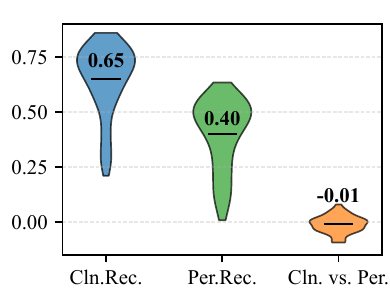}
        \caption{DINO similarity}
        \label{fig:dino_violin}
    \end{subfigure}
    \caption{Quantitative reconstruction quality, showing high fidelity for both reconstructed images and perturbations.}
    \label{fig:reconstruction_quality}
\end{figure}
\subsection{Persistence of Intrinsic Features}
\label{sec:limitations_feature}
We first investigate the semantic integrity of latent representations across 24 state-of-the-art defenses, all of which adhere to the semantic-steering paradigm. Specifically, we project original images, protected samples, and isolated perturbations from the TI-Dataset~\cite{ti} into the VAE~\cite{vae} latent space of Stable Diffusion v1.5~\cite{sdv1.5}. By computing pairwise cosine similarities among these components, we quantify the semantic correlations between protected outputs and original identities to evaluate the actual degree of identity erasure achieved by current methods.

\begin{observationbox}
\textit{\textbf{Observation 1:} Existing defenses fail to decouple the latent representation from the original identity, leading to semantic persistence.}
\end{observationbox}

Ideally, an effective defense should displace the original semantic features so that the resulting latent representation aligns more closely with a deceptive target than with the clean identity. However, \cref{fig:latent_similarity} reveals that protected features consistently maintain a disproportionately high similarity to the original samples, showing minimal correlation with the intended deceptive targets. This phenomenon indicates that current methods merely overlay additive noise onto the images rather than fundamentally redirecting or corrupting the underlying semantic anchor. Consequently, the original identity remains the dominant signal in the latent space, allowing the generative restoration mechanism of the LDM to effectively ignore the defensive signals during the denoising process.

\begin{observationbox}
\textit{\textbf{Observation 2:} Diffusion models can naturally separate distinct representations from mixed features.}
\end{observationbox}

We further evaluate whether LDMs can decompose protected latent representations into their constituent parts: the clean original identity and the additive perturbation. To verify this separability, we apply DDIM inversion~\cite{ddim} to protected latents and perform denoising guided by neutral prompts (e.g., ``image'') to minimize semantic leakage. Quantitative evaluations across 24 defenses (\cref{fig:reconstruction_quality}) reveal high CLIP~\cite{clip} and DINO~\cite{dino} similarities between the reconstructed components and their ground truths. We provide qualitative evidence of this separation in \cref{sec:appendix_reconstruction}. These results demonstrate the model's capacity to isolate and recover original semantics from superficial protective overlays, explaining the re-emergence of protected identities.

\begin{figure}[tbp!]
  \centering
  \begin{minipage}[c]{0.05\columnwidth}
    \mbox{}
  \end{minipage}
  \includegraphics[width=0.7\linewidth]{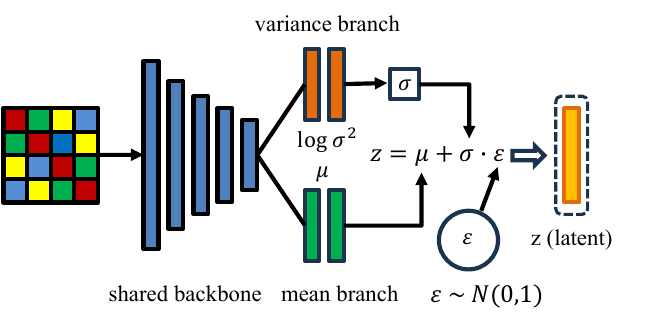}
  \caption{Dual-branch structure of the VAE encoder.}
  \label{fig:vae_branches}
\end{figure}
\subsection{Degradation of Defensive Signals}
\label{sec:limitations_signal}

\begin{observationbox}
\textit{\textbf{Observation 3:} The VAE mean branch structurally attenuates defensive perturbations during latent mapping.}
\end{observationbox}

In addition to semantic persistence, we investigate how defensive perturbations survive through the LDM pipeline. As illustrated in \cref{fig:vae_branches}, the latent representation is jointly determined by the mean ($\mu$) and variance ($\sigma$) branches of the VAE encoder. We evaluate the sensitivity of these branches by computing their Lipschitz constants ($K$) to quantify the survival of defensive perturbations passing through the encoder bottleneck. Since semantic-steering defenses aim to alter deterministic identity features, they fundamentally rely on manipulating the VAE's mean ($\mu$) branch. However, results in \cref{tab:vae_lipschitz} reveal a structural bias across five mainstream VAEs. Specifically, the mean branch exhibits a contractive mapping, effectively attenuating these steering signals. In contrast, the variance branch remains approximately eight times more sensitive to input changes. This disparity exposes a fundamental bottleneck for defenses that rely primarily on the manipulation of the mean latent vector.

\begin{figure}[tbp!]
  \centering
  \includegraphics[width=0.645\linewidth]{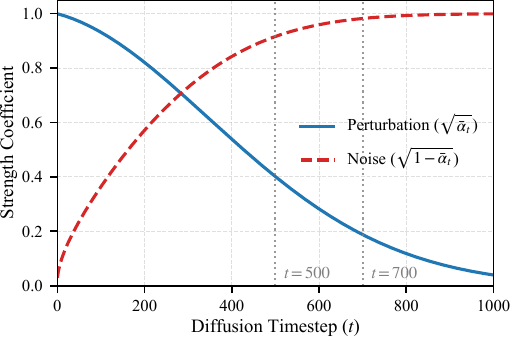}
  \caption{Decay of protective signals during Stable Diffusion v1.5~\cite{sdv1.5} forward diffusion process.}
  \label{fig:signal_decay}
\end{figure}

Following the VAE bottleneck, we further examine the impact of systematic noise injection on the surviving protective signals. In LDM-based mimicry, the perturbed latent $z_0 + \delta_z$ is transformed into a corresponding noisy state $z_t$ according to the forward process:
\begin{equation}
\label{eq:diffusion_forward}
z_t = \sqrt{\bar{\alpha}_t} (z_0 + \delta_{z}) + \sqrt{1 - \bar{\alpha}_t} \epsilon, \quad \epsilon \sim \mathcal{N}(0, \mathbf{I}),
\end{equation}
where the scaling coefficient $\sqrt{\bar{\alpha}_t}$ is governed by the LDM noise schedule. As the timestep $t$ increases to facilitate substantial content modification, $\sqrt{\bar{\alpha}_t}$ decreases monotonically according to the schedule, such as the scaled-linear schedule in Stable Diffusion v1.5~\cite{sdv1.5}. As shown in \cref{fig:signal_decay}, the perturbation weakens because the residual signal decays while noise becomes dominant. For example, at $t=700$, which is a standard setting for robust mimicry tasks, the protective signal is attenuated to about 0.2 of its initial magnitude. This indicates that the noise injection inherent to the editing pipeline effectively washes out surviving defensive signals, making steering-based defenses fragile in practical attacks.

\begin{observationbox}
\textit{\textbf{Observation 4:} Initial noise in the editing process further degrades the remaining protective signals.}
\end{observationbox}

\begin{table}[tbp!]
    \centering
    \caption{Lipschitz constants of five VAE encoders.}
    
    \label{tab:vae_lipschitz}
    \resizebox{0.8\linewidth}{!}{
        \begin{tabular}{lcc|c} 
            \toprule
            VAE Encoder & $K_\mu$ & $K_\sigma$ & Ratio ($K_\sigma / K_\mu$) \\
            \midrule
            SDv1-VAE~\cite{sdv1.5}          & 0.6747 & 5.5859   & 8.2791    \\
            SDv2-D16-VAE~\cite{sdv2.1}      & 0.8759 & 12.7562  & 14.5635   \\
            SDv3-VAE~\cite{sdv3}            & 0.6334 & 6.1018   & 9.6334    \\
            SDXL-VAE~\cite{sdxl}            & 0.8590 & 3.9507   & 4.5992    \\
            SDXL-EQ-VAE~\cite{eqvae}        & 0.8708 & 3.9087   & 4.4886    \\
            \midrule
            AVG                             & 0.7828 & 6.4607   & 8.2533    \\
            \bottomrule
        \end{tabular}
    }
\end{table}

These observations expose that the semantic-steering paradigm is fundamentally vulnerable due to its reliance on external deceptive perturbations. Being extrinsic to the generative process, these signals are inevitably identified as additive noise and purified by the model's restoration mechanisms, including VAE structural compression and diffusion stochasticity, rather than preserved as stable semantic changes. Consequently, the efficacy of such defenses is inherently upper-bounded by the model's restoration capability, rendering them structurally insufficient for robust protection.
\begin{figure*}[t]
    \centering
    \includegraphics[width=0.9\textwidth]{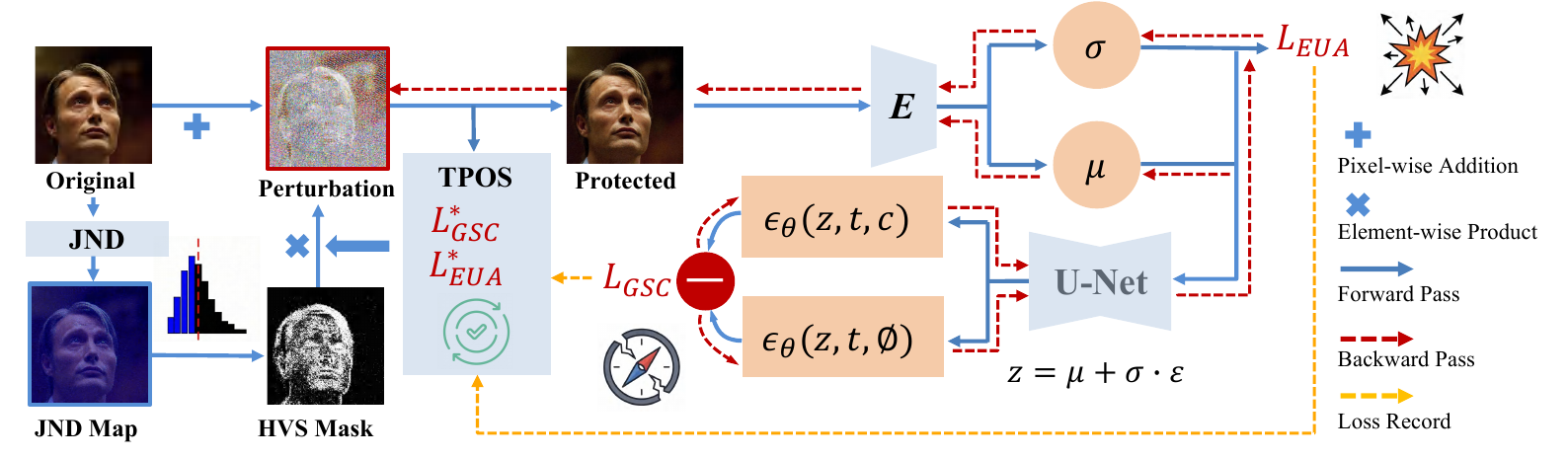}
    \caption{Overview of the proposed \sys framework. The pipeline jointly optimizes the protective perturbation by maximizing encoding uncertainty ($L_{EUA}$) during VAE encoding and counteracting guidance signals ($L_{GSC}$) during U-Net denoising.}
    \label{fig:void_overview}
\end{figure*}

\section{Design of {\sys}}
\label{sec:method}

\subsection{Defense Overview}

Our analysis in \cref{sec:fragility_analysis} establishes that the LDM's restoration priors act as a robust semantic filter, effectively stripping deceptive perturbations introduced by semantic-steering defenses as additive noise. Consequently, a robust defense cannot rely on extrinsic confrontation, as the model is trained to rectify such deviations.

To overcome this, we exploit a critical structural vulnerability. Specifically, while the model effectively purifies external noise, it is strictly compelled to preserve its own intrinsic probabilistic errors to maintain the continuity of the latent manifold. Driven by this insight, we propose a paradigm shift to semantic-corruption. Instead of fighting the model's restoration mechanism, \sys weaponizes this necessity, amplifying internal deviations to trigger a cascading collapse of the generative trajectory.

\sys operationalizes this strategy by targeting two distinct forms of system uncertainty: the encoding deviation inherent in VAE sampling and the stochastic randomness injected throughout the diffusion chain. To exploit the former, we introduce encoding uncertainty amplification (EUA), which penetrates the VAE bottleneck by escalating latent encoding errors into irreducible stochasticity. This ensures that the initial semantic structure is shattered at the very source of the generation process. Simultaneously, to manipulate the latter, we employ guidance signal counteraction (GSC), which leverages diffusion stochasticity to collapse the Classifier-Free Guidance~\cite{cfg} vector. By neutralizing the steering force required for model restoration, GSC prevents the LDM from rectifying the chaos seeded by EUA.

Realizing this paradigm, however, requires reconciling aggressive corruption with high visual utility. We resolve this tension via HVS-guided perturbation masking (HGPM), which restricts artifacts to human-imperceptible regions. To stabilize optimization under these strict spatial constraints, we further integrate timestep-partitioned optimal selection (TPOS). As illustrated in \cref{fig:void_overview}, \sys synthesizes these components to achieve a robust equilibrium between unprecedented mimicry prevention and superior image fidelity.

\subsection{Encoding Uncertainty Amplification}

To initiate intrinsic corruption at the source of the diffusion chain, we first target the VAE bottleneck. Standard LDMs employ a VAE~\cite{vae} to map images into a latent space distribution $\mathcal{N}(\mu,\sigma^2)$. Structurally, the encoder $\mathcal{E}$ consists of two parallel branches: a mean branch (${\mu}$) that captures deterministic semantic features, and a variance branch (${\sigma}$) that represents encoding uncertainty.

Our analysis in \cref{sec:fragility_analysis} reveals that the mean branch acts as a contractive mapping, effectively attenuating external perturbations to restore original semantics. Instead of engaging in a futile struggle against this robust mean branch, \sys pivots to exploit the highly sensitive variance branch. Our Lipschitz evaluation confirms this strategic shift: the variance branch exhibits a sensitivity coefficient ($K_{\sigma} \approx 6.5$) nearly an order of magnitude higher than that of the mean branch ($K_{\mu} \approx 0.8$). By targeting this sensitive component, we can aggressively amplify inherent encoding uncertainty to penetrate the VAE bottleneck without being suppressed.

In the latent diffusion framework, this stochasticity is not mere noise but a structural necessity defined by the reparameterization trick: $\mathbf{z} = {\mu} + {\sigma} \odot \mathbf{n}$. Unlike deterministic features, the variance ${\sigma}$ cannot be filtered by the model's restoration mechanism without destroying the continuity of the latent manifold. Consequently, maximizing this variance fundamentally alters the topology of the latent representation, creating a probabilistic shield that forces the model to interpret the perturbation as inherent ambiguity within the source data. We operationalize this by formulating the EUA loss to maximize the aggregate predicted uncertainty:

\begin{equation}
L_{\text{EUA}} = \sum_{i} \sigma_{\mathcal{E}, i}(\mathbf{x}+{\delta}). 
\label{eq:eua_loss}
\end{equation}

By forcing the encoder to map the image into a region of maximal uncertainty, this objective converts the defense into irreducible stochasticity. This seeds the diffusion pipeline with high-entropy chaos, ensuring that subsequent mimicry collapses into meaningless noise.

\subsection{Guidance Signal Counteraction}
\begin{figure}[t]
\centering
\includegraphics[width=1.0\columnwidth]{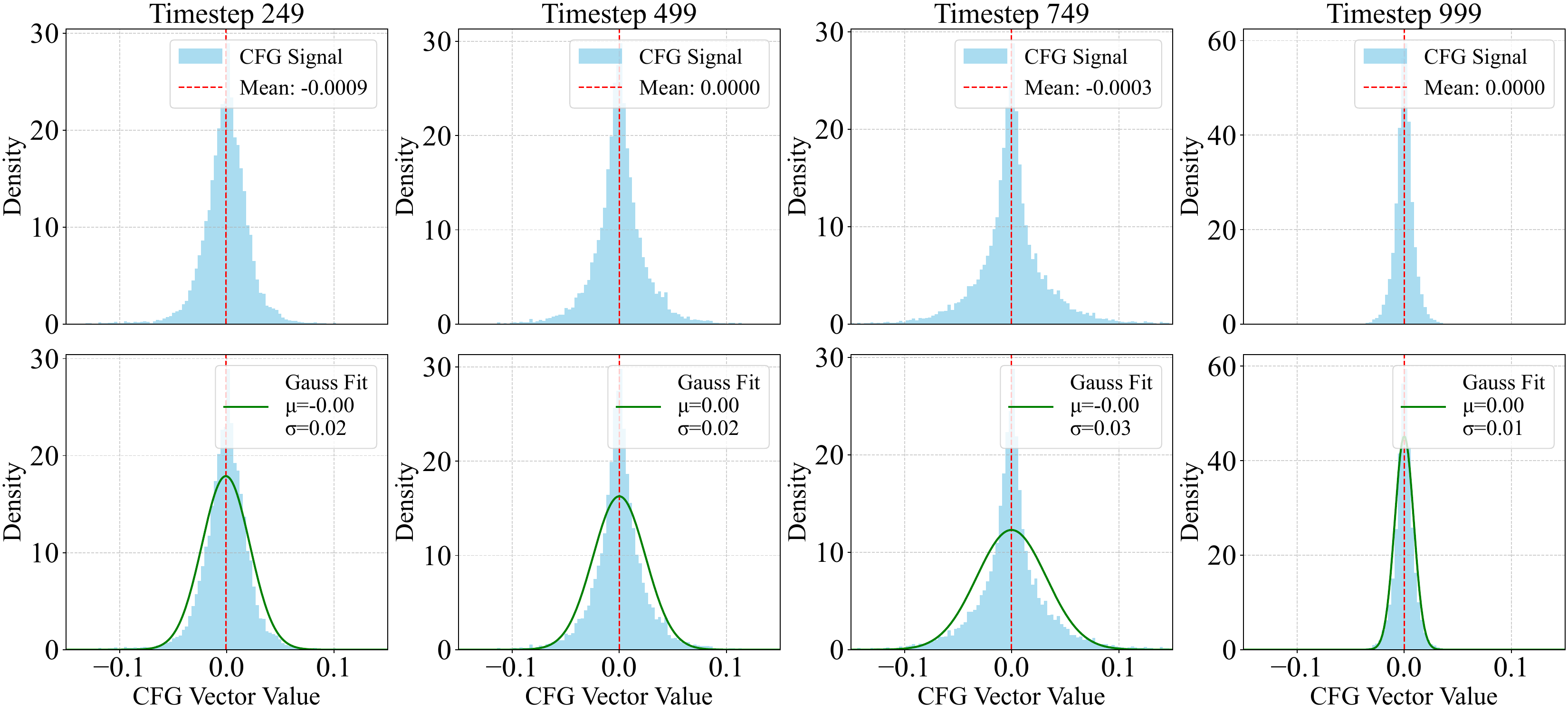}
\caption{Distribution analysis of the CFG vector across different timestep ranges $t$. The distributions exhibit near-zero mean and small variance, resembling Gaussian distributions.}
\label{fig:cfg_distribution}
\end{figure}

While EUA initiates latent chaos at the source, the LDM inherently employs Classifier-Free Guidance (CFG)~\cite{cfg} to rectify such deviations during the denoising process. CFG functions as the primary restoration engine by extrapolating a guidance vector $\Delta {\epsilon}_t$, which represents the semantic discrepancy between conditional and unconditional noise predictions. This vector acts as the steering force that aligns the generation trajectory with the target prompt $c$ as follows:

\begin{equation}
\Delta {\epsilon}_t({\delta}) = \epsilon_\theta(\mathbf{z}_t, t, c) - \epsilon_\theta(\mathbf{z}_t, t, \emptyset).
\label{eq:guidance_vector}
\end{equation}

To prevent the model from recovering protected features via this mechanism, guidance signal counteraction (GSC) aims to fundamentally collapse the LDM's restoration capability. Our key insight is that the effectiveness of CFG depends not merely on the magnitude of $\Delta {\epsilon}_t$ but on its directional variance, which provides the necessary gradient information for semantic alignment. As visualized in \cref{fig:cfg_distribution}, the values of $\Delta {\epsilon}_t$ follow an approximately Gaussian distribution with near-zero mean. Leveraging this stochastic nature, GSC neutralizes the restoration force by collapsing the signal distribution rather than simply suppressing its magnitude. We achieve this by minimizing the standard deviation of the guidance vector to compress the signal into a non-informative state:

\begin{equation}
L_{\text{GSC}}({\delta}) = \mathbb{E}_{t, {\epsilon}}\Big[ \sigma\left( \Delta {\epsilon}_t({\delta}) \right) \Big].
\label{eq:gsc_loss}
\end{equation}

Minimizing this objective effectively decouples the generation from its semantic anchor. By inducing a non-informative guidance signal, we ensure that the model receives no consistent directional information to recover original identities, thereby forcing the denoising process to propagate the chaos seeded by EUA throughout the entire diffusion chain.

To handle practical black-box scenarios where the adversary's attack prompt $c$ is unknown, we employ a shadow condition strategy. We use blind captioning (e.g., BLIP~\cite{blip}) to generate a proxy prompt $\hat{c}$ that captures the inherent visual attributes of the reference image. Because unauthorized mimicry fundamentally targets these visual features, $\hat{c}$ represents a critical semantic intersection with potential attack prompts. This approach allows \sys to disrupt the fundamental visual features targeted by mimicry, ensuring the defense remains effective even when attack prompts are unknown.

\subsection{Joint Optimization Framework}
\label{sec:optimization}
\SetKwInOut{Input}{Input}
\SetKwInOut{Output}{Output}
\SetKwProg{For}{for}{}{end}

\begin{algorithm}[t]
    \caption{\sys Protection Framework}
    \label{algo:medu}
    \DontPrintSemicolon
    \SetAlgoLined

    \Input{Original image $\mathbf{x}$ and proxy prompt $\hat{c}$, VAE Encoder $E$ and LDM $\epsilon_\theta$, Perturbation budget $\alpha$, Step size $\eta$, Total iterations $N$, JND masking ratio $\rho$, TPOS partition number $K$, Loss weight $\lambda$.}
    \Output{Protected image $\mathbf{x}'$}

    Initialize $\delta \leftarrow 0, \delta^* \leftarrow 0, L_{\text{best}} \in \mathbb{R}^K \leftarrow -\infty$\;
    
    $M_{\text{HVS}} \leftarrow \mathbb{I}(\text{TD-JND}(\mathbf{x}) \in \text{Top}_{\rho})$\;

    \For{$n = 1$ \textbf{to} $N$}{
        $t \sim U[1, T],k \leftarrow \lfloor t \cdot K / T \rfloor$\;

        $z'_0 \leftarrow E(\mathbf{x} + \delta), \epsilon \sim N(0, \mathbf{I})$\;

        \tcp{EUA}
        $L_{\text{EUA}} \leftarrow \sum \sigma(z'_0)$\; 

        $z'_t \leftarrow \sqrt{\bar{\alpha}_t} z'_0 + \sqrt{1 - \bar{\alpha}_t} \epsilon$\;
                
        \tcp{GSC}
        $L_{\text{GSC}} \leftarrow \sigma(\epsilon_\theta(z'_t, t, \hat{c}) - \epsilon_\theta(z'_t, t, \emptyset))$\;

        $L \leftarrow \lambda L_{\text{EUA}} - L_{\text{GSC}}$\;

        \tcp{HGPM}
        $\delta \leftarrow \text{Proj}_{\ell_\infty \le \alpha}\left(\delta + \eta \cdot \text{sign}(\nabla_{\delta} L)\right) \odot M_{\text{HVS}}$\;

        \tcp{TPOS}
        \If{$L > L_{\text{best}}[k]$}{
             $L_{\text{best}}[k] \leftarrow L, \delta^* \leftarrow \delta$\;
        }
    }
    
    \Return $\mathbf{x}' \leftarrow \mathbf{x} + \delta^*$\;
\end{algorithm}

To operationalize semantic corruption, we integrate the EUA and GSC into a joint optimization objective:
\begin{equation}
\label{eq:combined_loss_short}
\max_{\delta} J(\delta) = \lambda L_{\text{EUA}}(\mathbf{x}+\delta) - L_{\text{GSC}}(\mathbf{x}+\delta),
\end{equation}
where $\lambda$ balances the trade-off between latent chaos and guidance suppression ($\lambda = 10$ by default). To preserve visual utility, we frame this as a constrained optimization problem using two mechanisms to ensure stealthiness and stability.

\noindent\textbf{HVS-Guided Perturbation Masking.}
Standard defenses typically impose a spatially uniform $\ell_\infty$-norm budget, which disregards local texture sensitivity and introduces visible artifacts in smooth regions. To rectify this, \sys employs HVS-guided perturbation masking (HGPM) to adaptively allocate the security budget. Leveraging the Top-Down Just Noticeable Difference (TD-JND)~\cite{tdjnd} model, we compute a redundancy map that quantifies the perceptual tolerance of each pixel. We then construct a binary mask $M_{\text{HVS}}$ by retaining only the top $\rho\%$ regions with the highest JND values, effectively confining the perturbation to a texture-aware subspace:
\begin{equation}
\delta_{k+1} \leftarrow \text{Proj}_{\ell_\infty \le \alpha} (\delta_{k+1} \odot M_{\text{HVS}}).
\end{equation}
This approach ensures that aggressive corruption signals are embedded exclusively within human-insensitive regions, maintaining high visual fidelity.

\noindent\textbf{Timestep-Partitioned Optimal Selection.}
While HGPM ensures stealthiness, strictly enforcing spatial constraints can limit the search space and cause optimization oscillation. Furthermore, because timesteps $t$ are randomly sampled in \cref{eq:combined_loss_short}, a high loss value often reflects sampling bias rather than actual perturbation quality. To recover performance, we introduce timestep-partitioned optimal selection (TPOS). We divide diffusion timelines into $K$ disjoint intervals and maintain separate optima for each partition. This allows the optimization to decouple sampling variance and independently recover the optimal perturbation for each timestep partition.

\noindent\textbf{Overall Algorithm.}
The complete execution flow is detailed in \cref{algo:medu}. By synthesizing static perceptual constraints from HGPM with the dynamic joint optimization of EUA and GSC, \sys achieves a robust equilibrium between unprecedented mimicry prevention and superior image fidelity.
\section{Evaluation}
\label{sec:evaluation}

\subsection{Evaluation Setup}
\label{sec:evaluation_setup}

\noindent\textbf{Datasets.} We evaluate unauthorized mimicry on 5 datasets across 3 tasks: \emph{identity mimicry} on CelebA-HQ~\cite{celeba-hq} and VGGFace2~\cite{vggface2}, \emph{object mimicry} on TI-Dataset~\cite{ti} and DB-Dataset~\cite{dreambooth}, and \emph{style mimicry} on WikiArt~\cite{wikiart}.

\noindent\textbf{Models.}
We evaluate 4 LDMs: Stable Diffusion (SD) v1.5~\cite{sdv1.5}, SD v2.1~\cite{sdv2.1}, DreamShaper~\cite{dreamshaper}, and Latent Consistency Models~\cite{lcm}. We also evaluate two transformer-based LDMs, FLUX.1 Kontext~\cite{flux1kontext} and FLUX.2 Klein~\cite{flux2klein}. Following prior works~\cite{pretender,gapdiff}, we use SD v1.5~\cite{sdv1.5} as the surrogate.

\noindent\textbf{Mimicry Attacks.}
We evaluate 10 mimicry attacks spanning \emph{training-based personalization} and \emph{inference-based editing}. The former includes Fine-tuning (FT)~\cite{ldm}, DreamBooth (DB)~\cite{dreambooth}, LoRA (LR)~\cite{lora}, SVDiff (SV)~\cite{svd}, and Textual Inversion (TI)~\cite{ti}; the latter includes SDEdit (SE)~\cite{sdedit}, Inpainting (IP)~\cite{ldm}, ControlNet (CN)~\cite{controlnet}, DDIM Inversion (DI)~\cite{vddimiedit}, and DiffEdit (DF)~\cite{diffedit}.

\noindent\textbf{Mimicry Defenses.}
We compare against 24 defenses, including 12 evasion-based methods, LDMR~\cite{ldmr}, AdvDM~\cite{advdm}, Mist~\cite{mist}, PhotoGuard~\cite{photoguard}, Glaze~\cite{glaze}, Smooth~\cite{smooth}, SDST~\cite{sdst}, PID~\cite{pid}, DiffGuard~\cite{diffusionguard}, Nightshade~\cite{nightshade}, GAPDiff~\cite{gapdiff}, and LDS~\cite{lds}, and 12 poisoning-based methods, EUDP~\cite{eudp}, AntiDB~\cite{antidb}, ACE~\cite{ace}, SimAC~\cite{simac}, Metacloak~\cite{metacloak}, DisDiff~\cite{disdiff}, CAAT~\cite{caat}, PAP~\cite{pap}, AntiDiff~\cite{antidiffusion}, GoodAC~\cite{goodac}, Pretender~\cite{pretender}, and HAAD~\cite{haad}.

\noindent\textbf{Metrics.}
We evaluate two dimensions: \emph{Effectiveness}, measured by FID~\cite{fid}, CLIP-S~\cite{clipscore}, and CLIP-I~\cite{clip}, and \emph{Utility}, using visual quality metrics, including PSNR~\cite{psnr}, SSIM~\cite{ssim}, FSIM~\cite{fsim}, VIF~\cite{vif}, GMSD~\cite{gmsd}, LPIPS~\cite{lpips}, DeepIQA~\cite{deepiqa}, PieAPP~\cite{pieapp}, DISTS~\cite{dists}, and ContentLoss~\cite{content}.

\noindent\textbf{Implementation.}
All defenses use the same perturbation budget, $\epsilon=8/255$ under the $\ell_\infty$-norm. Detailed configurations are provided in \cref{sec:implement_details}.

\subsection{Validation of Defense Mechanisms}
\label{sec:rq1_verification}
\begin{figure}[tbp!]
    \captionsetup[subfigure]{labelformat=empty}
    \centering
    \begin{subfigure}[b]{0.3\linewidth}
        \centering
        \includegraphics[width=\linewidth, keepaspectratio]{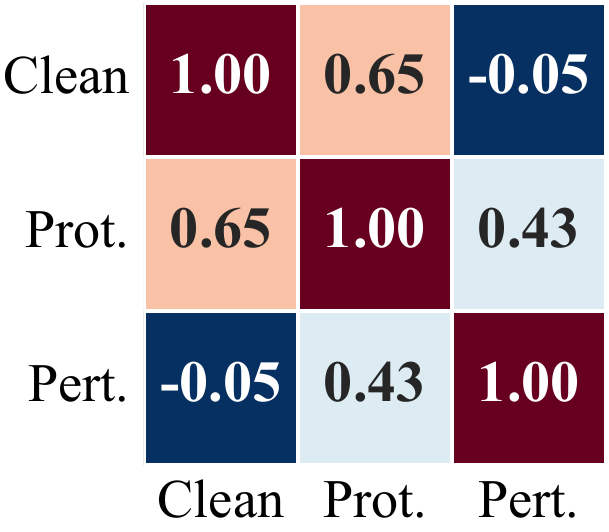}
        \label{fig:adversarial1}
    \end{subfigure}
    \hspace{10pt}
    \begin{subfigure}[b]{0.3\linewidth}
        \centering
        \includegraphics[width=\linewidth, keepaspectratio]{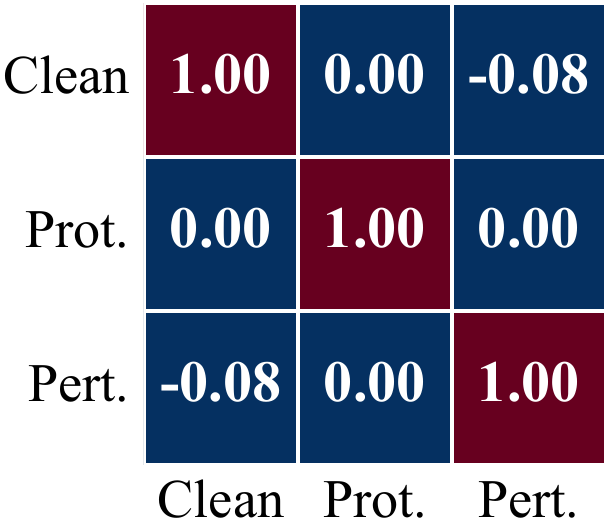}
        \label{fig:poisoning1}
    \end{subfigure}
    \hspace{10pt}
    \raisebox{1.5em}{\includegraphics[height=2.15cm]{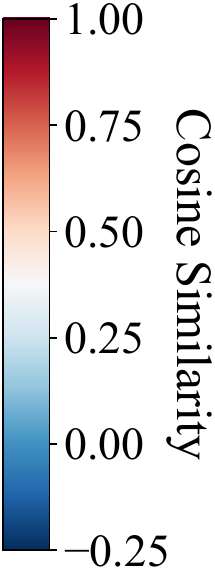}}
    \vspace{-0.4cm}
    \caption{Latent semantic erasure analysis.}
    \label{fig:semantic_erasure}
\end{figure}
\begin{table}[tbp!]
  \centering
  \caption{Resilience against intense diffusion. Comparison of protection effectiveness (FID $\uparrow$) across varying timesteps ($t$).}
  \label{tab:diffusion_resilience}
  \setlength{\tabcolsep}{16pt}
  \renewcommand{\arraystretch}{0.6}
  \resizebox{\linewidth}{!}{
  \begin{tabular}{lccccc}
    \toprule
    \multirow{2}{*}{\textbf{Defense}} & \multicolumn{5}{c}{\textbf{Diffusion Timestep ($t$)}} \\
    \cmidrule(lr){2-6}
    & \textbf{100} & \textbf{300} & \textbf{500} & \textbf{700} & \textbf{900} \\
    \midrule
    Smooth & 127 & 148 & 158 & 124 & 7 \\
    ACE    & 168 & 142 & 57 & 21 & 10 \\
    \midrule
    \rowcolor{Gray}
    {\sys (Ours)} & \textbf{286} & \textbf{293} & \textbf{296} & \textbf{301} & \textbf{306} \\
    \bottomrule
  \end{tabular}
  }
\end{table}

We first validate whether semantic corruption achieves latent semantic erasure and persistence under diffusion-induced signal degradation. For latent erasure, we evaluate VAE~\cite{vae} latent identity correlation. As shown in \cref{fig:semantic_erasure}, steering-based defenses preserve correlation after VAE compression (Avg. Sim $\approx$ 0.65), whereas \sys reduces it to nearly zero (0.00), decoupling protected representations from semantics. \cref{sec:appendix_specificity} further shows that this corruption is concept-specific rather than global pipeline failure.

We then evaluate whether the protection remains stable under diffusion noise injection. As shown in \cref{tab:diffusion_resilience}, Smooth~\cite{smooth} and ACE~\cite{ace} collapse at large timesteps (FID $\approx$ 10), while \sys maintains strong disruption (Avg. FID $\approx$ 296) and even strengthens as the timestep increases. Based on this result, we adopt adversary-favored settings in subsequent experiments. For inference-based image editing, we use a large timestep of $t=700$ to increase noise injection and editing strength, as in prior works~\cite{ldmr,smooth}. For training-based personalization, we use $t=300$ as the starting point of the diffusion training range. These high-step settings impose stronger signal degradation and provide a rigorous stress test for defense persistence.

\begin{table*}[ht]
\centering

\caption{Quantitative evaluation of defense efficacy. Results are averaged across 5 datasets (TI-Dataset~\cite{ti}, DB-Dataset~\cite{dreambooth}, CelebA-HQ~\cite{celeba-hq}, VGGFace2~\cite{vggface2} and WikiArt~\cite{wikiart}) and 6 mimicry methods: FT~\cite{ldm}, SV~\cite{svd}, LR~\cite{lora}, SE~\cite{sdedit}, IP~\cite{ldm} and CN~\cite{controlnet}). Arrows indicate the direction of better protection performance. \textbf{Bold} and \underline{underlined} values denote the best and second-best results, respectively. Detailed breakdowns per mimicry method and dataset are provided in \crefrange{tab:infer_fid}{tab:train_clip_s}.}

\label{tab:main_effectiveness}
\setlength{\tabcolsep}{5.0pt}
\renewcommand{\arraystretch}{0.8}
\resizebox{\textwidth}{!}{
\begin{tabular}{cl ccc ccc ccc}
\toprule
\multirow{2}{*}{\textbf{Type}} & \multirow{2}{*}{\textbf{Defense}} & \multicolumn{3}{c}{\textbf{Training-based mimicry}} & \multicolumn{3}{c}{\textbf{Inference-based mimicry}} & \multicolumn{3}{c}{\textbf{Average}} \\
\cmidrule(lr){3-5} \cmidrule(lr){6-8} \cmidrule(lr){9-11}
 & & FID~\cite{fid} $\uparrow$ & CLIP-S~\cite{clipscore} $\downarrow$ & CLIP-I~\cite{clip} $\downarrow$ & FID~\cite{fid} $\uparrow$ & CLIP-S~\cite{clipscore} $\downarrow$ & CLIP-I~\cite{clip} $\downarrow$ & FID~\cite{fid} $\uparrow$ & CLIP-S~\cite{clipscore} $\downarrow$ & CLIP-I~\cite{clip} $\downarrow$ \\
\midrule
\multirow{11}{*}{\rotatebox{90}{\textbf{Poisoning}}} & ACE~\cite{ace} & \underline{146.22} & \underline{24.52} & \underline{0.69} & 47.87 & 27.04 & 0.80 & 97.04 & \underline{25.78} & \underline{0.74} \\
 & AntiDB~\cite{antidb} & 69.32 & 28.49 & 0.77 & 65.86 & 26.24 & 0.84 & 67.59 & 27.36 & 0.81 \\
 & AntiDiff~\cite{antidiffusion} & 55.37 & 28.97 & 0.79 & 48.55 & 26.46 & 0.86 & 51.96 & 27.71 & 0.83 \\
 & CAAT~\cite{caat} & 67.90 & 28.73 & 0.77 & 45.40 & 26.74 & 0.86 & 56.65 & 27.74 & 0.81 \\
 & DisDiff~\cite{disdiff} & 71.97 & 28.75 & 0.77 & 28.11 & 26.79 & 0.88 & 50.04 & 27.77 & 0.82 \\
 & EUDP~\cite{eudp} & 76.59 & 28.62 & 0.76 & 29.69 & 27.12 & 0.87 & 53.14 & 27.87 & 0.82 \\
 & GoodAC~\cite{goodac} & 70.63 & 28.38 & 0.77 & 74.73 & 26.11 & 0.84 & 72.68 & 27.25 & 0.80 \\
 & MetaCloak~\cite{metacloak} & 44.03 & 29.09 & 0.81 & 38.48 & 26.65 & 0.87 & 41.26 & 27.87 & 0.84 \\
 & PAP~\cite{pap} & 72.26 & 28.36 & 0.76 & 87.86 & 26.33 & 0.83 & 80.06 & 27.35 & 0.80 \\
 & Pretender~\cite{pretender} & 56.95 & 28.98 & 0.79 & 41.79 & 26.79 & 0.86 & 49.37 & 27.88 & 0.82 \\
 & SimAC~\cite{simac} & 44.75 & 29.19 & 0.82 & 22.67 & 26.60 & 0.89 & 33.71 & 27.90 & 0.85 \\
 & HAAD~\cite{haad} & 70.95 & 28.56 & 0.76 & 78.26 & 26.89 & 0.83 & 74.60 & 27.72 & 0.80 \\
\midrule
\textbf{-} & RandomNoise & 22.15 & 29.16 & 0.88 & 5.64 & 26.55 & 0.93 & 13.90 & 27.85 & 0.90 \\
\midrule
\multirow{12}{*}{\rotatebox{90}{\textbf{Evasion}}} & AdvDM~\cite{advdm} & 53.27 & 29.18 & 0.80 & 39.85 & 27.07 & 0.85 & 46.56 & 28.13 & 0.82 \\
 & DiffGuard~\cite{diffusionguard} & 31.29 & 29.65 & 0.84 & 14.85 & 26.56 & 0.89 & 23.07 & 28.11 & 0.87 \\
 & GAPDiff~\cite{gapdiff} & 103.96 & 26.47 & 0.71 & 33.48 & 26.85 & 0.86 & 68.72 & 26.66 & 0.78 \\
 & Glaze~\cite{glaze} & 138.80 & 24.57 & 0.69 & 48.96 & 27.31 & 0.81 & 93.88 & 25.94 & 0.75 \\
 & LDMR~\cite{ldmr} & 46.70 & 29.03 & 0.79 & 43.66 & 26.45 & 0.82 & 45.18 & 27.74 & 0.81 \\
 & MIST~\cite{mist} & 143.17 & 24.53 & 0.69 & 48.32 & 27.39 & 0.81 & 95.74 & 25.96 & 0.75 \\
 & NightShade~\cite{nightshade} & 140.97 & 24.58 & 0.69 & 49.98 & 27.09 & 0.80 & 95.48 & 25.83 & 0.75 \\
 & PID~\cite{pid} & 34.14 & 29.26 & 0.82 & 27.68 & 27.28 & 0.84 & 30.91 & 28.27 & 0.83 \\
 & PhotoGuard~\cite{photoguard} & 32.18 & 29.20 & 0.83 & 23.19 & 27.01 & 0.85 & 27.69 & 28.11 & 0.84 \\
 & SDST~\cite{sdst} & 48.09 & 28.20 & 0.79 & 35.61 & 27.39 & 0.82 & 41.85 & 27.79 & 0.81 \\
 & Smooth~\cite{smooth} & 65.96 & 28.68 & 0.78 & \underline{160.92} & \underline{25.50} & \underline{0.79} & \underline{113.44} & 27.09 & 0.78 \\
 & LDS~\cite{lds} & 23.55 & 28.77 & 0.84 & 11.42 & 27.21 & 0.91 & 17.48 & 27.99 & 0.87 \\
 \cmidrule{2-11}
 \rowcolor{Gray}
 & {\sys (Ours)} & \textbf{404.94} & \textbf{20.08} & \textbf{0.53} & \textbf{325.80} & \textbf{18.82} & \textbf{0.49} & \textbf{365.37} & \textbf{19.45} & \textbf{0.51} \\
\bottomrule
\end{tabular}
}
\end{table*}
\subsection{Effectiveness against Mimicry Attacks}
\label{sec:rq2_effectiveness}
\begin{figure}[tbp!]
  \centering
  \includegraphics[width=0.9\linewidth]{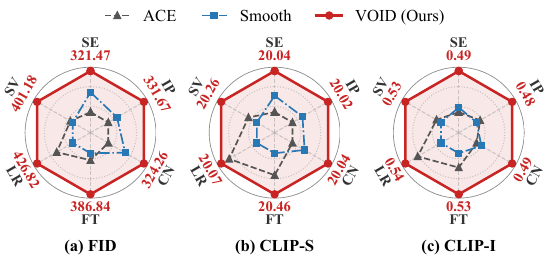}
    \caption{Defense effectiveness across diverse mimicry scenarios. Unlike baselines that exhibit jagged profiles across different attacks, \sys consistently encompasses others, demonstrating comprehensive protection.}
  \label{fig:attack_coverage}
\end{figure}

\noindent\textbf{Quantitative Results.} \cref{tab:main_effectiveness} reports the mimicry defense effectiveness averaged across 5 datasets and 6 mimicry attacks. To benchmark performance against the most capable counterparts, we identify ACE~\cite{ace} and Smooth~\cite{smooth} as the leading representatives of poisoning and evasion defenses, respectively. Despite their relative effectiveness, \sys demonstrates overwhelming superiority, outperforming these baselines by an average factor of \textbf{3.2$\times$} in FID~\cite{fid} (elevating the average disruption level from 113.44 to \textbf{365.37}). Beyond substantial visual degradation, the CLIP-S~\cite{clipscore} metric reveals a fundamental mechanistic distinction: even the strongest baselines retain high semantic alignment (25.83) perilously close to unprotected images, indicating persistent identity leakage, whereas \sys drastically reduces CLIP-S~\cite{clipscore} to \textbf{19.45}. Such semantic persistence in existing defenses explains their specialized yet fragile nature, as the latent identity remains accessible to diverse attack vectors. Furthermore, \cref{fig:attack_coverage} reveals that existing semantic-steering paradigms fail to provide uniform protection across diverse mimicry threats. Specifically, leading defenses such as ACE~\cite{ace} and Smooth~\cite{smooth} exhibit significant performance disparities across different mimicry paradigms, as they often provide effective defense against limited attack vectors while remaining vulnerable to others. \sys overcomes this limitation by maintaining a comprehensive protection coverage, consistently achieving superior mimicry disruption in both training-based model personalization and inference-based image editing scenarios.

\begin{figure*}[htbp!]
    \centering
    \includegraphics[width=\textwidth]{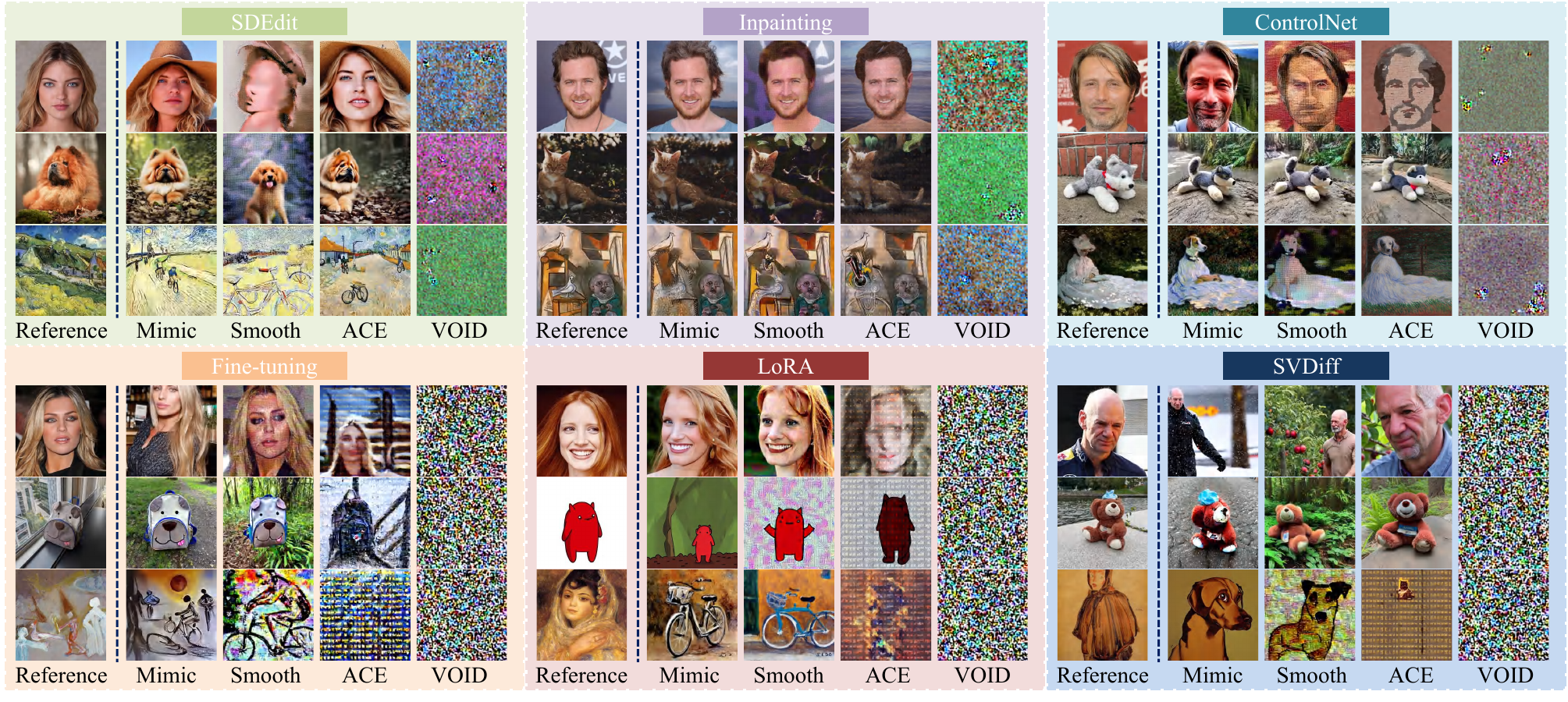}
    \caption{Qualitative comparison of defense efficacy.
    Rows 1-3 correspond to \emph{inference-based image mimicry} (SE~\cite{sdedit}, IP~\cite{ldm}, CN~\cite{controlnet}), and Rows 4-6 correspond to \emph{training-based model personalization} (FT~\cite{ldm}, LoRA~\cite{lora}, SVDiff~\cite{svd}). Columns compare the unprotected baseline (Mimic), the top-performing specialized defenses (Smooth~\cite{smooth} and ACE~\cite{ace}), and our \sys.}
    \label{fig:qualitative_comparison}
\end{figure*}

\noindent\textbf{Qualitative Results.}
Qualitative results in \cref{fig:qualitative_comparison} confirm that although steering-based baselines like ACE~\cite{ace} and Smooth~\cite{smooth} introduce distortion into mimicry results, original features frequently re-emerge. In contrast, \sys induces a robust semantic collapse that consistently renders mimicry outputs as semantically void noise. Visual inspection reveals that these outputs manifest as pure random noise with no perceptible trace of the original identity or style, confirming that \sys prevents the model from reconstructing any meaningful content from the corrupted latents. By corrupting the denoising trajectory to neutralize the model's restoration capability, \sys effectively severs the semantic link between original features and outputs to render mimicry attempts infeasible.

\begin{table*}[t]
\centering
\caption{Imperceptibility evaluation. \sys is compared against the top-4 visual quality baselines among 24 methods.}
\label{tab:imperceptibility}
\setlength{\tabcolsep}{4pt}
\renewcommand{\arraystretch}{1}
\resizebox{\textwidth}{!}{
\begin{tabular}{l ccccc ccccc}
\toprule
\multirow{2}{*}{\textbf{Defense}} & \multicolumn{5}{c}{\textbf{Signal-level Metrics}} & \multicolumn{5}{c}{\textbf{Neural-based Metrics}} \\
\cmidrule(lr){2-6} \cmidrule(lr){7-11}
 & PSNR~\cite{psnr}$\uparrow$ & SSIM~\cite{ssim}$\uparrow$ & FSIM~\cite{fsim}$\uparrow$ & VIF~\cite{vif}$\uparrow$ & GMSD~\cite{gmsd}$\downarrow$ & LPIPS~\cite{lpips}$\downarrow$ & DISTS~\cite{dists}$\downarrow$ & DeepIQA~\cite{deepiqa}$\uparrow$ & PieAPP~\cite{pieapp}$\downarrow$ & ContentLoss~\cite{content}$\downarrow$ \\
\midrule
AntiDiff~\cite{antidiffusion} & 33.708 & \underline{0.870} & 0.965 & 0.598 & 0.048 & 0.162 & 0.199 & -0.026 & 1.330 & 1297.6 \\
DiffGuard~\cite{diffusionguard} & \underline{33.841} & 0.854 & 0.975 & \underline{0.611} & 0.036 & 0.135 & \underline{0.192} & -0.019 & 0.912 & \underline{1247.8} \\
PhotoGuard~\cite{photoguard} & 33.367 & 0.843 & \underline{0.979} & 0.584 & \underline{0.028} & 0.122 & 0.219 & \underline{-0.014} & 0.728 & 1589.5 \\
SDST~\cite{sdst} & 32.903 & 0.823 & 0.977 & 0.534 & 0.028 & \underline{0.116} & 0.291 & -0.017 & \underline{0.716} & 2182.5 \\
\midrule
\rowcolor{Gray}
{\sys (Ours)} & \textbf{34.408} & \textbf{0.888} & \textbf{0.985} & \textbf{0.624} & \textbf{0.019} & \textbf{0.096} & \textbf{0.177} & \textbf{-0.010} & \textbf{0.679} & \textbf{1204.9} \\
\bottomrule
\end{tabular}
}
\end{table*}
\subsection{Practicality for Real-world Deployment}

\begin{figure}[tbp!]
  \centering
  \includegraphics[width=\linewidth]{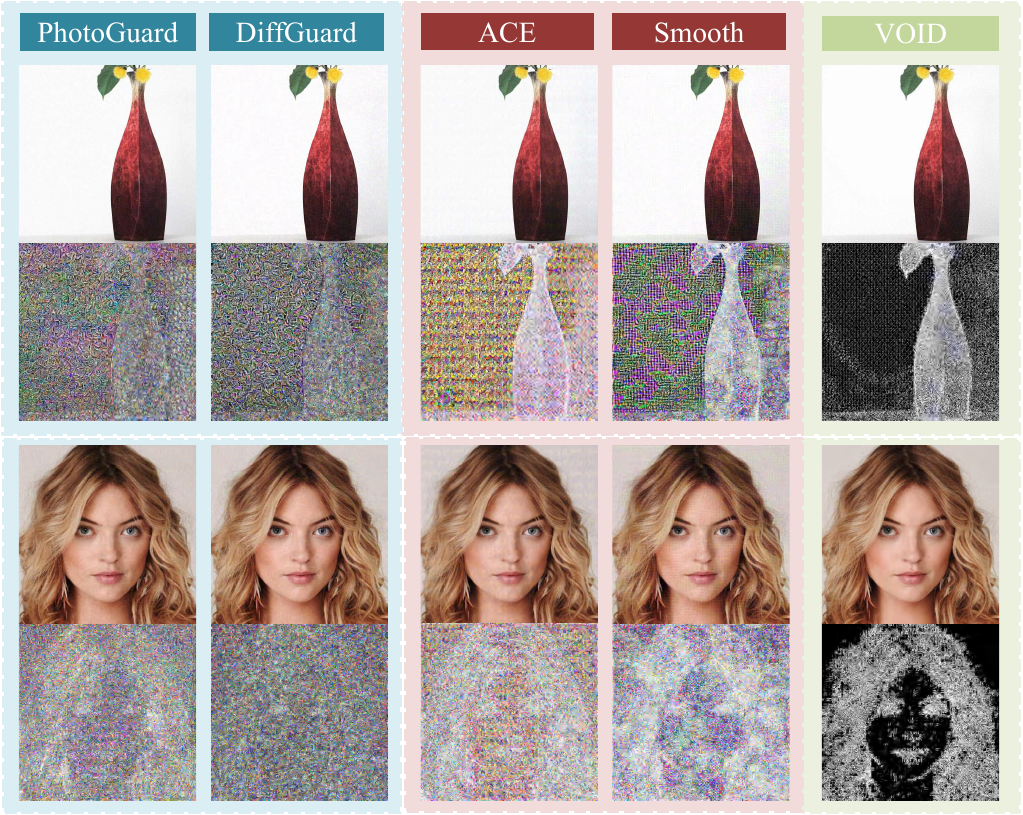}
    \caption{Visual quality comparison. Protected samples and their corresponding perturbation maps are presented, with perturbations magnified by $30\times$ for visibility.}
  \label{fig:visual_quality}
\end{figure}

\noindent\textbf{Visual Quality.} 
We evaluate the visual quality of the protected samples produced by \sys using 10 visual quality metrics. This evaluation encompasses both traditional signal-level measures, such as PSNR~\cite{psnr} and SSIM~\cite{ssim}, and neural-based perceptual models, including LPIPS~\cite{lpips} and DeepIQA~\cite{deepiqa}. To ensure a rigorous assessment of utility, we benchmark \sys against 24 state-of-the-art defenses across diverse image domains. \cref{tab:imperceptibility} compares \sys against the top-4 baselines (see \cref{tab:full_imperceptibility} for full results). As reported, \sys consistently outperforms existing defenses across all evaluated metrics, achieving \textbf{34.408} dB PSNR and \textbf{0.888} SSIM. Notably, \sys is the only method achieving an LPIPS below 0.1 (\textbf{0.096}), confirming high visual quality. While leading baselines such as PhotoGuard~\cite{photoguard} and SDST~\cite{sdst} apply indiscriminate global perturbations, the protective noise from \sys is remarkably sparse (\cref{fig:visual_quality}). This superior image utility stems from our HVS-guided perturbation masking strategy, which leverages the Just Noticeable Difference (JND)~\cite{tdjnd} model to adaptively allocate perturbations to human-insensitive regions.

\begin{table*}[htbp]
    \centering
    \caption{Cross-model transferability evaluation on TI-Dataset~\cite{ti}. {Bold} indicates the best performance among defenses.}
    \label{tab:cross_model_transfer}
    
    \setlength{\tabcolsep}{4pt} 
    \renewcommand{\arraystretch}{0.8}
    
    \resizebox{\textwidth}{!}{
        \begin{tabular}{l l ccc ccc ccc ccc}
        \toprule
        \multirow{2}{*}{{Defense}} & \multirow{2}{*}{{Attack}} & 
        \multicolumn{3}{c}{{SD v1.5}$^\ast$~\cite{sdv1.5}} & 
        \multicolumn{3}{c}{{SD v2.1}~\cite{sdv2.1}} & 
        \multicolumn{3}{c}{{DreamShaper}~\cite{dreamshaper}} & 
        \multicolumn{3}{c}{{LCM}~\cite{lcm}} \\
        
        \cmidrule(lr){3-5} \cmidrule(lr){6-8} \cmidrule(lr){9-11} \cmidrule(lr){12-14} 
        
        & & FID$\uparrow$ & CLIP-S$\downarrow$ & CLIP-I$\downarrow$ & FID$\uparrow$ & CLIP-S$\downarrow$ & CLIP-I$\downarrow$ & FID$\uparrow$ & CLIP-S$\downarrow$ & CLIP-I$\downarrow$ & FID$\uparrow$ & CLIP-S$\downarrow$ & CLIP-I$\downarrow$ \\
        \midrule
        
        \multirow{3}{*}{{Smooth~\cite{smooth}}} 
        & SDEdit~\cite{sdedit} & 121.34 & 30.20 & 0.79 & 108.15 & 27.00 & 0.77 & 63.72 & 33.55 & 0.84 & 86.73 & 29.50 & 0.85 \\
        & Inpainting~\cite{ldm} & 119.15 & 28.86 & 0.85 & 141.21 & 29.27 & 0.84 & 77.87 & 30.63 & 0.87 & 66.05 & 30.35 & 0.90 \\
        & ControlNet~\cite{controlnet} & 110.32 & 30.09 & 0.82 & 103.77 & 27.45 & 0.80 & 55.46 & 32.75 & 0.85 & 113.88 & 30.64 & 0.84 \\
        \midrule

        \multirow{3}{*}{{ACE~\cite{ace}}} 
        & Fine-tuning~\cite{ldm} & 219.69 & 24.26 & 0.64 & 202.78 & 23.33 & 0.64 & 214.95 & 25.15 & 0.66 & 139.47 & 22.26 & 0.76 \\
        & LoRA~\cite{lora} & 208.58 & 23.30 & 0.67 & 253.61 & 20.97 & 0.65 & 197.52 & 23.83 & 0.66 & 221.76 & 19.93 & 0.79 \\
        & SVDiff~\cite{svd} & 84.60 & 29.42 & 0.74 & 51.74 & 28.66 & 0.80 & 103.88 & 29.10 & 0.72 & 96.49 & 21.51 & 0.84 \\
        \midrule

        \rowcolor{Gray}
        & SDEdit~\cite{sdedit} & {299.18} & {19.67} & {0.52} & {301.15} & {19.58} & {0.58} & {310.11} & {19.57} & {0.52} & {304.00} & {19.67} & {0.56} \\
        
        \rowcolor{Gray}
        & Inpainting~\cite{ldm} & {319.47} & {19.68} & {0.55} & {319.13} & {19.54} & {0.54} & {323.30} & {19.57} & {0.56} & {331.84} & {19.69} & {0.57} \\
        
        \rowcolor{Gray}
        & ControlNet~\cite{controlnet} & {300.53} & {19.67} & {0.53} & {300.93} & {19.58} & {0.59} & {305.93} & {19.57} & {0.53} & {316.77} & {19.68} & {0.60} \\
        
        \rowcolor{Gray}
        & Fine-tuning~\cite{ldm} & {416.51} & {19.71} & {0.56} & {368.45} & {18.86} & {0.53} & {377.75} & {20.40} & {0.54} & {455.36} & {19.42} & {0.65} \\
        
        \rowcolor{Gray}
        & LoRA~\cite{lora} & {384.01} & {19.20} & {0.57} & {351.80} & {19.07} & {0.57} & {445.43} & {19.86} & {0.56} & {498.19} & {19.32} & {0.69} \\
        
        \rowcolor{Gray}
        \multirow{-6}{*}{{\sys (Ours)}}
        & SVDiff~\cite{svd} & {343.43} & {19.60} & {0.54} & {365.25} & {18.82} & {0.52} & {334.13} & {19.96} & {0.53} & {414.78} & {19.17} & {0.70} \\
        \bottomrule
        \end{tabular}
    }
\end{table*}
\begin{table}[tbp]
    \centering
    \caption{Practicality comparison across defenses.}
    \label{tab:practicality}
    
    \setlength{\tabcolsep}{3.5pt}
    \renewcommand{\arraystretch}{0.8}
    
    \resizebox{\columnwidth}{!}{
        \begin{tabular}{l c c c c c c}
        \toprule
        \multirow{2}{*}{Defense} 
        & {Time} 
        & {Memory} 
        & \multicolumn{2}{c}{Effectiveness} 
        & \multicolumn{2}{c}{Visual Quality} \\
        \cmidrule(lr){2-2}
        \cmidrule(lr){3-3}
        \cmidrule(lr){4-5}
        \cmidrule(lr){6-7}
        & {min$\downarrow$} 
        & {GB$\downarrow$} 
        & {CLIP-S$\downarrow$} 
        & {CLIP-I$\downarrow$} 
        & {SSIM$\uparrow$} 
        & {LPIPS$\downarrow$} \\
        \midrule
        Smooth~\cite{smooth} & \underline{3.52} & \textbf{8.63} & 27.09 & 0.78 & 0.817 & 0.186 \\
        PhotoGuard~\cite{photoguard} & 14.40 & 33.09 & 28.11 & 0.84 & 0.843 & 0.122 \\
        SDST~\cite{sdst} & \textbf{2.55} & \underline{9.80} & 27.79  & 0.81 & 0.823 & \underline{0.116} \\
        \midrule
        ACE~\cite{ace} & 6.13 & 13.17 & \underline{25.78} & \underline{0.74} & 0.813 & 0.158 \\
        AntiDiff~\cite{antidiffusion} & 7.71 & 16.92 & 27.71 & 0.83 & \underline{0.870} & 0.162 \\
        Pretender~\cite{pretender} & 5.86 & 13.34 & 27.88 & 0.82 & 0.842 & 0.183 \\
        \midrule
        \rowcolor{Gray}
        {\sys (Ours)} & {3.79} & {9.95} & \textbf{19.45} & \textbf{0.51} & \textbf{0.888} & \textbf{0.096} \\
        \bottomrule
        \end{tabular}
    }
\end{table}
\begin{table}[tbp]
    \centering
    \caption{Cross-model transferability on transformer-based latent diffusion models. Comparison on TI-Dataset~\cite{ti}.}
    \label{tab:flux_transfer}
    
    \setlength{\tabcolsep}{3.5pt}
    \renewcommand{\arraystretch}{0.8}
    
    \resizebox{\columnwidth}{!}{
        \begin{tabular}{l c c c c c c}
        \toprule
        \multirow{2}{*}{Defense} 
        & \multicolumn{3}{c}{FLUX.1 Kontext~\cite{flux1kontext}} 
        & \multicolumn{3}{c}{FLUX.2 Klein~\cite{flux2klein}} \\
        \cmidrule(lr){2-4}
        \cmidrule(lr){5-7}
        & FID$\uparrow$ 
        & CLIP-S$\downarrow$ 
        & CLIP-I$\downarrow$ 
        & FID$\uparrow$ 
        & CLIP-S$\downarrow$ 
        & CLIP-I$\downarrow$ \\
        \midrule
        RandomNoise & 7.64 & 28.42 & 0.93 & 11.13 & 28.11 & 0.91 \\
        Smooth~\cite{smooth} & \underline{27.63} & \underline{27.42} & \underline{0.84} & \underline{19.87} & \underline{27.64} & \underline{0.86} \\
        ACE~\cite{ace} & 12.44 & 28.01 & 0.90 & 7.30 & 28.36 & 0.92 \\
        \midrule
        \rowcolor{Gray}
        \sys (Ours) & \textbf{105.27} & \textbf{24.58} & \textbf{0.68} & \textbf{86.61} & \textbf{25.15} & \textbf{0.71} \\
        \bottomrule
        \end{tabular}
    }
\end{table}
\noindent\textbf{Computational Overhead.}
As shown in \cref{tab:practicality}, we compare \sys with six representative defenses covering the main cost profiles of prior work, including Smooth~\cite{smooth} with vanilla evasion, SDST~\cite{sdst} with SDS acceleration, PhotoGuard~\cite{photoguard} with full-pipeline gradients, ACE~\cite{ace} with vanilla poisoning, AntiDiff~\cite{antidiffusion} with prompt tuning, and Pretender~\cite{pretender} with SGBP acceleration. Runtime and memory are measured on a single NVIDIA A100 GPU. \sys requires 3.79 minutes and 9.95 GB memory, which is comparable to lightweight methods such as Smooth~\cite{smooth} and SDST~\cite{sdst} and substantially lower than PhotoGuard~\cite{photoguard}. \sys achieves the best effectiveness and visual quality, with the lowest CLIP-S/CLIP-I scores and the best SSIM/LPIPS results. These results demonstrate that \sys provides a favorable trade-off between deployment efficiency, protection strength, and image utility.

\begin{figure}[tbp!]
  \centering
  \includegraphics[width=\linewidth]{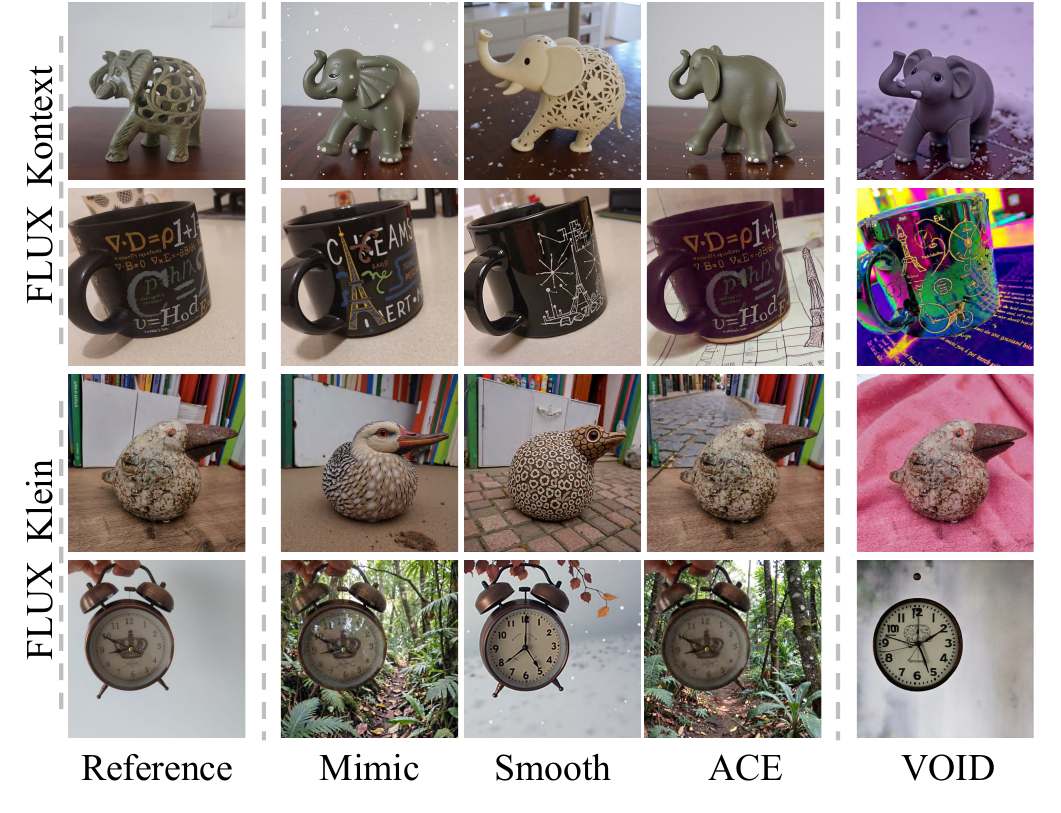}
    \caption{Comparison of defense efficacy on FLUX models.}
  \label{fig:flux_transfer}
\end{figure}
\subsection{Transferability across Diverse Scenarios}
\noindent\textbf{Cross-model transferability.} We assess the transferability of \sys by crafting perturbations on SD v1.5~\cite{sdv1.5} and evaluating them on SD v2.1~\cite{sdv2.1}, DreamShaper~\cite{dreamshaper}, and LCM~\cite{lcm}, which differ in text encoders, community fine-tuning, and distillation mechanisms. \Cref{tab:cross_model_transfer} reports quantitative results. Unlike baselines that suffer significant performance degradation on unseen models, \sys exhibits superior robustness. Notably, while ACE~\cite{ace}'s protection collapses on the distilled LCM~\cite{lcm} during fine-tuning (FID 139.47), \sys maintains a dominant FID of \textbf{455.36} (a \textbf{$3.2\times$} improvement). This shows that \sys targets the intrinsic generative consensus shared across diffusion models rather than model-specific surrogate parameters. We further evaluate \sys on transformer-based LDMs, including FLUX.1 Kontext~\cite{flux1kontext} and FLUX.2 Klein~\cite{flux2klein}. As shown in \cref{tab:flux_transfer}, \sys achieves the best results on both models, with the highest FID and lowest CLIP-S/CLIP-I. \Cref{fig:flux_transfer} shows that FLUX outputs mainly exhibit severe color drift and semantic distortion rather than pure-noise collapse, still making the edited images unusable.

\begin{table*}[htbp]
    \centering
    \caption{Robustness evaluation of \sys on TI-Dataset~\cite{ti} against various countermeasures.}
    \label{tab:countermeasures}
    
    \setlength{\tabcolsep}{3pt}
    \renewcommand{\arraystretch}{0.8} 

    \resizebox{\textwidth}{!}{
        \begin{tabular}{l l ccc ccc cc}
        \toprule
        \multirow{2}{*}{{Defense}} & 
        \multirow{2}{*}{{Attack}} & 
        \multicolumn{3}{c}{{Model-Agnostic Transformation}} & 
        \multicolumn{3}{c}{{Prior-Guided Purification}} & 
        \multicolumn{2}{c}{{Mechanism-Aware Adaptation}} \\
        
        \cmidrule(lr){3-5} \cmidrule(lr){6-8} \cmidrule(lr){9-10}
        
        & & GNoise & GSmooth & JPEG Comp. & Impress~\cite{impress} & Impress++~\cite{impress++} & Upscale~\cite{upscaler} & DiffShortcut~\cite{diffshortcut} & TimestepSel. \\
        \midrule
        \multirow{3}{*}{{Smooth~\cite{smooth}}}
        & SDEdit~\cite{sdedit} & 111.34 & 25.50 & 101.90 & 62.43 & 50.20 & 36.26 & 76.97 & 7.81 \\
        
        & Inpaint~\cite{ldm} & 109.15 & 72.39 & 109.15 & 109.15 & 97.75 & 109.15 & 86.74 & 5.27 \\
        
        & ControlNet~\cite{controlnet} & 100.32 & 29.88 & 92.46 & 61.62 & 47.98 & 83.33 & 61.06 & 5.88 \\
        
        \midrule
        
        \multirow{3}{*}{{ACE~\cite{ace}}}
        
        & Fine-tuning~\cite{ldm} & 209.69 & 73.33 & 134.53 & 197.91 & 152.96 & 192.34 & 87.99 & 40.99 \\
        
        & LoRA~\cite{lora} & 198.58 & 66.78 & 114.88 & 177.25 & 143.08 & 140.94 & 75.72 & 38.89 \\
        
        & SVDiff~\cite{disdiff} & 74.60 & 49.04 & 51.19 & 74.60 & 66.01 & 72.32 & 37.22 & 32.24 \\
        \midrule
        
        \rowcolor{Gray}
        & SDEdit~\cite{sdedit} & {205.26} & {162.34} & {200.54} & {180.81} & {175.70} & {167.72} & {113.24} & {316.67} \\
        
        \rowcolor{Gray}
        & Inpaint~\cite{ldm} & {204.49} & {195.93} & {207.55} & {224.96} & {206.78} & {254.78} & {122.16} & {339.21} \\
        
        \rowcolor{Gray}
        & ControlNet~\cite{controlnet} & {200.42} & {165.20} & {196.49} & {181.08} & {173.08} & {191.93} & {131.43} & {316.11} \\
        
        \rowcolor{Gray}
        & Fine-tuning~\cite{ldm} & {313.10} & {244.92} & {275.52} & {307.21} & {295.50} & {304.42} & {163.32} & {417.38} \\
        
        \rowcolor{Gray}
        & LoRA~\cite{lora} & {291.30} & {225.40} & {249.45} & {280.63} & {267.76} & {262.47} & {172.33} & {377.56} \\
        
        \rowcolor{Gray}
        \multirow{-6}{*}{{\sys (Ours)}}
        & SVDiff~\cite{disdiff} & {209.01} & {196.23} & {197.31} & {209.01} & {203.09} & {207.87} & {123.51} & {349.42} \\
        
        \bottomrule
        \end{tabular}
    }
\end{table*}
\begin{table}[htbp]
    \centering
    \caption{Defense effectiveness across heterogeneous mimicry Defenses. Comparison on TI-Dataset~\cite{ti}.}
    \label{tab:heterogeneous_mimicry}
    
    \setlength{\tabcolsep}{3.5pt}
    
    \resizebox{\columnwidth}{!}{
        \begin{tabular}{l ccc ccc}
        \toprule
        \multirow{2}{*}{\textbf{Defense}} & 
        \multicolumn{3}{c}{\textbf{DB~\cite{dreambooth}}} & \multicolumn{3}{c}{\textbf{TI~\cite{ti}}} \\
        \cmidrule(lr){2-4} \cmidrule(lr){5-7}
        & FID $\uparrow$ & CLIP-S $\downarrow$ & CLIP-I $\downarrow$ & FID $\uparrow$ & CLIP-S $\downarrow$ & CLIP-I $\downarrow$ \\
        \midrule
        Smooth\cite{smooth} & 51.78 & 31.60 & 0.81 & 113.69 & 24.01 & \underline{0.64} \\
        ACE\cite{ace} & \underline{87.64} & \underline{28.98} & \underline{0.74} & \underline{149.28} & \underline{23.39} & 0.66 \\
        \rowcolor{Gray}
        {\sys (Ours)} & \textbf{222.32} & \textbf{28.05} & \textbf{0.69} & \textbf{247.47} & \textbf{19.85} & \textbf{0.60} \\
        \bottomrule
        \end{tabular}
    }
    \resizebox{\columnwidth}{!}{
        \begin{tabular}{l ccc ccc}
        \toprule
        \multirow{2}{*}{\textbf{Defense}} & 
        \multicolumn{3}{c}{\textbf{DI~\cite{vddimiedit}}} & \multicolumn{3}{c}{\textbf{DF~\cite{diffedit}}} \\
        \cmidrule(lr){2-4} \cmidrule(lr){5-7}
        & FID $\uparrow$ & CLIP-S $\downarrow$ & CLIP-I $\downarrow$ & FID $\uparrow$ & CLIP-S $\downarrow$ & CLIP-I $\downarrow$ \\
        \midrule
        Smooth\cite{smooth} & \underline{152.13} & 28.76 & 0.78 & 120.08 & 27.00 & 0.79 \\
        ACE\cite{ace} & 124.22 & \underline{26.12} & \underline{0.72} & \underline{159.28} & \underline{26.19} & \underline{0.77} \\
        \rowcolor{Gray}
        {\sys (Ours)} & \textbf{244.35} & \textbf{20.32} & \textbf{0.58} & \textbf{627.33} & \textbf{20.96} & \textbf{0.57} \\
        \bottomrule
        \end{tabular}
    }
\end{table}

\begin{figure}[tbp!]
  \centering
  \includegraphics[width=0.9\linewidth]{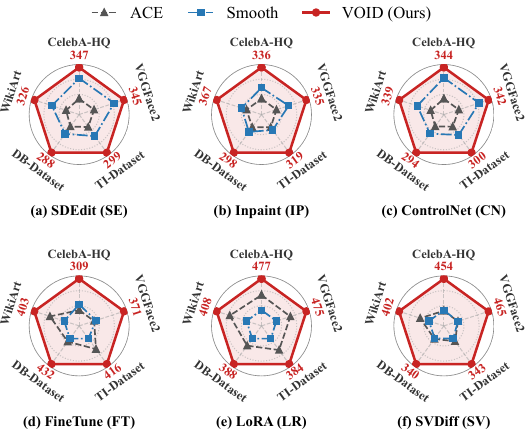}
    \caption{Evaluation of defense consistency across semantic domains. Radar charts illustrate FID scores (higher is better) across 5 diverse datasets under 6 mimicry attacks.}
    \label{fig:domain_consistency}
\end{figure}

\noindent\textbf{Cross-method transferability.} \cref{fig:domain_consistency} verifies the defense consistency of \sys across 6 mimicry attacks; we further extend the scope to 4 heterogeneous paradigms, spanning \emph{training-based personalization} (DreamBooth (DB)~\cite{dreambooth}, Textual Inversion (TI)~\cite{ti}) and \emph{inference-based editing} (DDIM Inversion (DI)~\cite{vddimiedit}, DiffEdit (DF)~\cite{diffedit}). As detailed in \cref{tab:heterogeneous_mimicry}, specialized baselines like ACE~\cite{ace} and Smooth~\cite{smooth} exhibit distinct performance polarity. In contrast, \sys demonstrates universal superiority, notably achieving a dominant FID~\cite{fid} of \textbf{627.33} in the DiffEdit~\cite{diffedit} setting, surpassing the runner-up ACE~\cite{ace} by nearly \textbf{4$\times$}. This confirms that the semantic-corruption paradigm serves as a highly transferable defense capable of neutralizing diverse mimicry mechanisms.

\noindent\textbf{Cross-domain transferability.}
We evaluate \sys across 5 distinct domains, including facial identities, general objects, and artistic styles. The results, visualized in \cref{fig:domain_consistency}, demonstrate that \sys maintains consistent protection stability across all categories. Quantitatively, \sys achieves high disruption levels with FID scores consistently exceeding 300 (peaking at \textbf{477} in LoRA~\cite{lora}) and exhibits negligible variance. In contrast, baselines like ACE and Smooth display significant domain bias, evidenced by their inconsistent radar profiles. This confirms that \sys is content-agnostic, ensuring robust protection regardless of the target image semantics.

\begin{table*}[ht]
    \centering
    \caption{Ablation study of \sys components. We evaluate Effectiveness, Utility, Transferability, and Robustness.}
    \label{tab:ablation}

    \setlength{\tabcolsep}{5pt}
    \renewcommand{\arraystretch}{0.9}
    
    \resizebox{\textwidth}{!}{
        \begin{tabular}{l ccc ccc cc cc}
            \toprule

            \multirow{2}{*}{\textbf{Defense Component}} & 
            \multicolumn{3}{c}{\textbf{Effectiveness} (FID $\uparrow$)} & 
            \multicolumn{3}{c}{\textbf{Utility} (Visual Quality)} & 
            \multicolumn{2}{c}{\textbf{Transferability} (FID $\uparrow$)} & 
            \multicolumn{2}{c}{\textbf{Robustness} (FID $\uparrow$)} \\
            
            \cmidrule(lr){2-4} \cmidrule(lr){5-7} \cmidrule(lr){8-9} \cmidrule(lr){10-11}
            
            & Training-based & Inference-based & Avg. & SSIM ($\uparrow$) & PSNR ($\uparrow$) & LPIPS ($\downarrow$) & SD v2.1~\cite{sdv2.1} & LCM~\cite{lcm} & Impress~\cite{impress} & TimestepSel. \\
            \midrule
            
             EUA (Base)              & \textbf{413.66} & \textbf{317.91} & \textbf{365.79} & 0.789 & 32.766 & 0.182 & 365.67 & 370.24 & 263.44 & 354.78 \\
            
            + GSC                   & 392.30 & 308.06 & 350.18 & 0.789 & 32.775 & 0.180 & \textbf{384.64} & \textbf{398.23} & \textbf{284.13} & \textbf{378.90} \\
            
            + HGPM              & 376.64 & 292.43 & 334.54 & \textbf{0.894} & \textbf{35.327} & \textbf{0.081} & 328.03 & 325.75 & 215.70 & 330.27 \\

            + TPOS (\sys)  & 381.31 & 306.39 & 343.85 & 0.875 & 34.405 & 0.097 & 334.45 & 331.78 & 230.62 & 346.06 \\
            
            \bottomrule
        \end{tabular}
    }
\end{table*}
\subsection{Robustness against Countermeasures}
\label{sec:robustness_eval}

We evaluate \sys against three countermeasure categories on TI-Dataset~\cite{ti}. As shown in \cref{tab:countermeasures}, \sys consistently outperforms baselines across all settings.

\noindent\textbf{Model-Agnostic Transformations.}
Against standard transformations (Gaussian noise, smoothing, JPEG~\cite{jpeg}), \sys retains substantial efficacy even under aggressive smoothing. In contrast, baselines like ACE~\cite{ace} collapse to non-disruptive levels under identical conditions. This disparity confirms that \sys relies on resilient structural corruption rather than fragile high-frequency perturbations.

\noindent\textbf{Prior-Guided Purification.} \sys exhibits exceptional stability under advanced purification methods such as Impress~\cite{impress}. Specifically, for SDEdit~\cite{sdedit}, it retains a dominant FID of \textbf{180.81}, significantly outperforming Smooth~\cite{smooth} (62.43) and ACE~\cite{ace} (14.78). This indicates that the semantic-corruption is deeply entangled with generative features, rendering it highly resistant to gradient-based inversion.

\noindent\textbf{Mechanism-Aware Adaptation.}
Motivated by our decay analysis (\cref{sec:rq1_verification}), we employ \emph{Timestep Selection} to maximize diffusion intensity (inference strength 0.9, training start $t=500$). Unlike baselines that falter under such high-noise regimes, \sys exhibits exceptional resilience, achieving FIDs of \textbf{316.67} for SDEdit~\cite{sdedit} and \textbf{417.38} for Fine-tuning~\cite{ldm}. Moreover, its sustained effectiveness against DiffShortcut~\cite{diffshortcut} (FID $>$ \textbf{113}) further validates the defense's robustness against targeted, mechanism-specific countermeasures.

\begin{figure}[t]
    \centering
    \includegraphics[width=0.8\linewidth]{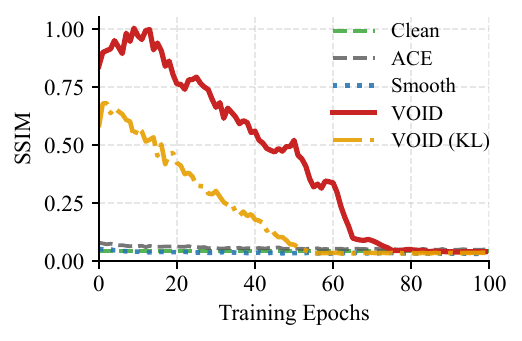}
    \caption{Optimization barrier under strong adaptive attacks.}
    \label{fig:adaptive_barrier}
\end{figure}
\begin{table}[tbp]
    \centering
    \caption{Strong adaptive attack results on TI-Dataset~\cite{ti}. Perturbations are crafted on SD v1.5 and evaluated on different target models under standard and KL-regularized training.}
    \label{tab:strong_adaptive}
    
    \setlength{\tabcolsep}{3pt}
    \renewcommand{\arraystretch}{0.8}
    
    \resizebox{\columnwidth}{!}{
        \begin{tabular}{l c c c c c}
        \toprule
        Target Model & FID$\uparrow$ & CLIP-S$\downarrow$ & CLIP-I$\downarrow$ & Memory(GB)$\downarrow$ & Time(H)$\downarrow$ \\
        \midrule
        SD v1.5~\cite{sdv1.5}$^{*}$ & 247.23 & 25.11 & 0.66 & 78.53 & 2.13 \\
        SD v1.5~\cite{sdv1.5}$^{*}$ + KL & 218.64 & 25.78 & 0.69 & 78.74 & 1.68 \\
        SD v2.1~\cite{sdv2.1} & 198.71 & 26.83 & 0.70 & 79.12 & 2.44 \\
        SD v2.1~\cite{sdv2.1} + KL & 176.32 & 27.21 & 0.72 & 79.35 & 1.92 \\
        \bottomrule
        \end{tabular}
    }
\end{table}

\noindent\textbf{Strong Adaptive Adversary.}
Existing adaptive evaluations~\cite{pretender} of defenses against LDM-based mimicry mainly fine-tune the U-Net, which largely overlaps with the FT attack~\cite{ldm} evaluated above. We implement a stronger adaptive attack that fine-tunes the full editing pipeline, including the VAE and U-Net, by editing the protected input with its own caption and forcing the edited output to reconstruct the input image in pixel space. We also test \sys (KL), which adds a KL regularizer to pull the VAE encoder's variance output toward the standard Gaussian, thereby reducing the EUA-induced variance shift. As shown in \cref{fig:adaptive_barrier}, \sys still creates a strong optimization barrier, with the attack loss stagnating for over 70 steps. We further report adaptive results in \cref{tab:strong_adaptive}, where protections crafted on SD v1.5~\cite{sdv1.5} are evaluated on both SD v1.5~\cite{sdv1.5} and SD v2.1~\cite{sdv2.1} under standard and KL-regularized training. Although the adapted models improve the quantitative editing metrics, they require about 79GB memory and over 2 hours on a single A100 GPU, while the outputs remain severely blurred as shown in \cref{fig:adaptive_collapse}. This shows that adaptive training is costly and still fails to recover usable mimicry. Detailed settings are provided in \cref{sec:appendix_adaptive}.

\begin{figure}[t]
    \centering
    
    \includegraphics[width=0.85\linewidth]{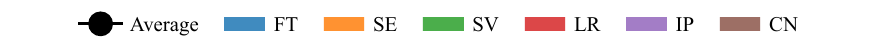}
    \begin{subfigure}[b]{0.495\linewidth}
        \centering
        \includegraphics[width=\linewidth]{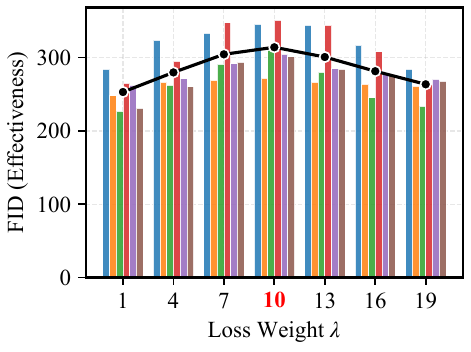}
        \caption{Loss Weight ($\lambda$)}
        \label{fig:sens_lambda}
    \end{subfigure}%
    \hfill
    \begin{subfigure}[b]{0.495\linewidth}
        \centering
        \includegraphics[width=\linewidth]{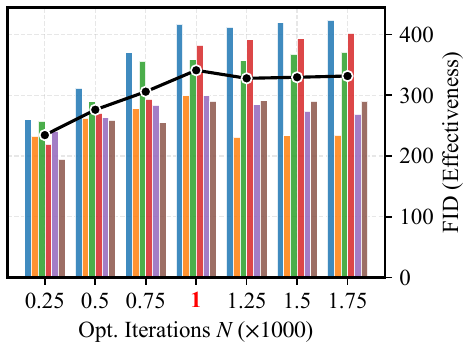}
        \caption{Optimization Iterations ($N$)}
        \label{fig:sens_iter}
    \end{subfigure}

    \vspace{5pt}
    
    \includegraphics[width=0.95\linewidth]{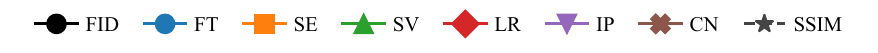}
    \begin{subfigure}[b]{0.495\linewidth}
        \centering
        \includegraphics[width=\linewidth]{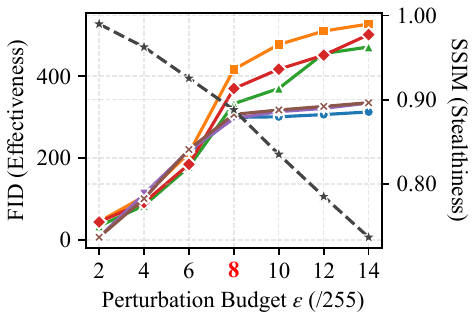}
        \caption{Perturbation Budget ($\alpha$)}
        \label{fig:sens_epsilon}
    \end{subfigure}%
    \hfill
    \begin{subfigure}[b]{0.495\linewidth}
        \centering
        \includegraphics[width=\linewidth]{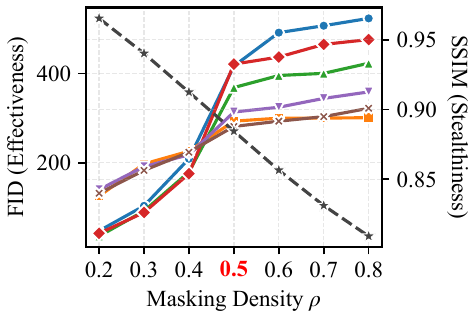}
        \caption{Masking Density ($\rho$)}
        \label{fig:sens_mask}
    \end{subfigure}
    \caption{Parameters sensitivity analysis.}
    \label{fig:parameter_sensitivity}
\end{figure}
\subsection{Ablation and Sensitivity Analysis}

\begin{figure}[t]
    \centering
    \includegraphics[width=\linewidth]{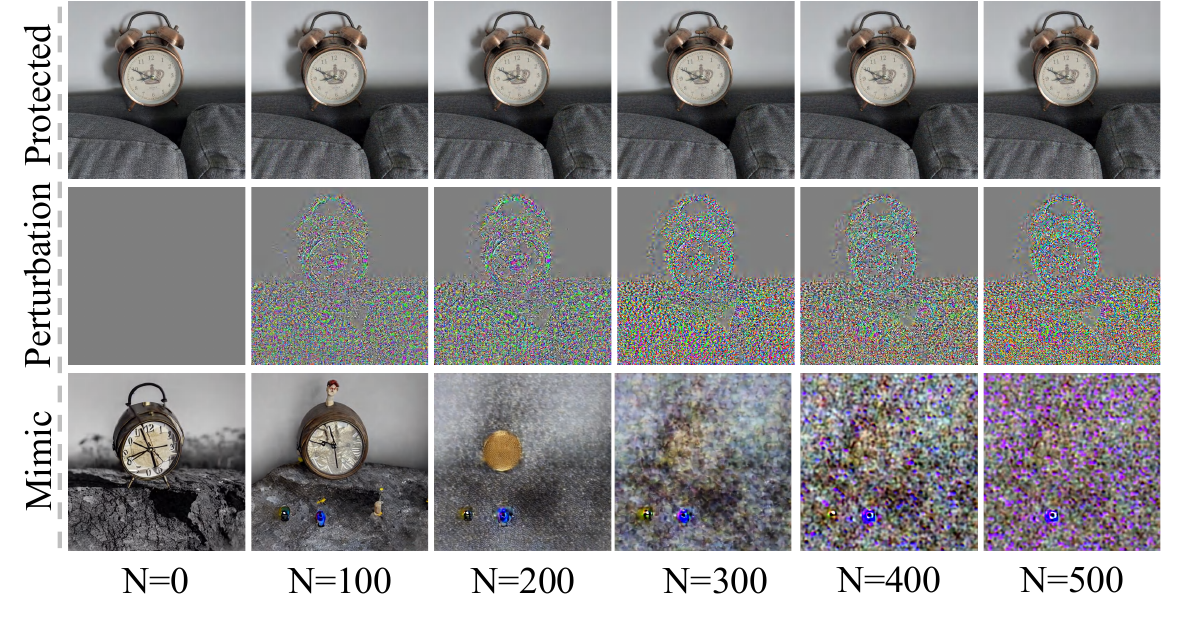}
    \caption{Visualization of optimization evolution.}
    \label{fig:optimization_evolution}
\end{figure}

\noindent\textbf{Component Ablation.}
We isolate the contribution of each \sys component in \cref{tab:ablation}. Encoding Uncertainty Amplification (EUA) is the primary source of trajectory collapse, yielding the highest baseline effectiveness (Avg. FID \textbf{365.79}) by maximizing latent variance, though it compromises visual utility. Guidance Signal Counteraction (GSC) improves generalization, achieving the best transferability (FID \textbf{398.23}) and robustness (FID \textbf{284.13}), confirming that neutralizing semantic guidance prevents surrogate overfitting. HVS-Guided Perturbation Masking (HGPM) confines perturbations to texture-rich regions and boosts PSNR to \textbf{35.327}, with a trade-off in raw attack strength. Timestep-Partitioned Optimal Selection (TPOS) mitigates stochastic sampling variance, recovering effectiveness (Avg. FID \textbf{343.85}) while preserving high fidelity. \cref{fig:optimization_evolution} further shows the evolution across optimization steps, where the protected image remains stable while mimicry outputs progressively degrade into global noise.

\begin{table}[tbp]
    \centering
    \caption{Sensitivity to perturbation budget and surrogate model setting on TI-Dataset~\cite{ti}.}
    \label{tab:budget_surrogate}
    
    \renewcommand{\arraystretch}{0.8}
    
    \resizebox{\columnwidth}{!}{
        \begin{tabular}{c l l c c}
        \toprule
        \textbf{Budget} & \textbf{Surrogate} & \textbf{Defense} & \textbf{SDEdit~\cite{sdedit}-FID$\uparrow$} & \textbf{FT~\cite{ldm}-FID$\uparrow$} \\
        \midrule
        \multicolumn{5}{c}{\textit{Perturbation budget}} \\
        \midrule
        \multirow{3}{*}{$8/255$}
        & \multirow{3}{*}{SDv1.5~\cite{sdv1.5}} 
        & Smooth~\cite{smooth} & 121.34 & 81.48 \\
        & & ACE~\cite{ace} & 24.78 & 219.69 \\
        & & \cellcolor{Gray}\sys (Ours) & \cellcolor{Gray}\textbf{299.18} & \cellcolor{Gray}\textbf{416.51} \\
        \cmidrule(lr){1-5}
        \multirow{3}{*}{$16/255$}
        & \multirow{3}{*}{SDv1.5~\cite{sdv1.5}} 
        & Smooth~\cite{smooth} & 187.66 & 154.21 \\
        & & ACE~\cite{ace} & 40.05 & 301.34 \\
        & & \cellcolor{Gray}\sys (Ours) & \cellcolor{Gray}\textbf{337.53} & \cellcolor{Gray}\textbf{481.40} \\
        \midrule
        \multicolumn{5}{c}{\textit{Surrogate setting}} \\
        \midrule
        \multirow{3}{*}{$8/255$}
        & SDv1.5~\cite{sdv1.5} 
        & \multirow{3}{*}{\sys (Ours)} & 307.07 & 415.56 \\
        & SDv2.1~\cite{sdv2.1}
        & & 300.28 & 387.30 \\
        & SDv1.5~\cite{sdv1.5}+v2.1~\cite{sdv2.1}
        & & \textbf{362.61} & \textbf{460.34} \\
        \cmidrule(lr){1-5}
        $16/255$
        & SDv1.5~\cite{sdv1.5}+2.1~\cite{sdv2.1}
        & \sys (Ours) & \textbf{429.87} & \textbf{499.51} \\
        \bottomrule
        \end{tabular}
    }
\end{table}

\noindent\textbf{Parameter Sensitivity.}
We assess the sensitivity of \sys to four key hyperparameters on TI-Dataset~\cite{ti}, as shown in \cref{fig:parameter_sensitivity}. Increasing the perturbation budget $\alpha$ improves protection but degrades visual quality, and $\alpha=8/255$ balances strong disruption (FID $>300$) with high fidelity (SSIM $\approx 0.88$). Effectiveness plateaus after 1000 iterations, so we set $N=1000$. The loss weight $\lambda$ balances EUA and GSC: moderate weighting improves transferability on SD v2.1~\cite{sdv2.1}, while excessive weighting suppresses EUA-driven corruption; we use $\lambda=10$. The masking density $\rho$ controls the trade-off between search space and stealthiness, and $\rho=0.5$ restricts perturbations to the top 50\% texture-rich regions. \cref{tab:budget_surrogate} further quantifies perturbation-budget and surrogate-model sensitivity, with results averaged over LCM~\cite{lcm} and DreamShaper~\cite{dreamshaper} as target models. Increasing the budget to $16/255$ improves protection, and using both SD v1.5~\cite{sdv1.5} and SD v2.1~\cite{sdv2.1} as surrogates boosts transferability, showing the potential of ensemble-based optimization.
\section{Discussion}
\label{sec:limitations}

\subsection{\sys as a Red-Teaming Tool}
\label{subsec:red_teaming}
Beyond privacy protection against mimicry, \sys can also evaluate LDM-based adversarial purification defenses. In this setting, generative purifiers such as OSCP~\cite{oscp} and DBLP~\cite{dblp} use LDM priors to remove adversarial perturbations and restore clean data manifolds before classification. However, their reliance on stable generation trajectories makes them vulnerable to semantic corruption. To test this, we apply the core EUA and GSC objectives without perceptual masking under a standard $\ell_\infty$ budget ($\epsilon=4/255$). As shown in \cref{tab:purifier_robustness}, purifier robust accuracy drops to nearly 0\%, even though these models remain robust against PGD~\cite{pgd} and AutoAttack~\cite{autoattack}. This gap shows that standard adversarial attacks treat LDMs as generic differentiable functions, while \sys directly targets their intrinsic generative dynamics. Therefore, \sys provides a complementary benchmark for exposing hidden fragility in LDM-based security modules.
\begin{table}[t]
    \centering
    \caption{Robustness evaluation (Accuracy \%) of LDM-based purifiers on ImageNet using ResNet-50. The perturbation budget is $\epsilon=4/255$, and PGD attack runs for 100 iterations.}
    \label{tab:purifier_robustness}
    
    \renewcommand{\arraystretch}{0.8} 
    \setlength{\tabcolsep}{8pt}

    \resizebox{\linewidth}{!}{
        \begin{tabular}{l c ccc}
            \toprule
            
            \multirow{2}{*}{{Defense}} & 
            \multirow{2}{*}{{Clean}} & 
            \multicolumn{3}{c}{{Evaluation Methods}} \\
            
            \cmidrule(lr){3-5}
            
            & & {PGD~\cite{pgd}} & {AutoAttack~\cite{autoattack}} & {\sys (Ours)} \\
            \midrule
            
            W/O Purifier & 80.55 & 0.01 & 0.00 & --- \\
            \midrule
            OSCP~\cite{oscp} & 77.63 & 73.89 & 74.19 & \textbf{0.11} \\
            DBLP~\cite{dblp} & 78.00 & 75.60 & 74.80 & \textbf{0.10} \\
            \bottomrule
        \end{tabular}
    }
\end{table}

\subsection{Limitations}
\label{subsec:limitations}
\noindent\textbf{Generalizability across Modalities.} 
Our current evaluation focuses exclusively on image-domain LDMs. While the generative capabilities of LDMs have expanded to other modalities such as audio, video, and 3D meshes, the applicability of \sys's semantic corruption paradigm in these domains remains unverified. Different modalities may exhibit distinct feature persistence dynamics and latent structures, potentially requiring domain-specific adaptations of our EUA and GSC modules. We leave the exploration of extending \sys to multimodal protection as future work.

\noindent\textbf{Computational Overhead.} 
The GSC module requires gradients from the guidance vector and thus performs both conditional and unconditional forward passes during optimization. This introduces a modest overhead compared to single-pass baselines such as AdvDM~\cite{advdm}. Given the significant performance advantage, we argue that this cost is acceptable for offline privacy protection and remains considerably lower than complex bilevel-optimization-based poisoning attacks. It is also a one-time cost incurred only during the protection phase, with no impact on subsequent image storage or usage.
\section{Conclusion}

The great potential of LDMs also brings unprecedented privacy threats. Existing semantic-steering mimicry defense mechanisms overlook the model's internal machinery that can render them futile, as we have elaborated in this paper. We explore an alternative principled semantic-corruption approach, which leverages the intrinsic stochastic nature of LDMs, and demonstrate its effectiveness through the design, implementation, and extensive evaluation of the \sys framework. 

While this paper focuses primarily on image synthesis, we believe that the proposed semantic-corruption paradigm has unveiled a fundamental trait of diffusion priors, in particular their reliance on continuous latent evolution. This suggests transferability to other modalities such as voice cloning and 3D synthesis. An interesting research direction is to turn \sys into a unified framework that can support non-image domains and thwart unforeseen mimicry attacks. 
\section*{Acknowledgments}

We thank the anonymous reviewers for their valuable feedback. This work was supported by National Key Research and Development Program of China (2023YFB3107400), the NSFC under Grants  U2441240 (``Ye Qisun'' Science Foundation), 62441238, 62302344, 62441237, 62502350, 62521002, 62132011, U24B20185, the Fundamental and Interdisciplinary Disciplines Breakthrough Plan of the Ministry of Education of China (JYB2025XDXM114). Thanks to the New Cornerstone Science Foundation and the Xplorer Prize.
\section*{Ethical Considerations}

Since \sys studies adversarial perturbations against generative models, it may create misuse risks. We discuss its impacts on four stakeholder groups.

\noindent\textbf{Protected data and content owners.}
This group includes individuals, creators, artists, and IP holders who may face unauthorized identity or style mimicry. \sys provides a proactive opt-out defense, but its protection is not absolute and may be weakened by future adaptive attacks. It should therefore complement consent-based data protection, platform safeguards, and policy interventions.

\noindent\textbf{Adversaries and malicious actors.}
\sys is designed to raise the cost of unauthorized mimicry. However, perturbation-based techniques may be repurposed for data poisoning, model disruption, or adaptive circumvention. We reduce these risks by framing \sys as a defensive tool and releasing artifacts with restrictive licensing and careful documentation.

\noindent\textbf{Benign AI ecosystem actors.}
Developers, dataset maintainers, and model providers may use \sys for robustness evaluation and non-consensual data-use analysis. However, widespread perturbation-based protection may increase filtering costs, introduce false positives, or interfere with legitimate data use. Such defenses should complement transparent data policies and platform-level consent mechanisms.

\noindent\textbf{Research and publication stakeholders.}
Our experiments use public datasets and collect no new private information. Still, represented people or works may raise consent or redistribution concerns, especially in face-related settings. We obscure sensitive facial attributes and gestures in mimicry results and conduct no human-subject study. Although publication may inform circumvention attempts, we mitigate this risk through restrictive licensing.

Overall, publication is ethically warranted because unauthorized mimicry causes concrete privacy and intellectual-property harms. \sys should be viewed as one defensive layer, not a replacement for broader consent, provenance, and governance mechanisms.

\section*{Open Science}

We make the artifacts supporting our core contribution publicly available. The release includes:
(1) the source code of \sys;
(2) the mimicry attack implementations used in our evaluation;
and (3) evaluation scripts for measuring defense effectiveness, including FID, CLIP-S, and CLIP-I. The artifact package and setup instructions are available at:

\begin{center}
\url{https://doi.org/10.5281/zenodo.20233998}
\end{center}

\bibliographystyle{plain}
\bibliography{reference}
\appendix

\begin{figure}[t]
\centering
\includegraphics[width=\columnwidth]{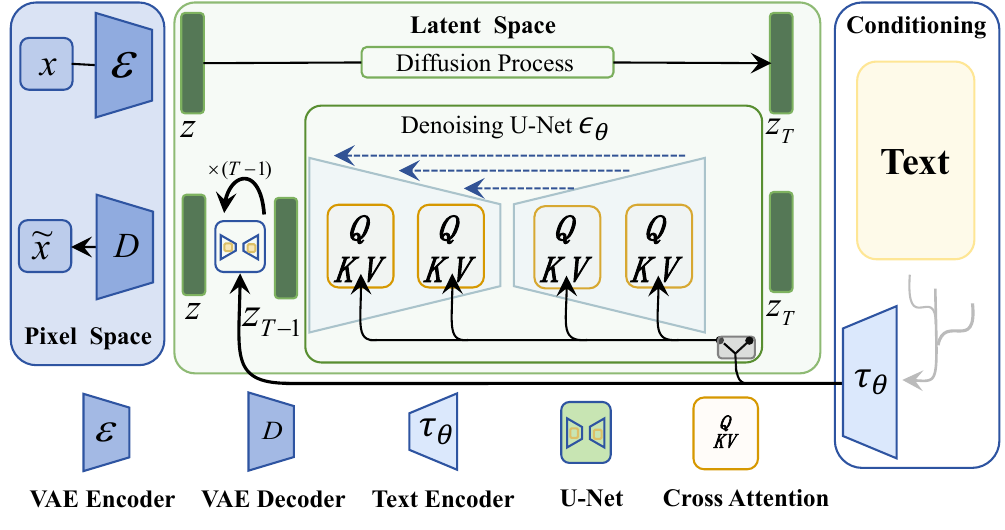}
\caption{Overall architecture of Latent Diffusion Models.}
\label{fig:ldm_architecture}
\end{figure}
\section{Latent Diffusion Models}
\label{sec:appendix_ldm}
Latent Diffusion Models (LDMs)~\cite{ldm,flux} have become the de facto standard for high-fidelity image synthesis. Structurally, LDMs decouple the generative process into two distinct stages: perceptual compression and semantic generation. As shown in \cref{fig:ldm_architecture}, a VAE encoder $\mathcal{E}$~\cite{vae} first maps the input $x$ to a low-dimensional latent $z_0 = \mathcal{E}(x)$, and a decoder $\mathcal{D}$ reconstructs the image from the latent.

The core generation occurs in the latent space via a probabilistic diffusion process. Formally, the \emph{forward process} is a fixed Markov chain that gradually destroys the data structure by injecting Gaussian noise $n \sim \mathcal{N}(0, \mathbf{I})$ over $T$ steps. At any timestep $t$, the noisy latent $z_t$ can be expressed as:
\begin{equation}
    \label{eq:forward_process_latent}
    z_t = \sqrt{\bar{\alpha}_t} z_0 + \sqrt{1 - \bar{\alpha}_t} n,
\end{equation}
where $\{\bar{\alpha}_t\}$ follows a fixed variance schedule.
The \emph{reverse process} employs a time-conditional U-Net $\epsilon_\theta$~\cite{unet} to reverse this corruption.
Trained by minimizing the mean squared error between the predicted and actual noise, the objective function is:
\begin{equation}
\mathcal{L}_{\text{diff}} = \mathbb{E}_{t, x_0, n}\left[\|\epsilon - \epsilon_\theta(z_t, t, c)\|^2\right],
\label{eq:diff}
\end{equation}
where $c$ denotes the conditioning signal (e.g., text prompts).
Minimizing this objective allows the model to learn the data distribution conditioned on $c$. During inference, to strictly enforce semantic alignment with $c$, LDMs utilize Classifier-Free Guidance (CFG)~\cite{cfg}. The final noise prediction $\hat{\epsilon}$ is extrapolated by scaling a \emph{guidance vector}:
\begin{equation}
    \hat{\epsilon}_\theta(z_t, t, c) = \epsilon_\theta(z_t, t, \emptyset) + w \cdot \underbrace{(\epsilon_\theta(z_t, t, c) - \epsilon_\theta(z_t, t, \emptyset))}_{\text{guidance vector}},
    \label{eq:cfg}
\end{equation}
where $w$ is the guidance scale. This vector dictates the direction towards the conditional semantics $c$, effectively serving as the primary control mechanism for generation.

The computational efficiency and open availability of LDM weights have fundamentally lowered the barrier for high-fidelity synthesis~\cite{ldm,lora}.
Unlike API-based models restricted by centralized safeguards, these models are typically deployed on personal devices in uncontrolled environments. This grants adversaries full control over the inference pipeline, allowing them to bypass safety checkers entirely and weaponize community-developed tools (e.g., LoRA~\cite{lora}, ControlNet~\cite{controlnet}) for unauthorized mimicry of identities and artistic styles~\cite{photoguard,mist}. Consequently, passive restrictions are insufficient, creating a compelling need for proactive defenses embedded directly within the data.

\begin{figure}[t]
  \centering
  
  \begin{minipage}[c]{0.05\columnwidth} 
    \scriptsize \centering
    \rotatebox{90}{ACE~\cite{ace}}
  \end{minipage}%
  \begin{minipage}[c]{0.85\columnwidth}
    \includegraphics[width=\linewidth]{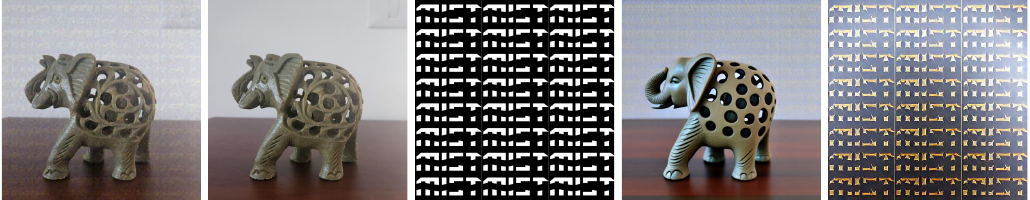}
  \end{minipage}
  \begin{minipage}[c]{0.05\columnwidth}
    \scriptsize \centering
    \rotatebox{90}{AntiDB~\cite{antidb}}
  \end{minipage}%
  \begin{minipage}[c]{0.85\columnwidth}
    \includegraphics[width=\linewidth]{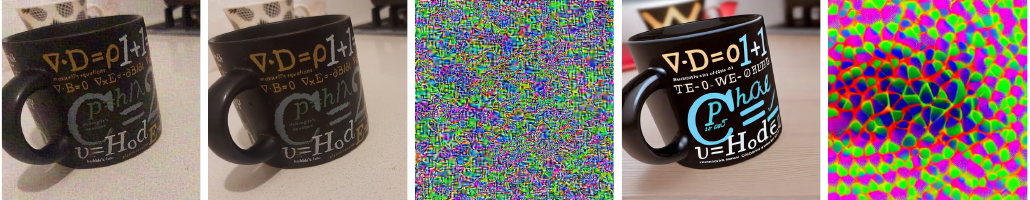}
  \end{minipage}
  \begin{minipage}[c]{0.05\columnwidth}
    \scriptsize \centering
    \rotatebox{90}{PhotoGuard~\cite{photoguard}}
  \end{minipage}%
  \begin{minipage}[c]{0.85\columnwidth}
    \includegraphics[width=\linewidth]{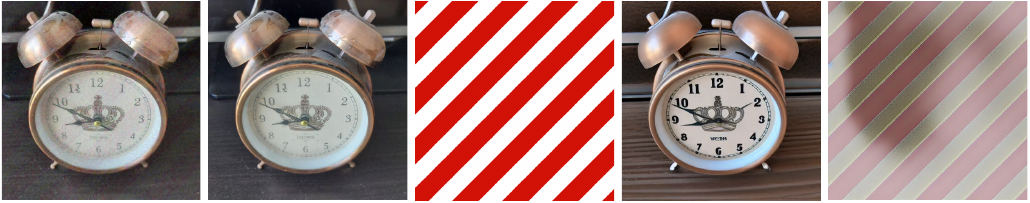}
  \end{minipage}
  \begin{minipage}[c]{0.05\columnwidth}
    \scriptsize \centering
    \rotatebox{90}{Smooth~\cite{smooth}}
  \end{minipage}%
  \begin{minipage}[c]{0.85\columnwidth}
    \includegraphics[width=\linewidth]{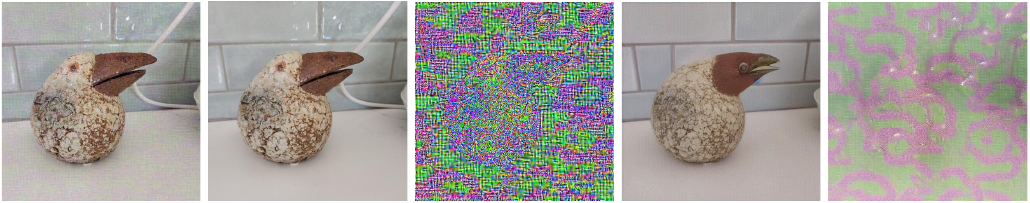}
  \end{minipage}
  \begin{minipage}[c]{0.05\columnwidth}
    \mbox{}
  \end{minipage}%
  \begin{minipage}[c]{0.85\columnwidth}
    \includegraphics[width=\linewidth]{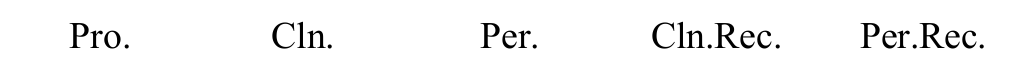} 
  \end{minipage}
    \caption{Visual reconstruction of original images and protective perturbations from protected samples.}
  \label{fig:feature_reconstruction}
\end{figure}
\begin{figure}[tbp!]
    \centering
    \includegraphics[width=\columnwidth]{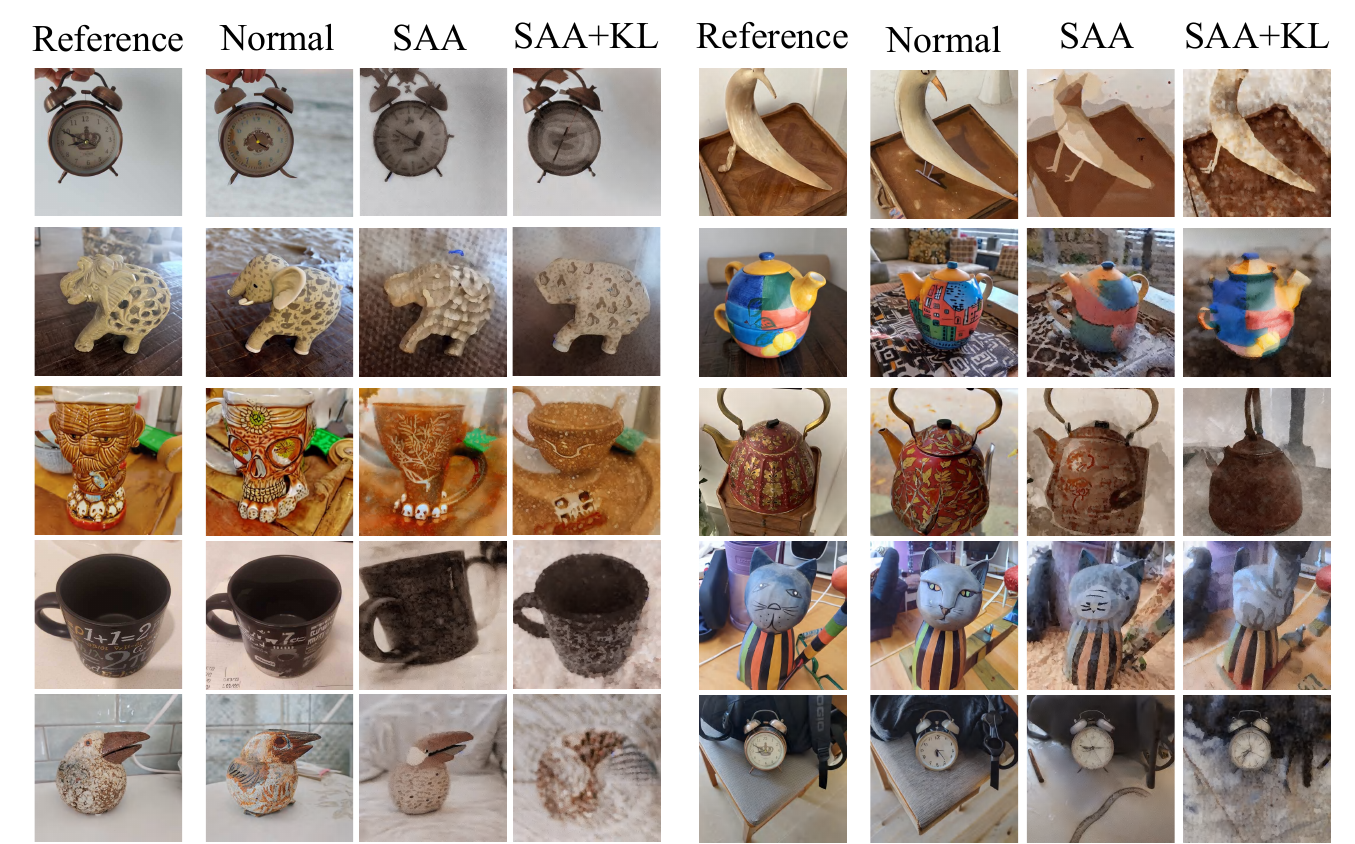}
    \caption{Functional collapse under adaptive adversary.}
    \label{fig:adaptive_collapse}
\end{figure}
\begin{table}[tbp]
    \centering
    \caption{Semantic specificity analysis on SD v1.5~\cite{sdv1.5}.}
    \label{tab:semantic_specificity}
    
    \setlength{\tabcolsep}{10pt}
    \renewcommand{\arraystretch}{0.8}
    
    \resizebox{\columnwidth}{!}{
        \begin{tabular}{l c c c c}
        \toprule
        \multirow{2}{*}{{Prompt Group}} 
        & \multicolumn{2}{c}{{FID$\uparrow$}} 
        & \multicolumn{2}{c}{{CLIP-S$\downarrow$}} \\
        \cmidrule(lr){2-3}
        \cmidrule(lr){4-5}
        & w/o & \sys & w/o & \sys \\
        \midrule
        Protected-concept & 23.75 & 415.35 & 28.89 & 19.12 \\
        Near-concept      & 20.37 & 37.42  & 28.03 & 24.39 \\
        Unrelated-concept & 19.52 & 19.93  & 27.88 & 27.08 \\
        \bottomrule
        \end{tabular}
    }
\end{table}
\section{Visual Reconstruction of Mixed Features}
\label{sec:appendix_reconstruction}
This section provides qualitative evidence for Observation 2 in \cref{sec:limitations_feature}. We apply DDIM inversion~\cite{ddim} to protected latents and denoise them with neutral prompts. As shown in \cref{fig:feature_reconstruction}, the model separates a protected sample into a reconstructed original identity and an isolated protective perturbation. This shows that steering-based defenses preserve semantic links, allowing the model's restoration mechanism to recover protected content.

\section{Implementation Details}
\label{sec:implement_details}
For \textit{inference-based editing}, we set the editing strength to 0.7. For \textit{training-based personalization}, we use $t=300$ as the starting point of the diffusion training range and sample timesteps from $300$ to $1000$. All defenses use an $L_\infty$ budget of $\alpha=8/255$ and 1000 optimization steps. Following PhotoGuard~\cite{photoguard}, protection generation uses \texttt{float32} to reduce quantization errors, while attacks use mixed precision. For \sys, we set $\lambda=10$ and $\rho=50$ to select the top 50\% JND regions. Experiments mainly run on NVIDIA RTX 4090 GPUs, except PhotoGuard generation and the holistic adaptive attack, which run on NVIDIA A100 (80GB) GPUs.

\section{Details of Strong Adaptive Adversary}
\label{sec:appendix_adaptive}
We design a holistic adaptive adversary that exploits $x_{prot} \approx x_{clean}$ by using the protected image as the reconstruction target. The attack fine-tunes the full editing pipeline, including the VAE Encoder/Decoder and U-Net. It injects noise into $x_{prot}$ at timestep $t=600$, performs 15-step unrolled denoising with the image caption, and backpropagates the pixel-space reconstruction loss through the loop. This attack requires about 79GB VRAM on a single A100 GPU. Despite this cost, \sys creates a strong optimization barrier. As shown in \cref{fig:adaptive_barrier}, optimization stagnates for over 70 steps due to unstable gradients from EUA-amplified latent variance. We also evaluate VOID (KL), where the adversary adds a KL regularizer to pull the VAE posterior toward the standard Gaussian and reduce the EUA-induced variance shift. Although KL regularization accelerates convergence, it does not restore usable mimicry. As shown in \cref{fig:adaptive_collapse}, the adapted model still produces severely blurred outputs, indicating functional degradation. Thus, even a holistic adaptive adversary improves reconstruction metrics only at high cost and fails to restore visually usable editing.

\section{Semantic Specificity Analysis}
\label{sec:appendix_specificity}
\sys protects images by corrupting the semantics of the protected concept, but this should not imply global generation failure. To verify semantic specificity and separate concept-specific corruption from global instability, we fine-tune SD v1.5~\cite{sdv1.5} on \sys-protected images and evaluate protected-concept, near-concept, and unrelated-concept prompts. If \sys destabilizes the whole diffusion pipeline, all prompt groups should degrade similarly. As shown in \cref{tab:semantic_specificity}, degradation is strongest for protected concepts, where FID~\cite{fid} increases from 23.75 to 415.35 and CLIP-S~\cite{clipscore} drops from 28.89 to 19.12. The effect weakens on near concepts and nearly disappears on unrelated concepts, confirming that \sys mainly corrupts protected semantics rather than the overall diffusion generation process.

\begin{table*}[htbp]
\centering
\caption{\textbf{FID scores} quantifying visual degradation under \textbf{inference-based mimicry attacks} (SDEdit~\cite{sdedit}, Inpainting~\cite{ldm}, ControlNet~\cite{controlnet}) across five datasets. The best and second-best results are highlighted in \textbf{bold} and \underline{underline}, respectively.}
\label{tab:infer_fid}

\setlength{\tabcolsep}{2.5pt}
\renewcommand{\arraystretch}{0.95}
\resizebox{\textwidth}{!}{
\begin{tabular}{l ccc ccc ccc ccc ccc c}
\toprule

\multirow{2}{*}{\textbf{Defense}} & \multicolumn{3}{c}{\textbf{TI-Dataset\cite{ti}}} & \multicolumn{3}{c}{\textbf{DB-Dataset\cite{dreambooth}}} & \multicolumn{3}{c}{\textbf{CelebA-HQ\cite{celeba-hq}}} & \multicolumn{3}{c}{\textbf{VGGFace2\cite{vggface2}}} & \multicolumn{3}{c}{\textbf{WikiArt\cite{wikiart}}} & \multirow{2}{*}{\textbf{AVG}} \\

\cmidrule(lr){2-4} \cmidrule(lr){5-7} \cmidrule(lr){8-10} \cmidrule(lr){11-13} \cmidrule(lr){14-16}
& SE\cite{sdedit} & IP\cite{ldm} & CN\cite{controlnet} & SE\cite{sdedit} & IP\cite{ldm} & CN\cite{controlnet} & SE\cite{sdedit} & IP\cite{ldm} & CN\cite{controlnet} & SE\cite{sdedit} & IP\cite{ldm} & CN\cite{controlnet} & SE\cite{sdedit} & IP\cite{ldm} & CN\cite{controlnet} & \\
\midrule
RandomNoise & 7.46 & 5.09 & 5.60 & 2.53 & 2.08 & 2.10 & 12.22 & 3.42 & 10.32 & 10.87 & 7.89 & 9.68 & 1.37 & 2.48 & 1.51 & 5.64 \\
LDMR\cite{ldmr} & 26.45 & 67.86 & 24.67 & 14.92 & 43.44 & 18.46 & 65.79 & 78.12 & 70.79 & 43.47 & 55.66 & 40.81 & 11.28 & 77.15 & 16.04 & 43.66 \\
AdvDM\cite{advdm} & 21.77 & 66.26 & 20.67 & 7.32 & 38.68 & 8.01 & 39.83 & 89.25 & 42.78 & 36.76 & 96.41 & 37.06 & 8.83 & 74.33 & 9.87 & 39.85 \\
MIST\cite{mist} & 20.92 & 85.00 & 18.20 & 15.20 & 46.96 & 14.87 & 72.03 & 95.23 & 54.05 & 65.37 & 96.50 & 53.49 & 9.53 & 66.66 & 10.72 & 48.32 \\
PhotoGuard\cite{photoguard} & 14.05 & 36.63 & 12.30 & 6.44 & 19.65 & 6.65 & 37.46 & 31.80 & 33.01 & 33.82 & 34.79 & 27.81 & 6.28 & 39.84 & 7.34 & 23.19 \\
Glaze\cite{glaze} & 21.72 & 86.92 & 18.92 & 15.07 & 46.38 & 14.91 & 70.92 & 87.52 & 54.12 & 65.93 & 100.42 & 53.36 & 15.47 & 63.86 & 18.81 & 48.96 \\
Smooth\cite{smooth} & \underline{121.34} & \underline{119.15} & \underline{110.32} & \underline{95.01} & \underline{96.76} & \underline{94.49} & \underline{250.34} & \underline{169.09} & \underline{248.35} & \underline{251.64} & \underline{182.55} & \underline{248.67} & \underline{152.54} & \underline{121.25} & \underline{152.27} & \underline{160.92} \\
SDST\cite{sdst} & 21.33 & 55.17 & 17.82 & 13.36 & 33.17 & 12.69 & 72.58 & 33.97 & 55.21 & 66.15 & 34.65 & 50.95 & 8.65 & 47.09 & 11.28 & 35.61 \\
PID\cite{pid} & 16.88 & 53.13 & 14.50 & 9.47 & 31.12 & 10.24 & 51.93 & 34.31 & 45.97 & 37.06 & 32.19 & 33.65 & 5.67 & 32.24 & 6.91 & 27.68 \\
DiffGuard\cite{diffusionguard} & 10.49 & 30.00 & 9.82 & 5.64 & 17.13 & 5.84 & 10.57 & 18.28 & 12.02 & 14.24 & 22.87 & 13.64 & 10.01 & 32.77 & 9.41 & 14.85 \\
NightShade\cite{nightshade} & 21.92 & 87.09 & 19.13 & 16.75 & 48.35 & 16.20 & 77.25 & 81.01 & 56.19 & 70.86 & 96.66 & 55.55 & 16.30 & 66.56 & 19.91 & 49.98 \\
GAPDiff\cite{gapdiff} & 12.71 & 57.20 & 13.25 & 6.28 & 28.66 & 7.78 & 36.89 & 81.67 & 36.56 & 36.49 & 81.87 & 35.96 & 7.98 & 50.03 & 8.89 & 33.48 \\
LDS\cite{lds} & 6.23 & 24.31 & 9.55 & 2.01 & 13.26 & 4.97 & 3.46 & 20.74 & 16.76 & 4.04 & 20.67 & 16.48 & 2.08 & 28.41 & 7.59 & 12.04 \\
EUDP\cite{eudp} & 10.34 & 63.18 & 10.25 & 5.34 & 37.93 & 6.53 & 18.65 & 77.69 & 21.15 & 17.30 & 74.54 & 19.04 & 4.01 & 74.41 & 5.07 & 29.69 \\
ACE\cite{ace} & 24.78 & 95.06 & 20.73 & 19.05 & 52.05 & 17.97 & 67.43 & 78.50 & 56.79 & 60.65 & 77.34 & 51.44 & 11.18 & 72.16 & 12.92 & 47.87 \\
AntiDB\cite{antidb} & 40.26 & 75.99 & 33.73 & 19.78 & 46.79 & 21.85 & 77.37 & 97.70 & 83.25 & 61.94 & 133.95 & 66.48 & 68.87 & 88.47 & 71.53 & 65.86 \\
SimAC\cite{simac} & 8.68 & 53.82 & 8.51 & 3.98 & 31.86 & 4.41 & 12.19 & 60.26 & 13.16 & 10.50 & 48.98 & 11.30 & 8.36 & 56.45 & 7.61 & 22.67 \\
MetaCloak\cite{metacloak} & 12.83 & 59.56 & 13.76 & 7.14 & 39.83 & 9.01 & 27.56 & 86.12 & 30.19 & 25.03 & 108.58 & 27.20 & 26.83 & 74.59 & 29.00 & 38.48 \\
DisDiff\cite{disdiff} & 10.47 & 58.80 & 10.30 & 5.61 & 36.81 & 6.77 & 11.01 & 70.25 & 14.04 & 10.55 & 70.06 & 12.25 & 13.96 & 75.85 & 15.00 & 28.11 \\
CAAT\cite{caat} & 14.22 & 64.77 & 14.40 & 9.21 & 40.61 & 11.44 & 39.69 & 85.39 & 44.55 & 31.56 & 116.43 & 36.14 & 42.25 & 85.07 & 45.23 & 45.40 \\
PAP\cite{pap} & 48.11 & 83.68 & 41.30 & 35.12 & 56.07 & 37.10 & 120.09 & 109.16 & 128.69 & 108.76 & 152.77 & 116.16 & 93.47 & 91.25 & 96.13 & 87.86 \\
AntiDiff\cite{antidiffusion} & 19.41 & 56.37 & 18.19 & 14.27 & 36.29 & 16.17 & 54.06 & 82.40 & 58.16 & 43.91 & 116.77 & 47.67 & 43.63 & 74.86 & 46.10 & 48.55 \\
GoodAC\cite{goodac} & 32.75 & 73.45 & 29.61 & 26.22 & 49.32 & 28.57 & 89.79 & 104.15 & 97.39 & 76.50 & 147.09 & 82.53 & 94.22 & 91.93 & 97.36 & 74.73 \\
Pretender\cite{pretender} & 14.39 & 68.70 & 14.81 & 8.33 & 39.51 & 10.01 & 34.79 & 82.12 & 38.11 & 30.18 & 103.94 & 33.31 & 34.56 & 77.15 & 36.97 & 41.79 \\
HAAD\cite{haad} & 33.41 & 90.21 & 20.87 & 24.55 & 65.46 & 24.62 & 109.77 & 155.06 & 86.14 & 100.78 & 149.52 & 92.72 & 21.46 & 99.45 & 27.55 & 73.44 \\
\midrule
\rowcolor{Gray}
{\sys (Ours)} & \textbf{299.18} & \textbf{319.47} & \textbf{300.53} & \textbf{288.63} & \textbf{298.80} & \textbf{294.48} & \textbf{347.23} & \textbf{336.90} & \textbf{344.39} & \textbf{345.85} & \textbf{335.22} & \textbf{342.18} & \textbf{326.46} & \textbf{367.94} & \textbf{339.71} & \textbf{325.80} \\
\bottomrule
\end{tabular}
}
\end{table*}
\begin{table*}[htbp]
\centering
\caption{\textbf{CLIP-S scores} measuring semantic alignment under \textbf{inference-based mimicry attacks} (SDEdit~\cite{sdedit}, Inpainting~\cite{ldm}, ControlNet~\cite{controlnet}) across five datasets. The best and second-best results are highlighted in \textbf{bold} and \underline{underline}, respectively.}
\label{tab:infer_clip_s}

\setlength{\tabcolsep}{2.5pt}
\renewcommand{\arraystretch}{0.95}
\resizebox{\textwidth}{!}{
\begin{tabular}{l ccc ccc ccc ccc ccc c}
\toprule

\multirow{2}{*}{\textbf{Defense}} & \multicolumn{3}{c}{\textbf{TI-Dataset\cite{ti}}} & \multicolumn{3}{c}{\textbf{DB-Dataset\cite{dreambooth}}} & \multicolumn{3}{c}{\textbf{CelebA-HQ\cite{celeba-hq}}} & \multicolumn{3}{c}{\textbf{VGGFace2\cite{vggface2}}} & \multicolumn{3}{c}{\textbf{WikiArt\cite{wikiart}}} & \multirow{2}{*}{\textbf{AVG}} \\

\cmidrule(lr){2-4} \cmidrule(lr){5-7} \cmidrule(lr){8-10} \cmidrule(lr){11-13} \cmidrule(lr){14-16}
& SE\cite{sdedit} & IP\cite{ldm} & CN\cite{controlnet} & SE\cite{sdedit} & IP\cite{ldm} & CN\cite{controlnet} & SE\cite{sdedit} & IP\cite{ldm} & CN\cite{controlnet} & SE\cite{sdedit} & IP\cite{ldm} & CN\cite{controlnet} & SE\cite{sdedit} & IP\cite{ldm} & CN\cite{controlnet} & \\
\midrule
RandomNoise & 32.87 & 29.94 & 31.78 & 30.76 & 28.49 & 29.63 & 18.38 & 15.35 & 17.12 & 23.63 & 21.80 & 22.56 & 34.34 & 29.37 & 32.24 & 26.55 \\
LDMR\cite{ldmr} & 32.54 & 30.01 & 31.88 & 30.79 & 28.89 & 30.00 & 18.55 & \underline{14.70} & 16.80 & 23.72 & 21.15 & 22.80 & 33.98 & 28.86 & 32.13 & 26.45 \\
AdvDM\cite{advdm} & 32.60 & 30.14 & 31.75 & 31.36 & 29.37 & 30.42 & 20.22 & 16.67 & 18.11 & 24.94 & 22.20 & 23.46 & 33.63 & 29.27 & 31.99 & 27.07 \\
MIST\cite{mist} & 33.08 & 29.86 & 32.26 & 31.36 & 29.22 & 30.54 & 21.53 & 16.60 & 18.40 & 25.59 & 23.21 & 23.42 & 34.37 & 29.18 & 32.27 & 27.39 \\
PhotoGuard\cite{photoguard} & 33.17 & 30.14 & 32.11 & 31.11 & 28.84 & 30.03 & 20.06 & 16.63 & 17.92 & 25.10 & 21.92 & 23.37 & 33.75 & 29.11 & 31.91 & 27.01 \\
Glaze\cite{glaze} & 33.10 & 29.65 & 32.22 & 31.38 & 29.20 & 30.55 & 21.55 & 16.87 & 18.43 & 25.60 & 23.47 & 23.43 & 34.21 & 28.75 & 31.22 & 27.31 \\
Smooth\cite{smooth} & \underline{30.20} & \underline{28.86} & \underline{30.09} & \underline{29.56} & \underline{28.28} & \underline{28.99} & \underline{18.25} & 18.98 & \underline{16.74} & \underline{23.10} & \underline{19.98} & \underline{22.48} & \underline{29.80} & \underline{28.38} & \underline{28.77} & \underline{25.50} \\
SDST\cite{sdst} & 33.47 & 30.53 & 32.48 & 31.55 & 29.47 & 30.61 & 21.56 & 17.47 & 18.15 & 25.76 & 22.33 & 23.35 & 33.74 & 28.57 & 31.81 & 27.39 \\
PID\cite{pid} & 33.34 & 31.02 & 32.39 & 31.41 & 29.49 & 30.51 & 20.56 & 17.09 & 18.10 & 24.85 & 22.16 & 23.35 & 33.66 & 29.40 & 31.92 & 27.28 \\
DiffGuard\cite{diffusionguard} & 32.95 & 30.24 & 31.96 & 30.86 & 28.72 & 29.85 & 18.79 & 15.66 & 17.66 & 23.89 & 20.95 & 22.87 & 33.38 & 28.94 & 31.76 & 26.56 \\
NightShade\cite{nightshade} & 33.06 & 29.74 & 32.21 & 31.35 & 29.17 & 30.55 & 21.74 & 16.62 & 18.43 & 25.72 & 21.61 & 23.44 & 33.02 & 28.48 & 31.15 & 27.09 \\
GAPDiff\cite{gapdiff} & 32.98 & 30.45 & 32.11 & 31.09 & 29.09 & 30.18 & 19.82 & 16.33 & 17.84 & 24.54 & 21.13 & 23.09 & 33.32 & 29.02 & 31.78 & 26.85 \\
LDS\cite{lds} & 32.87 & 31.18 & 32.23 & 30.78 & 29.54 & 30.09 & 18.81 & 17.07 & 17.93 & 23.62 & 22.80 & 23.04 & 34.26 & 30.91 & 32.21 & 27.15 \\
EUDP\cite{eudp} & 33.24 & 30.55 & 32.27 & 31.10 & 29.12 & 30.15 & 19.65 & 15.91 & 18.07 & 24.60 & 22.51 & 23.40 & 34.40 & 29.25 & 32.59 & 27.12 \\
ACE\cite{ace} & 32.83 & 29.40 & 32.04 & 31.24 & 29.06 & 30.51 & 21.40 & 16.60 & 18.39 & 25.53 & 21.62 & 23.40 & 33.08 & 28.42 & 32.10 & 27.04 \\
AntiDB\cite{antidb} & 32.09 & 29.75 & 31.39 & 30.79 & 29.12 & 29.86 & 18.97 & 15.22 & 17.31 & 24.20 & 20.38 & 22.96 & 32.14 & 28.64 & 30.73 & 26.24 \\
SimAC\cite{simac} & 33.03 & 30.00 & 32.10 & 30.89 & 28.77 & 29.91 & 19.02 & 15.64 & 17.65 & 24.05 & 20.87 & 22.92 & 33.35 & 28.98 & 31.83 & 26.60 \\
MetaCloak\cite{metacloak} & 32.80 & 30.22 & 31.93 & 30.91 & 28.90 & 29.97 & 19.63 & 15.63 & 18.01 & 24.52 & 20.64 & 23.28 & 32.93 & 28.89 & 31.52 & 26.65 \\
DisDiff\cite{disdiff} & 33.01 & 30.37 & 32.04 & 30.94 & 29.07 & 30.07 & 19.41 & 16.01 & 18.07 & 24.25 & 21.34 & 23.18 & 33.29 & 28.94 & 31.85 & 26.79 \\
CAAT\cite{caat} & 33.03 & 30.79 & 32.14 & 31.04 & 29.26 & 30.09 & 19.62 & 16.00 & 17.89 & 24.58 & 20.94 & 23.35 & 32.61 & 28.66 & 31.12 & 26.74 \\
PAP\cite{pap} & 31.93 & 29.70 & 31.21 & 30.46 & 28.85 & 29.61 & 19.02 & 14.77 & 19.02 & 23.59 & 23.02 & 23.10 & 31.75 & 28.56 & 30.41 & 26.33 \\
AntiDiff\cite{antidiffusion} & 32.73 & 30.22 & 31.81 & 30.82 & 29.01 & 29.84 & 19.22 & 15.41 & 17.52 & 24.32 & 20.31 & 23.06 & 32.65 & 28.82 & 31.16 & 26.46 \\
GoodAC\cite{goodac} & 32.29 & 29.95 & 31.54 & 30.58 & 28.91 & 29.67 & 18.83 & 15.11 & 17.18 & 23.98 & 20.25 & 22.79 & 31.71 & 28.51 & 30.36 & 26.11 \\
Pretender\cite{pretender} & 33.17 & 30.68 & 32.21 & 31.14 & 29.27 & 30.19 & 19.61 & 15.94 & 17.89 & 24.55 & 20.97 & 23.30 & 32.80 & 28.78 & 31.29 & 26.79 \\
HAAD\cite{haad} & 32.37 & 30.91 & 32.13 & 30.66 & 29.54 & 30.06 & 18.50 & 16.39 & 17.67 & 23.73 & 22.55 & 23.03 & 33.83 & 29.80 & 32.10 & 26.88 \\
\midrule
\rowcolor{Gray}
{\sys (Ours)} & \textbf{19.67} & \textbf{19.68} & \textbf{19.67} & \textbf{20.30} & \textbf{20.30} & \textbf{20.30} & \textbf{16.02} & \textbf{14.35} & \textbf{15.21} & \textbf{21.66} & \textbf{19.70} & \textbf{21.03} & \textbf{18.10} & \textbf{18.13} & \textbf{18.10} & \textbf{18.82} \\
\bottomrule
\end{tabular}
}
\end{table*}
\begin{table*}[htbp]
\centering
\caption{\textbf{CLIP-I scores} assessing identity preservation under \textbf{inference-based mimicry attacks} (SDEdit~\cite{sdedit}, Inpainting~\cite{ldm}, ControlNet~\cite{controlnet}) across five datasets. The best and second-best results are highlighted in \textbf{bold} and \underline{underline}, respectively.}
\label{tab:infer_clip_i}

\setlength{\tabcolsep}{2.5pt}
\renewcommand{\arraystretch}{0.95}
\resizebox{\textwidth}{!}{
\begin{tabular}{l ccc ccc ccc ccc ccc c}
\toprule

\multirow{2}{*}{\textbf{Defense}} & \multicolumn{3}{c}{\textbf{TI-Dataset\cite{ti}}} & \multicolumn{3}{c}{\textbf{DB-Dataset\cite{dreambooth}}} & \multicolumn{3}{c}{\textbf{CelebA-HQ\cite{celeba-hq}}} & \multicolumn{3}{c}{\textbf{VGGFace2\cite{vggface2}}} & \multicolumn{3}{c}{\textbf{WikiArt\cite{wikiart}}} & \multirow{2}{*}{\textbf{AVG}} \\

\cmidrule(lr){2-4} \cmidrule(lr){5-7} \cmidrule(lr){8-10} \cmidrule(lr){11-13} \cmidrule(lr){14-16}
& SE\cite{sdedit} & IP\cite{ldm} & CN\cite{controlnet} & SE\cite{sdedit} & IP\cite{ldm} & CN\cite{controlnet} & SE\cite{sdedit} & IP\cite{ldm} & CN\cite{controlnet} & SE\cite{sdedit} & IP\cite{ldm} & CN\cite{controlnet} & SE\cite{sdedit} & IP\cite{ldm} & CN\cite{controlnet} & \\
\midrule
RandomNoise & 0.91 & 0.96 & 0.93 & 0.94 & 0.97 & 0.96 & 0.86 & 0.96 & 0.89 & 0.85 & 0.94 & 0.88 & 0.96 & 0.99 & 0.97 & 0.93 \\
LDMR\cite{ldmr} & 0.81 & 0.85 & 0.84 & 0.83 & 0.87 & 0.86 & 0.72 & 0.83 & 0.75 & 0.72 & 0.82 & 0.76 & 0.86 & 0.88 & 0.88 & 0.82 \\
AdvDM\cite{advdm} & 0.86 & 0.88 & 0.89 & 0.87 & 0.89 & 0.90 & 0.77 & 0.83 & 0.81 & 0.76 & 0.82 & 0.79 & 0.89 & 0.90 & 0.90 & 0.85 \\
MIST\cite{mist} & 0.82 & 0.84 & 0.86 & 0.82 & 0.85 & 0.86 & 0.71 & \underline{0.78} & 0.78 & 0.68 & \underline{0.76} & 0.76 & 0.87 & 0.87 & 0.89 & 0.81 \\
PhotoGuard\cite{photoguard} & 0.85 & 0.90 & 0.88 & 0.87 & 0.91 & 0.90 & 0.75 & 0.85 & 0.80 & 0.74 & 0.85 & 0.79 & 0.89 & 0.91 & 0.91 & 0.85 \\
Glaze\cite{glaze} & 0.81 & 0.84 & 0.85 & 0.82 & 0.85 & 0.86 & 0.71 & 0.78 & 0.77 & 0.68 & 0.77 & 0.76 & 0.86 & 0.87 & 0.87 & 0.81 \\
Smooth\cite{smooth} & \underline{0.79} & 0.85 & \underline{0.82} & 0.82 & 0.87 & 0.85 & \underline{0.67} & 0.83 & \underline{0.70} & \underline{0.66} & 0.82 & \underline{0.69} & \underline{0.78} & 0.89 & \underline{0.81} & \underline{0.79} \\
SDST\cite{sdst} & 0.82 & 0.86 & 0.86 & 0.82 & 0.87 & 0.86 & 0.70 & 0.85 & 0.77 & 0.68 & 0.84 & 0.75 & 0.87 & 0.87 & 0.89 & 0.82 \\
PID\cite{pid} & 0.83 & 0.86 & 0.87 & 0.85 & 0.88 & 0.88 & 0.74 & 0.86 & 0.79 & 0.73 & 0.85 & 0.78 & 0.88 & 0.92 & 0.90 & 0.84 \\
DiffGuard\cite{diffusionguard} & 0.88 & 0.91 & 0.90 & 0.89 & 0.93 & 0.92 & 0.86 & 0.91 & 0.88 & 0.82 & 0.89 & 0.85 & 0.87 & 0.93 & 0.91 & 0.89 \\
NightShade\cite{nightshade} & 0.81 & 0.84 & 0.85 & 0.81 & 0.85 & 0.85 & 0.70 & 0.78 & 0.77 & 0.67 & 0.77 & 0.75 & 0.85 & 0.86 & 0.87 & 0.80 \\
GAPDiff\cite{gapdiff} & 0.86 & 0.88 & 0.88 & 0.89 & 0.90 & 0.90 & 0.78 & 0.84 & 0.81 & 0.76 & 0.84 & 0.80 & 0.88 & 0.91 & 0.90 & 0.86 \\
LDS\cite{lds} & 0.92 & 0.90 & 0.90 & 0.93 & 0.91 & 0.92 & 0.91 & 0.87 & 0.87 & 0.90 & 0.85 & 0.85 & 0.94 & 0.90 & 0.91 & 0.90 \\
EUDP\cite{eudp} & 0.88 & 0.88 & 0.90 & 0.89 & 0.89 & 0.91 & 0.83 & 0.85 & 0.86 & 0.82 & 0.84 & 0.84 & 0.91 & 0.89 & 0.91 & 0.87 \\
ACE\cite{ace} & 0.81 & \underline{0.83} & 0.85 & \underline{0.81} & \underline{0.84} & \underline{0.85} & 0.72 & 0.79 & 0.77 & 0.69 & 0.77 & 0.76 & 0.86 & \underline{0.84} & 0.88 & 0.80 \\
AntiDB\cite{antidb} & 0.85 & 0.87 & 0.87 & 0.87 & 0.88 & 0.90 & 0.77 & 0.85 & 0.80 & 0.77 & 0.83 & 0.80 & 0.83 & 0.89 & 0.86 & 0.84 \\
SimAC\cite{simac} & 0.89 & 0.90 & 0.91 & 0.91 & 0.91 & 0.93 & 0.86 & 0.86 & 0.88 & 0.85 & 0.86 & 0.86 & 0.88 & 0.90 & 0.91 & 0.89 \\
MetaCloak\cite{metacloak} & 0.88 & 0.89 & 0.90 & 0.89 & 0.90 & 0.91 & 0.82 & 0.86 & 0.85 & 0.81 & 0.85 & 0.84 & 0.86 & 0.91 & 0.88 & 0.87 \\
DisDiff\cite{disdiff} & 0.88 & 0.89 & 0.90 & 0.89 & 0.90 & 0.91 & 0.85 & 0.85 & 0.87 & 0.84 & 0.85 & 0.86 & 0.87 & 0.89 & 0.89 & 0.88 \\
CAAT\cite{caat} & 0.87 & 0.88 & 0.89 & 0.88 & 0.89 & 0.91 & 0.80 & 0.84 & 0.83 & 0.80 & 0.83 & 0.82 & 0.84 & 0.89 & 0.87 & 0.86 \\
PAP\cite{pap} & 0.84 & 0.87 & 0.87 & 0.86 & 0.88 & 0.89 & 0.74 & 0.84 & 0.77 & 0.74 & 0.84 & 0.77 & 0.83 & 0.89 & 0.85 & 0.83 \\
AntiDiff\cite{antidiffusion} & 0.87 & 0.89 & 0.89 & 0.88 & 0.90 & 0.91 & 0.79 & 0.86 & 0.82 & 0.79 & 0.85 & 0.82 & 0.85 & 0.91 & 0.88 & 0.86 \\
GoodAC\cite{goodac} & 0.85 & 0.87 & 0.88 & 0.87 & 0.89 & 0.90 & 0.76 & 0.84 & 0.79 & 0.76 & 0.84 & 0.79 & 0.83 & 0.89 & 0.85 & 0.84 \\
Pretender\cite{pretender} & 0.87 & 0.87 & 0.89 & 0.88 & 0.89 & 0.90 & 0.81 & 0.85 & 0.83 & 0.80 & 0.84 & 0.82 & 0.85 & 0.90 & 0.88 & 0.86 \\
HAAD\cite{haad} & 0.85 & 0.85 & 0.88 & 0.87 & 0.86 & 0.90 & 0.75 & 0.79 & 0.78 & 0.74 & 0.78 & 0.78 & 0.88 & 0.86 & 0.88 & 0.83 \\
\midrule
\rowcolor{Gray}
{\sys (Ours)} & \textbf{0.52} & \textbf{0.55} & \textbf{0.53} & \textbf{0.51} & \textbf{0.51} & \textbf{0.51} & \textbf{0.46} & \textbf{0.43} & \textbf{0.47} & \textbf{0.46} & \textbf{0.42} & \textbf{0.45} & \textbf{0.50} & \textbf{0.48} & \textbf{0.50} & \textbf{0.49} \\
\bottomrule
\end{tabular}
}
\end{table*}
\begin{table*}[htbp]
\centering
\caption{\textbf{FID scores} quantifying visual degradation under \textbf{training-based mimicry attacks} (Fine-Tuning~\cite{ldm}, LoRA~\cite{lora}, SVDiff~\cite{svd}) across five datasets. The best and second-best results are highlighted in \textbf{bold} and \underline{underline}, respectively.}
\label{tab:train_fid}

\setlength{\tabcolsep}{2.5pt}
\renewcommand{\arraystretch}{0.95}
\resizebox{\textwidth}{!}{
\begin{tabular}{l ccc ccc ccc ccc ccc c}
\toprule

\multirow{2}{*}{\textbf{Defense}} & \multicolumn{3}{c}{\textbf{TI-Dataset\cite{ti}}} & \multicolumn{3}{c}{\textbf{DB-Dataset\cite{dreambooth}}} & \multicolumn{3}{c}{\textbf{CelebA-HQ\cite{celeba-hq}}} & \multicolumn{3}{c}{\textbf{VGGFace2\cite{vggface2}}} & \multicolumn{3}{c}{\textbf{WikiArt\cite{wikiart}}} & \multirow{2}{*}{\textbf{AVG}} \\

\cmidrule(lr){2-4} \cmidrule(lr){5-7} \cmidrule(lr){8-10} \cmidrule(lr){11-13} \cmidrule(lr){14-16}
& FT~\cite{ldm} & LR\cite{lora} & SV\cite{svd} & FT~\cite{ldm} & LR\cite{lora} & SV\cite{svd} & FT~\cite{ldm} & LR\cite{lora} & SV\cite{svd} & FT~\cite{ldm} & LR\cite{lora} & SV\cite{svd} & FT~\cite{ldm} & LR\cite{lora} & SV\cite{svd} & \\
\midrule
RandomNoise & 39.74 & 37.81 & 31.36 & 14.33 & 15.53 & 10.98 & 19.04 & 21.32 & 16.30 & 21.19 & 21.68 & 17.40 & 26.67 & 20.67 & 18.26 & 22.15 \\
LDMR\cite{ldmr} & 75.62 & 58.12 & 45.17 & 48.73 & 33.51 & 21.02 & 40.79 & 58.11 & 29.14 & 36.25 & 54.15 & 27.99 & 71.50 & 34.92 & 65.53 & 46.70 \\
AdvDM\cite{advdm} & 70.45 & 61.70 & 55.53 & 41.35 & 32.03 & 18.21 & 47.63 & 69.39 & 42.12 & 50.80 & 71.58 & 35.04 & 77.51 & 44.37 & 81.27 & 53.27 \\
MIST\cite{mist} & 214.15 & 207.87 & 78.80 & 107.54 & 123.91 & 33.62 & 92.88 & \underline{268.19} & 33.76 & 59.32 & 250.39 & 31.88 & 231.38 & 235.24 & 178.58 & 143.17 \\
PhotoGuard\cite{photoguard} & 46.77 & 51.74 & 36.75 & 25.45 & 20.87 & 13.54 & 25.87 & 36.17 & 21.71 & 29.68 & 36.33 & 22.17 & 47.86 & 29.92 & 37.90 & 32.18 \\
Glaze\cite{glaze} & \underline{221.54} & 197.81 & 75.96 & 101.67 & 111.43 & 32.36 & 76.74 & 258.31 & 34.77 & 58.82 & 242.94 & 32.29 & 221.35 & 227.83 & 188.21 & 138.80 \\
Smooth\cite{smooth} & 81.48 & 64.99 & 54.38 & 66.20 & 36.24 & 29.28 & \underline{115.46} & 62.22 & 39.95 & 71.58 & 61.70 & 30.71 & 120.09 & 42.62 & 112.48 & 65.96 \\
SDST\cite{sdst} & 97.42 & 60.43 & 46.34 & 43.76 & 30.81 & 21.09 & 47.80 & 49.39 & 28.25 & 44.37 & 52.27 & 27.03 & 72.58 & 35.04 & 64.79 & 48.09 \\
PID\cite{pid} & 60.47 & 53.67 & 40.10 & 28.65 & 25.08 & 18.08 & 29.49 & 40.70 & 24.75 & 28.84 & 37.19 & 23.43 & 41.44 & 26.04 & 34.19 & 34.14 \\
DiffGuard\cite{diffusionguard} & 47.56 & 52.05 & 37.95 & 28.91 & 21.32 & 15.38 & 23.44 & 30.57 & 26.10 & 25.21 & 30.30 & 24.40 & 41.44 & 25.40 & 39.28 & 31.29 \\
NightShade\cite{nightshade} & 218.33 & 188.80 & 77.74 & \underline{111.80} & 125.13 & \underline{34.72} & 91.84 & 259.70 & 34.43 & 51.37 & 247.19 & 31.35 & 232.74 & 227.16 & 182.29 & 140.97 \\
GAPDiff\cite{gapdiff} & 160.38 & 122.41 & 74.64 & 62.04 & 54.65 & 28.83 & 105.67 & 215.32 & 44.20 & \underline{88.71} & 203.19 & 38.72 & 186.43 & 86.09 & 88.05 & 103.96 \\
LDS\cite{lds} & 43.48 & 39.63 & 35.18 & 16.60 & 13.75 & 13.04 & 19.82 & 22.23 & 18.54 & 19.62 & 20.86 & 19.45 & 29.57 & 21.67 & 27.76 & 24.08 \\
EUDP\cite{eudp} & 94.02 & 73.24 & 62.19 & 60.63 & 37.76 & 30.98 & 74.20 & 72.75 & \underline{76.40} & 63.40 & 72.34 & \underline{54.67} & 172.21 & 41.32 & 162.77 & 76.59 \\
ACE\cite{ace} & 219.69 & \underline{208.58} & \underline{84.60} & 109.78 & \underline{133.78} & 34.50 & 72.88 & 262.11 & 32.93 & 50.98 & \underline{253.97} & 30.25 & \underline{254.56} & \underline{247.90} & \underline{196.74} & \underline{146.22} \\
AntiDB\cite{antidb} & 98.21 & 67.65 & 57.95 & 63.59 & 35.01 & 30.40 & 64.92 & 57.36 & 40.38 & 53.17 & 56.65 & 37.52 & 174.79 & 36.15 & 166.08 & 69.32 \\
SimAC\cite{simac} & 62.70 & 55.82 & 42.29 & 32.18 & 22.68 & 18.64 & 41.53 & 47.41 & 42.26 & 40.94 & 48.57 & 32.30 & 65.31 & 35.90 & 82.71 & 44.75 \\
MetaCloak\cite{metacloak} & 74.07 & 57.65 & 39.55 & 43.75 & 27.25 & 18.17 & 35.51 & 46.05 & 28.41 & 36.15 & 42.99 & 26.53 & 79.86 & 32.78 & 71.71 & 44.03 \\
DisDiff\cite{disdiff} & 98.25 & 63.43 & 62.73 & 55.21 & 34.42 & 29.12 & 68.17 & 70.91 & 75.18 & 66.78 & 75.35 & 54.41 & 143.21 & 41.80 & 140.55 & 71.97 \\
CAAT\cite{caat} & 94.65 & 67.94 & 54.10 & 58.48 & 34.88 & 27.40 & 63.78 & 56.75 & 45.20 & 59.45 & 59.30 & 38.82 & 169.68 & 37.89 & 150.20 & 67.90 \\
PAP\cite{pap} & 100.78 & 66.03 & 68.06 & 64.15 & 35.03 & 33.29 & 72.61 & 62.22 & 41.77 & 57.30 & 59.00 & 32.62 & 184.83 & 39.95 & 166.19 & 72.26 \\
AntiDiff\cite{antidiffusion} & 77.79 & 56.70 & 49.12 & 51.17 & 30.79 & 23.20 & 49.57 & 53.49 & 37.13 & 46.06 & 47.29 & 31.37 & 126.24 & 31.69 & 118.89 & 55.37 \\
GoodAC\cite{goodac} & 92.56 & 66.39 & 57.86 & 58.09 & 34.51 & 30.41 & 68.97 & 62.06 & 40.51 & 66.88 & 58.66 & 36.43 & 184.67 & 40.27 & 161.18 & 70.63 \\
Pretender\cite{pretender} & 91.25 & 70.87 & 52.76 & 53.04 & 32.06 & 23.43 & 46.48 & 51.35 & 37.66 & 43.87 & 48.41 & 35.49 & 125.97 & 33.55 & 108.05 & 56.95 \\
HAAD\cite{haad} & 98.58 & 66.15 & 61.24 & 61.43 & 34.06 & 30.41 & 69.77 & 59.68 & 45.23 & 59.37 & 61.45 & 39.12 & 177.91 & 39.30 & 161.61 & 71.02 \\
\midrule
\rowcolor{Gray}
{\sys (Ours)} & \textbf{416.51} & \textbf{384.01} & \textbf{343.43} & \textbf{432.78} & \textbf{388.56} & \textbf{340.42} & \textbf{309.78} & \textbf{477.48} & \textbf{454.15} & \textbf{371.79} & \textbf{475.75} & \textbf{465.24} & \textbf{403.33} & \textbf{408.27} & \textbf{402.64} & \textbf{404.94} \\
\bottomrule
\end{tabular}
}
\end{table*}
\begin{table*}[htbp]
\centering
\caption{\textbf{CLIP-S scores} measuring semantic alignment under \textbf{training-based mimicry attacks} (Fine-Tuning~\cite{ldm}, LoRA~\cite{lora}, SVDiff~\cite{svd}) across five datasets. The best and second-best results are highlighted in \textbf{bold} and \underline{underline}, respectively.}
\label{tab:train_clip_s}

\setlength{\tabcolsep}{2.5pt}
\renewcommand{\arraystretch}{0.95}
\resizebox{\textwidth}{!}{
\begin{tabular}{l ccc ccc ccc ccc ccc c}
\toprule

\multirow{2}{*}{\textbf{Defense}} & \multicolumn{3}{c}{\textbf{TI-Dataset\cite{ti}}} & \multicolumn{3}{c}{\textbf{DB-Dataset\cite{dreambooth}}} & \multicolumn{3}{c}{\textbf{CelebA-HQ\cite{celeba-hq}}} & \multicolumn{3}{c}{\textbf{VGGFace2\cite{vggface2}}} & \multicolumn{3}{c}{\textbf{WikiArt\cite{wikiart}}} & \multirow{2}{*}{\textbf{AVG}} \\

\cmidrule(lr){2-4} \cmidrule(lr){5-7} \cmidrule(lr){8-10} \cmidrule(lr){11-13} \cmidrule(lr){14-16}
& FT~\cite{ldm} & LR\cite{lora} & SV\cite{svd} & FT~\cite{ldm} & LR\cite{lora} & SV\cite{svd} & FT~\cite{ldm} & LR\cite{lora} & SV\cite{svd} & FT~\cite{ldm} & LR\cite{lora} & SV\cite{svd} & FT~\cite{ldm} & LR\cite{lora} & SV\cite{svd} & \\
\midrule
RandomNoise & 29.28 & 29.59 & 31.80 & 29.28 & 30.17 & 31.48 & 23.60 & 25.10 & 24.99 & 27.01 & 27.84 & \underline{27.82} & 33.27 & 33.14 & 32.99 & 29.16 \\
LDMR\cite{ldmr} & 29.94 & 29.40 & 31.39 & 29.81 & 30.52 & 31.75 & 23.80 & 24.25 & 25.22 & 27.70 & 27.61 & 28.46 & 31.58 & 32.70 & 31.40 & 29.03 \\
AdvDM\cite{advdm} & 30.47 & 29.95 & 31.36 & 29.58 & 30.90 & 31.96 & 22.91 & 24.54 & 25.31 & 26.87 & 28.07 & 28.42 & 32.37 & 33.42 & 31.60 & 29.18 \\
MIST\cite{mist} & 24.26 & 23.81 & 29.46 & 27.32 & 26.11 & 31.37 & 21.72 & 17.04 & 25.19 & 26.87 & \underline{22.68} & 28.01 & \underline{19.19} & 22.33 & 22.51 & 24.53 \\
PhotoGuard\cite{photoguard} & 29.97 & 30.29 & 31.60 & 29.83 & 30.65 & 31.68 & 23.96 & 24.82 & 25.30 & 27.26 & 28.09 & 28.24 & 31.98 & 32.94 & 31.38 & 29.20 \\
Glaze\cite{glaze} & 24.10 & 23.90 & 29.61 & 27.16 & 26.59 & 31.49 & 22.29 & 17.11 & 25.29 & 26.81 & 22.74 & 28.02 & 19.50 & 22.44 & \underline{21.57} & 24.57 \\
Smooth\cite{smooth} & 30.37 & 29.49 & 31.11 & 30.02 & 30.59 & 31.59 & 21.58 & 24.15 & \underline{24.52} & 26.62 & 27.98 & 28.12 & 30.56 & 32.52 & 31.01 & 28.68 \\
SDST\cite{sdst} & 27.99 & 29.71 & 30.92 & 28.15 & 30.54 & 31.56 & 23.05 & 24.40 & 25.42 & 27.10 & 27.66 & 28.19 & 28.87 & 31.10 & 28.33 & 28.20 \\
PID\cite{pid} & 29.54 & 29.80 & 31.21 & 29.52 & 31.10 & 31.83 & 24.21 & 24.89 & 25.47 & 27.86 & 28.31 & 28.28 & 32.34 & 32.89 & 31.67 & 29.26 \\
DiffGuard\cite{diffusionguard} & 30.62 & 30.22 & 31.72 & 30.18 & 31.15 & 31.90 & 24.27 & 25.28 & 25.44 & 27.50 & 28.23 & 28.41 & 32.78 & 33.41 & 33.66 & 29.65 \\
NightShade\cite{nightshade} & \underline{23.94} & 24.57 & 29.58 & \underline{26.89} & 26.15 & 31.42 & 21.45 & \underline{17.01} & 25.25 & 26.99 & 23.12 & 27.99 & 19.22 & 22.77 & 22.30 & 24.58 \\
GAPDiff\cite{gapdiff} & 27.23 & 26.83 & 29.70 & 29.01 & 29.33 & \underline{31.30} & \underline{20.86} & 18.06 & 24.73 & \underline{26.26} & 23.80 & 27.97 & 24.16 & 29.95 & 27.85 & 26.47 \\
LDS\cite{lds} & 29.97 & 30.24 & 31.55 & 29.31 & 30.26 & 31.54 & 23.49 & 24.86 & 25.06 & 27.28 & 27.68 & 28.18 & 28.37 & 32.61 & 28.49 & 28.59 \\
EUDP\cite{eudp} & 30.22 & 29.55 & 31.44 & 29.84 & 30.73 & 31.80 & 22.82 & 24.76 & 24.72 & 27.11 & 28.19 & 28.24 & 28.20 & 33.09 & 28.57 & 28.62 \\
ACE\cite{ace} & 24.26 & \underline{23.30} & \underline{29.42} & 27.05 & \underline{25.87} & 31.39 & 22.38 & 17.99 & 25.07 & 27.07 & 22.72 & 28.02 & 20.01 & \underline{21.44} & 21.83 & \underline{24.52} \\
AntiDB\cite{antidb} & 29.78 & 29.90 & 30.94 & 29.88 & 30.69 & 31.68 & 22.84 & 24.85 & 24.87 & 26.92 & 28.15 & 28.16 & 27.58 & 32.89 & 28.14 & 28.49 \\
SimAC\cite{simac} & 30.31 & 29.57 & 32.11 & 29.89 & 30.82 & 31.90 & 22.97 & 24.68 & 25.16 & 26.85 & 27.86 & 28.38 & 32.34 & 33.25 & 31.75 & 29.19 \\
MetaCloak\cite{metacloak} & 30.25 & 29.89 & 31.55 & 29.86 & 30.79 & 31.82 & 23.24 & 24.89 & 24.95 & 26.81 & 28.01 & 27.94 & 31.51 & 32.93 & 31.99 & 29.09 \\
DisDiff\cite{disdiff} & 29.93 & 29.93 & 31.13 & 29.82 & 30.90 & 31.78 & 22.45 & 24.83 & 24.72 & 26.75 & 28.48 & 28.46 & 29.40 & 33.11 & 29.61 & 28.75 \\
CAAT\cite{caat} & 29.94 & 30.20 & 31.61 & 30.09 & 30.66 & 31.81 & 22.48 & 24.88 & 25.11 & 26.91 & 28.47 & 28.37 & 28.25 & 33.06 & 29.16 & 28.73 \\
PAP\cite{pap} & 29.67 & 29.78 & 30.74 & 29.76 & 30.63 & 31.39 & 22.62 & 24.57 & 24.74 & 26.99 & 28.22 & 28.12 & 27.47 & 32.63 & 28.03 & 28.36 \\
AntiDiff\cite{antidiffusion} & 30.44 & 30.31 & 31.39 & 30.28 & 30.78 & 31.84 & 23.01 & 25.03 & 24.78 & 26.93 & 28.29 & 28.08 & 30.25 & 32.97 & 30.13 & 28.97 \\
GoodAC\cite{goodac} & 29.94 & 29.69 & 31.12 & 29.83 & 30.50 & 31.50 & 22.61 & 24.62 & 24.94 & 26.78 & 28.25 & 28.15 & 27.16 & 32.66 & 27.92 & 28.38 \\
Pretender\cite{pretender} & 30.01 & 29.89 & 31.67 & 30.01 & 30.72 & 31.94 & 23.15 & 24.99 & 25.08 & 27.07 & 28.30 & 28.23 & 30.19 & 32.90 & 30.51 & 28.98 \\
HAAD\cite{haad} & 29.76 & 29.87 & 31.03 & 29.91 & 30.75 & 31.82 & 22.66 & 24.67 & 24.87 & 26.83 & 26.12 & 28.14 & 30.50 & 32.99 & 30.55 & 28.70 \\
\midrule
\rowcolor{Gray}
{\sys (Ours)} & \textbf{19.71} & \textbf{19.20} & \textbf{19.60} & \textbf{20.48} & \textbf{20.14} & \textbf{20.47} & \textbf{18.82} & \textbf{16.94} & \textbf{18.13} & \textbf{23.31} & \textbf{22.63} & \textbf{23.26} & \textbf{18.72} & \textbf{19.88} & \textbf{19.86} & \textbf{20.08} \\
\bottomrule
\end{tabular}
}
\end{table*}
\begin{table*}[htbp]
\centering
\caption{\textbf{CLIP-I scores} assessing identity preservation under \textbf{training-based mimicry attacks} (Fine-Tuning~\cite{ldm}, LoRA~\cite{lora}, SVDiff~\cite{svd}) across five datasets. The best and second-best results are highlighted in \textbf{bold} and \underline{underline}, respectively.}
\label{tab:train_clip_i}

\setlength{\tabcolsep}{2.5pt}
\renewcommand{\arraystretch}{0.95}
\resizebox{\textwidth}{!}{
\begin{tabular}{l ccc ccc ccc ccc ccc c}
\toprule

\multirow{2}{*}{\textbf{Defense}} & \multicolumn{3}{c}{\textbf{TI-Dataset\cite{ti}}} & \multicolumn{3}{c}{\textbf{DB-Dataset\cite{dreambooth}}} & \multicolumn{3}{c}{\textbf{CelebA-HQ\cite{celeba-hq}}} & \multicolumn{3}{c}{\textbf{VGGFace2\cite{vggface2}}} & \multicolumn{3}{c}{\textbf{WikiArt\cite{wikiart}}} & \multirow{2}{*}{\textbf{AVG}} \\

\cmidrule(lr){2-4} \cmidrule(lr){5-7} \cmidrule(lr){8-10} \cmidrule(lr){11-13} \cmidrule(lr){14-16}
& FT~\cite{ldm} & LR\cite{lora} & SV\cite{svd} & FT~\cite{ldm} & LR\cite{lora} & SV\cite{svd} & FT~\cite{ldm} & LR\cite{lora} & SV\cite{svd} & FT~\cite{ldm} & LR\cite{lora} & SV\cite{svd} & FT~\cite{ldm} & LR\cite{lora} & SV\cite{svd} & \\
\midrule
RandomNoise & 0.87 & 0.86 & 0.87 & 0.89 & 0.88 & 0.91 & 0.84 & 0.86 & 0.90 & 0.84 & 0.85 & 0.89 & 0.86 & 0.88 & 0.92 & 0.88 \\
LDMR\cite{ldmr} & 0.79 & 0.80 & 0.80 & 0.80 & 0.82 & 0.84 & 0.76 & 0.75 & 0.83 & 0.78 & 0.75 & 0.83 & 0.75 & 0.84 & 0.76 & 0.79 \\
AdvDM\cite{advdm} & 0.79 & 0.79 & 0.78 & 0.81 & 0.83 & 0.86 & 0.76 & 0.77 & 0.82 & 0.76 & 0.77 & 0.83 & 0.79 & 0.85 & 0.77 & 0.80 \\
MIST\cite{mist} & \underline{0.64} & 0.68 & 0.75 & 0.69 & 0.67 & 0.82 & 0.70 & 0.57 & 0.82 & 0.73 & \underline{0.56} & 0.81 & 0.56 & 0.65 & 0.64 & 0.69 \\
PhotoGuard\cite{photoguard} & 0.83 & 0.82 & 0.84 & 0.85 & 0.86 & 0.88 & 0.80 & 0.80 & 0.86 & 0.79 & 0.79 & 0.85 & 0.79 & 0.85 & 0.80 & 0.83 \\
Glaze\cite{glaze} & 0.64 & 0.68 & 0.75 & 0.70 & 0.69 & 0.82 & 0.71 & 0.57 & 0.82 & 0.73 & 0.57 & 0.81 & 0.56 & 0.66 & \underline{0.62} & 0.69 \\
Smooth\cite{smooth} & 0.78 & 0.78 & 0.79 & 0.79 & 0.81 & 0.83 & 0.69 & 0.75 & 0.80 & 0.73 & 0.75 & 0.83 & 0.72 & 0.83 & 0.74 & 0.78 \\
SDST\cite{sdst} & 0.79 & 0.81 & 0.81 & 0.81 & 0.85 & 0.85 & 0.74 & 0.77 & 0.83 & 0.75 & 0.75 & 0.83 & 0.73 & 0.82 & 0.74 & 0.79 \\
PID\cite{pid} & 0.82 & 0.81 & 0.83 & 0.84 & 0.85 & 0.87 & 0.78 & 0.78 & 0.84 & 0.80 & 0.79 & 0.84 & 0.81 & 0.86 & 0.82 & 0.82 \\
DiffGuard\cite{diffusionguard} & 0.83 & 0.82 & 0.83 & 0.84 & 0.85 & 0.87 & 0.82 & 0.83 & 0.86 & 0.83 & 0.83 & 0.87 & 0.83 & 0.87 & 0.85 & 0.84 \\
NightShade\cite{nightshade} & 0.64 & 0.69 & 0.74 & 0.69 & 0.68 & 0.82 & 0.69 & 0.57 & 0.82 & 0.74 & 0.57 & 0.81 & 0.56 & 0.67 & 0.63 & 0.69 \\
GAPDiff\cite{gapdiff} & 0.68 & 0.71 & 0.75 & 0.77 & 0.76 & 0.84 & \underline{0.67} & 0.57 & 0.79 & \underline{0.69} & 0.58 & 0.80 & 0.61 & 0.75 & 0.71 & 0.71 \\
LDS\cite{lds} & 0.86 & 0.83 & 0.84 & 0.88 & 0.87 & 0.89 & 0.83 & 0.83 & 0.86 & 0.74 & 0.76 & 0.81 & 0.68 & 0.84 & 0.70 & 0.81 \\
EUDP\cite{eudp} & 0.76 & 0.77 & 0.77 & 0.77 & 0.80 & 0.81 & 0.74 & 0.76 & \underline{0.76} & 0.74 & 0.75 & 0.79 & 0.67 & 0.84 & 0.69 & 0.76 \\
ACE\cite{ace} & 0.64 & \underline{0.67} & \underline{0.74} & \underline{0.69} & \underline{0.67} & 0.81 & 0.71 & \underline{0.56} & 0.82 & 0.74 & 0.56 & 0.82 & \underline{0.56} & \underline{0.64} & 0.63 & \underline{0.69} \\
AntiDB\cite{antidb} & 0.76 & 0.78 & 0.77 & 0.78 & 0.80 & 0.81 & 0.74 & 0.76 & 0.81 & 0.75 & 0.76 & 0.81 & 0.65 & 0.84 & 0.68 & 0.77 \\
SimAC\cite{simac} & 0.81 & 0.81 & 0.82 & 0.84 & 0.86 & 0.87 & 0.78 & 0.80 & 0.82 & 0.79 & 0.80 & 0.85 & 0.82 & 0.87 & 0.80 & 0.82 \\
MetaCloak\cite{metacloak} & 0.80 & 0.80 & 0.82 & 0.82 & 0.84 & 0.86 & 0.80 & 0.79 & 0.84 & 0.79 & 0.80 & 0.84 & 0.79 & 0.85 & 0.78 & 0.81 \\
DisDiff\cite{disdiff} & 0.76 & 0.79 & 0.77 & 0.78 & 0.81 & 0.81 & 0.74 & 0.77 & 0.76 & 0.73 & 0.77 & \underline{0.79} & 0.71 & 0.85 & 0.73 & 0.77 \\
CAAT\cite{caat} & 0.76 & 0.78 & 0.78 & 0.78 & 0.80 & 0.82 & 0.74 & 0.76 & 0.80 & 0.74 & 0.75 & 0.81 & 0.68 & 0.85 & 0.70 & 0.77 \\
PAP\cite{pap} & 0.75 & 0.78 & 0.76 & 0.77 & 0.80 & \underline{0.81} & 0.72 & 0.76 & 0.81 & 0.75 & 0.76 & 0.82 & 0.65 & 0.84 & 0.68 & 0.76 \\
AntiDiff\cite{antidiffusion} & 0.79 & 0.80 & 0.79 & 0.80 & 0.82 & 0.84 & 0.76 & 0.78 & 0.82 & 0.77 & 0.78 & 0.82 & 0.72 & 0.85 & 0.73 & 0.79 \\
GoodAC\cite{goodac} & 0.76 & 0.79 & 0.77 & 0.78 & 0.81 & 0.81 & 0.72 & 0.76 & 0.81 & 0.74 & 0.76 & 0.81 & 0.65 & 0.83 & 0.68 & 0.77 \\
Pretender\cite{pretender} & 0.77 & 0.78 & 0.79 & 0.80 & 0.81 & 0.84 & 0.76 & 0.77 & 0.81 & 0.77 & 0.77 & 0.82 & 0.73 & 0.85 & 0.73 & 0.79 \\
HAAD\cite{haad} & 0.75 & 0.78 & 0.77 & 0.78 & 0.80 & 0.81 & 0.73 & 0.76 & 0.81 & 0.74 & 0.66 & 0.81 & 0.68 & 0.84 & 0.70 & 0.76 \\
\midrule
\rowcolor{Gray}
{\sys (Ours)} & \textbf{0.56} & \textbf{0.57} & \textbf{0.54} & \textbf{0.55} & \textbf{0.53} & \textbf{0.52} & \textbf{0.50} & \textbf{0.50} & \textbf{0.49} & \textbf{0.49} & \textbf{0.50} & \textbf{0.49} & \textbf{0.55} & \textbf{0.59} & \textbf{0.59} & \textbf{0.53} \\
\bottomrule
\end{tabular}
}
\end{table*}

\begin{table*}[htbp!]
\centering
\caption{Complete Imperceptibility Evaluation. Expanding on Table~\ref{tab:imperceptibility} in the main text.}
\label{tab:full_imperceptibility}
\setlength{\tabcolsep}{0.56pt}
\renewcommand{\arraystretch}{0.8}
\resizebox{\textwidth}{!}{
\begin{tabular}{ll ccccc ccccc}
\toprule
\multirow{2}{*}{\textbf{Type }} & \multirow{2}{*}{\textbf{Defense}} & \multicolumn{5}{c}{\textbf{Traditional Metrics}} & \multicolumn{5}{c}{\textbf{Neural Metrics}} \\
\cmidrule(lr){3-7} \cmidrule(lr){8-12}
& & PSNR~\cite{psnr}$\uparrow$ & SSIM~\cite{ssim}$\uparrow$ & FSIM~\cite{fsim}$\uparrow$ & VIF~\cite{vif}$\uparrow$ & GMSD~\cite{gmsd}$\downarrow$ & LPIPS~\cite{lpips}$\downarrow$ & DISTS~\cite{dists}$\downarrow$ & DeepIQA~\cite{deepiqa}$\uparrow$ & PieAPP~\cite{pieapp}$\downarrow$ & ContentLoss~\cite{content}$\downarrow$ \\
\midrule
\multirow{11}{*}{\rotatebox{90}{\textbf{Poisoning}}} & ACE~\cite{ace} & 31.957 & 0.813 & 0.964 & 0.517 & 0.049 & 0.158 & 0.317 & -0.023 & 0.922 & 2729.8 \\
 & AntiDB~\cite{antidb} & 32.223 & 0.824 & 0.956 & 0.549 & 0.058 & 0.179 & 0.207 & -0.030 & 1.535 & 1418.4 \\
 & AntiDiff~\cite{antidiffusion} & 33.708 & \underline{0.870} & 0.965 & 0.598 & 0.048 & 0.162 & 0.199 & -0.026 & 1.330 & 1297.6 \\
 & CAAT~\cite{caat} & 32.325 & 0.828 & 0.955 & 0.557 & 0.060 & 0.180 & 0.209 & -0.029 & 1.554 & 1401.1 \\
 & DisDiff~\cite{disdiff} & 32.854 & 0.847 & 0.959 & 0.577 & 0.054 & 0.167 & 0.221 & -0.027 & 1.442 & 1443.1 \\
 & EUDP~\cite{eudp} & 32.426 & 0.832 & 0.956 & 0.560 & 0.057 & 0.173 & 0.223 & -0.028 & 1.490 & 1469.7 \\
 & GoodAC~\cite{goodac} & 32.214 & 0.824 & 0.956 & 0.549 & 0.058 & 0.178 & 0.205 & -0.030 & 1.545 & 1412.9 \\
 & MetaCloak~\cite{metacloak} & 32.293 & 0.836 & 0.933 & 0.546 & 0.083 & 0.189 & 0.217 & -0.032 & 1.702 & 1387.1 \\
 & PAP~\cite{pap} & 32.218 & 0.825 & 0.956 & 0.550 & 0.058 & 0.179 & 0.204 & -0.030 & 1.553 & 1409.7 \\
 & Pretender~\cite{pretender} & 32.735 & 0.842 & 0.958 & 0.572 & 0.056 & 0.183 & 0.213 & -0.029 & 1.455 & 1399.9 \\
 & SimAC~\cite{simac} & 32.976 & 0.845 & 0.961 & 0.573 & 0.051 & 0.145 & 0.219 & -0.025 & 1.313 & 1449.8 \\
 & HAAD~\cite{haad} & 32.304 & 0.826 & 0.956 & 0.552 & 0.058 & 0.179 & 0.206 & -0.024 & 1.541 & 1413.4 \\
\midrule
\multirow{12}{*}{\rotatebox{90}{\textbf{Evasion}}} & AdvDM~\cite{advdm} & 33.253 & 0.846 & 0.966 & 0.584 & 0.046 & 0.161 & 0.250 & -0.023 & 1.161 & 1716.1 \\
 & DiffGuard~\cite{diffusionguard} & \underline{33.841} & 0.854 & 0.975 & \underline{0.611} & 0.036 & 0.135 & \underline{0.192} & -0.019 & 0.912 & \underline{1247.8} \\
 & GAPDiff~\cite{gapdiff} & 31.205 & 0.788 & 0.943 & 0.495 & 0.075 & 0.222 & 0.259 & -0.040 & 1.356 & 2025.6 \\
 & Glaze~\cite{glaze} & 31.890 & 0.814 & 0.964 & 0.513 & 0.049 & 0.158 & 0.313 & -0.024 & 0.890 & 2724.3 \\
 & LDMR~\cite{ldmr} & 31.680 & 0.781 & 0.955 & 0.507 & 0.061 & 0.229 & 0.244 & -0.035 & 1.328 & 1688.7 \\
 & MIST~\cite{mist} & 32.014 & 0.821 & 0.965 & 0.521 & 0.049 & 0.159 & 0.309 & -0.023 & 0.887 & 2693.0 \\
 & NightShade~\cite{nightshade} & 31.921 & 0.810 & 0.965 & 0.514 & 0.047 & 0.160 & 0.326 & -0.023 & 0.890 & 2896.1 \\
 & PID~\cite{pid} & 32.428 & 0.809 & 0.963 & 0.543 & 0.044 & 0.174 & 0.238 & -0.021 & 1.167 & 1570.1 \\
 & PhotoGuard~\cite{photoguard} & 33.367 & 0.843 & \underline{0.979} & 0.584 & \underline{0.028} & 0.122 & 0.219 & \underline{-0.014} & 0.728 & 1589.5 \\
 & SDST~\cite{sdst} & 32.903 & 0.823 & 0.977 & 0.534 & 0.028 & \underline{0.116} & 0.291 & -0.017 & \underline{0.716} & 2182.5 \\
 & Smooth~\cite{smooth} & 32.173 & 0.817 & 0.954 & 0.535 & 0.062 & 0.186 & 0.204 & -0.029 & 1.509 & 1328.8 \\
 & LDS~\cite{lds} & 31.254 & 0.864 & 0.976 & 0.604 & 0.037 & 0.125 & 0.198 & -0.021 & 0.717 & 1352.6 \\
\cmidrule{2-12}
\rowcolor{Gray}
 & {\sys (Ours)} & \textbf{34.408} & \textbf{0.888} & \textbf{0.985} & \textbf{0.624} & \textbf{0.019} & \textbf{0.096} & \textbf{0.177} & \textbf{-0.010} & \textbf{0.679} & \textbf{1204.9} \\
\bottomrule
\end{tabular}
}
\end{table*}

\end{document}